\documentclass{article}

\PassOptionsToPackage{numbers, sort&compress}{natbib}

\usepackage[main,final]{neurips_2026}

\usepackage[utf8]{inputenc} 
\usepackage[T1]{fontenc}    
\usepackage{hyperref}       
\usepackage{url}            
\usepackage{booktabs}       
\usepackage{amsfonts}       
\usepackage{nicefrac}       
\usepackage{microtype}      
\usepackage[table]{xcolor}         

\usepackage{graphicx}
\usepackage{amsmath}
\usepackage{amssymb}
\usepackage{mathabx}
\usepackage{caption}
\definecolor{darkred}{RGB}{245,0,0}
\usepackage{algorithm}
\usepackage{algorithmic}
\usepackage{multirow}
\usepackage{subfigure}
\usepackage{makecell}
\usepackage{tabularx}
\usepackage{titletoc}
\usepackage{enumitem}
\usepackage[normalem]{ulem}
\usepackage{xspace}
\newcommand{\ie}{i.e.,\xspace}
\newcommand{\eg}{e.g.,\xspace}

\title{\fontsize{16.1pt}{19.3pt}\selectfont VIGOR: Zero-Shot Visual Generalization via Latent-Space Consistency in Model-Based Reinforcement Learning}

\author{%
  Mingyu Park\thanks{Equal contribution.}\\
  Robotics Program\\
  KAIST, ETRI\\
  \texttt{min9yu@kaist.ac.kr}\\
  \texttt{mngupark@etri.re.kr}\\  
  \And
  Samyeul Noh\footnotemark[1]\\
  Electrical Engineering\\
  KAIST, ETRI\\
  \texttt{samuel@kaist.ac.kr}\\
  \texttt{samuel@etri.re.kr}\\
  \AND
  Hyun Myung\thanks{Corresponding authors. Correspondence to: hmyung@kaist.ac.kr, donghwan@kaist.ac.kr}\\
  Electrical Engineering\\
  KAIST\\
  \texttt{hmyung@kaist.ac.kr}\\
  \And
  Donghwan Lee\footnotemark[2]\\
  Electrical Engineering\\
  KAIST\\
  \texttt{donghwan@kaist.ac.kr}\\
}

\begin{document}

\maketitle

\vspace{-0.10cm}
\begin{abstract}
\vspace{-0.10cm}
  Model-based reinforcement learning (MBRL) achieves strong sample efficiency by planning within learned latent dynamics, yet its performance degrades substantially under unseen visual distractions such as background variations, lighting changes, or camera shifts. Unlike model-free RL, where encoder perturbations affect only single-step predictions, MBRL suffers from a two-level vulnerability: visual distractions first push encoder outputs out of distribution, and these errors then compound through recursive latent rollouts over the planning horizon. We propose \textbf{VI}sual \textbf{G}eneralization via latent-space c\textbf{O}nsistency in model-based \textbf{R}L (\textbf{VIGOR}), a framework that enables zero-shot generalization to unseen visual distractions while retaining the sample efficiency of its MBRL backbone. VIGOR enforces latent-space consistency through three interdependent components: (i)~\textit{asymmetric weak-to-strong augmentation}, which pairs weak-only and weak-to-strong latent views within a single batch; (ii)~\textit{dynamics-level consistency}, which enforces augmentation-invariant transition predictions through direct latent regression; and (iii)~\textit{encoder-level stabilization}, which prevents encoder drift under the cross-augmentation supervision imposed by dynamics-level consistency. Evaluations on the DeepMind Control Suite (DMC) and Robosuite show that VIGOR, instantiated on TD-MPC2, outperforms state-of-the-art model-free and model-based baselines in zero-shot generalization, surpassing the second-best baseline by $3.4\%$ on DMC and $43.6\%$ on Robosuite. Moreover, VIGOR achieves these improvements while maintaining sample efficiency comparable to that of its backbone and requiring fewer environment interactions than the other evaluated baselines. Ablations further show that VIGOR's robustness is augmentation-agnostic: replacing the default augmentation with alternatives from distinct perturbation families preserves strong generalization, confirming that latent-space consistency, not the augmentation choice, drives robustness. Finally, a second instantiation of VIGOR on DreamerV3, a structurally distinct recurrent state-space world model, also improves zero-shot generalization over the DreamerV3 baseline on both benchmarks, suggesting that the benefits of VIGOR are not limited to the TD-MPC2 backbone. Code and videos are available on our project page: \url{https://vigor-gen.github.io/}.
\end{abstract}

\vspace{-0.10cm}
\section{Introduction} \label{sec:introduction}
\vspace{-0.15cm}
Visual reinforcement learning (visual RL) trains control policies directly from pixels, eliminating task-specific state estimation~\citep{laskin2020reinforcement,laskin2020curl,yarats2021image,yarats2021mastering}. Despite success in locomotion and manipulation~\citep{haarnoja2018soft,kalashnikov2018scalable,hansen2023td}, visual RL remains sensitive to \textit{unseen} visual distractions: even minor shifts in backgrounds, lighting, or camera position can cause substantial policy degradation.

\begin{figure}[t!]
\centering
\includegraphics[width=0.80\textwidth]{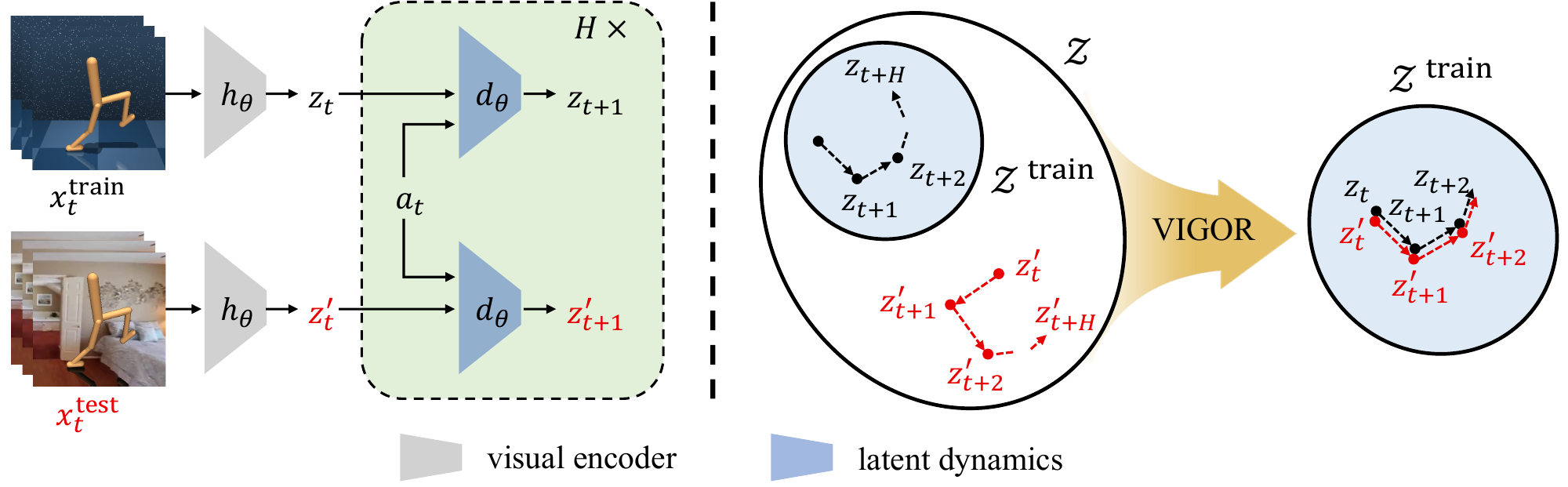}
\captionsetup{font=small}
\caption{\textbf{Challenges of unseen visual distractions.} Left: During training (top), clean observations $x_t^{\text{train}}$ are encoded into in-distribution latents $z_t \in \mathcal{Z}^{\text{train}}$, where the dynamics model $d_\theta$ produces accurate transitions. At test time (bottom), visually distracted observations $x_t^{\text{test}}$ yield out-of-distribution latents $z'_t \notin \mathcal{Z}^{\text{train}}$, causing erroneous transitions. Right: These deviations compound over horizon $H$, leading to divergent rollouts and planning failures. The proposed VIGOR framework targets this failure mode by aligning perturbed rollouts $(z'_t, z'_{t+1}, \dots, z'_{t+H})$ with their in-distribution counterparts $(z_t, z_{t+1}, \dots, z_{t+H})$, anchoring predictions to the task-relevant manifold $\mathcal{Z}^{\text{train}}$ (mechanism described in Section~\ref{sec:VIGOR}).}
\label{fig:distribution_shift}
\end{figure}

In model-free RL (MFRL), visual generalization has been addressed through data augmentation~\citep{hansen2021stabilizing,huang2022spectrum}, robust representation learning~\citep{hansen2021generalization,bertoin2022look,yuan2022pre}, and stabilized value-function training~\citep{liu2023improving,huang2024learning}. Model-based RL (MBRL) offers superior sample efficiency by planning in learned latent dynamics~\citep{hansen2023td,hafner2023mastering}, but achieving visual generalization in MBRL is \emph{structurally} more challenging. In MFRL, encoder perturbations affect only a single-step prediction. In MBRL, the problem is two-level: visual distractions first push encoder outputs out of the training distribution, and these out-of-distribution (OOD) latents are then recursively propagated by the dynamics model over a planning horizon~$H$, causing errors to compound across steps (Figure~\ref{fig:distribution_shift}). This two-level vulnerability, which does not arise in MFRL, means that techniques effective for single-step robustness are insufficient when the encoder simultaneously serves a recursive dynamics model, Q-value networks, and a planner.

We propose \textbf{VI}sual \textbf{G}eneralization via latent-space c\textbf{O}nsistency in model-based \textbf{R}L~(\textbf{VIGOR}), a framework that enables zero-shot generalization while retaining the sample efficiency of its TD-MPC2 backbone. VIGOR addresses the two-level vulnerability through three interdependent components: (i)~\textit{asymmetric weak-to-strong augmentation} (AW), a batch-splitting strategy that produces paired weak-only and weak-to-strong latents; (ii)~\textit{dynamics-level consistency} (DC), which enforces augmentation-invariant transition predictions by anchoring dynamics outputs to stable targets; and (iii)~\textit{encoder-level stabilization} (ES), which prevents the shared encoder from drifting under the cross-augmentation pressure imposed by DC.

We evaluate VIGOR on the DeepMind Control Suite (DMC)~\citep{tassa2018deepmind} and Robosuite~\citep{zhu2020robosuite} under the RL-ViGen zero-shot protocol~\citep{yuan2024rl}, covering seven visual distraction types---a substantially broader evaluation than the two types (\textit{background-color}, \textit{background-video}) typically considered in prior work~\citep{hansen2021generalization,huang2022spectrum,bertoin2022look,ha2023dream} (Appendix~\ref{sec:appendix_distractions}). VIGOR outperforms state-of-the-art MFRL and MBRL baselines, improving zero-shot generalization over the second-best baseline by $3.4\%$ on DMC and $43.6\%$ on Robosuite while retaining the sample efficiency of its TD-MPC2 backbone. 
Ablation studies further reveal that naive transfer of MFRL strategies fails in MBRL: applying strong augmentation uniformly across the batch, without the asymmetric splitting of AW, achieves only $48\%$ of VIGOR's performance on DMC (Section~\ref{sec:core_components}). Furthermore, VIGOR's robustness is augmentation-agnostic: latent-space consistency, rather than any specific augmentation choice, drives generalization (Section~\ref{sec:design_choices}; Appendix~\ref{sec:appendix_augmentation_ablation}).

Our contributions are:
\begin{itemize}
  \vspace{-0.13cm}
  \item \textbf{Problem identification and analysis.} We identify a two-level failure mode in MBRL: visual distractions first produce OOD encoder outputs, which then compound through recursive rollouts---a failure mode unique to recursive latent dynamics (Figure~\ref{fig:distribution_shift}; Section~\ref{sec:embedding_vis}).
  \vspace{-0.03cm}
  \item \textbf{Integrated framework for MBRL robustness.} We propose VIGOR, which addresses this failure mode through three interdependent components (AW, DC, and ES) designed for the recursive world-model setting. We show that naive transfer of MFRL strategies fails in MBRL, that VIGOR's integration is necessary, and that the resulting robustness is augmentation-agnostic (Sections~\ref{sec:core_components}--\ref{sec:embedding_vis}; Appendix~\ref{sec:appendix_augmentation_ablation}). We further show that VIGOR also improves zero-shot generalization when instantiated on DreamerV3, a structurally distinct recurrent state-space model (RSSM)-based backbone (Appendix~\ref{sec:vigor_on_dreamerv3}).
  \vspace{-0.03cm}
  \item \textbf{State-of-the-art zero-shot generalization.} VIGOR outperforms MFRL and MBRL baselines, surpassing the second-best baseline by $3.4\%$ on DMC and $43.6\%$ on Robosuite while preserving the strong sample efficiency of its TD-MPC2 backbone.
\end{itemize}

\vspace{-0.10cm}
\section{Related Work} \label{sec:related_work}
\vspace{-0.13cm}

\subsection{Visual Model-Based Reinforcement Learning}
\vspace{-0.13cm}
Latent-space MBRL learns compact world models for planning or imagination in latent space~\citep{ha2018world,hafner2019learning,hafner2023mastering,hansen2022temporal,hansen2023td}. The Dreamer series adopts a \textit{decoder-based} architecture that reconstructs pixel observations~\citep{hafner2019dream,hafner2020mastering,hafner2023mastering}, whereas TD-MPC uses a \textit{decoder-free} architecture that plans without reconstruction~\citep{hansen2022temporal,hansen2023td}. Both families assume 
identical training and test distributions; in practice, unseen visual distractions push encoder outputs out of distribution, causing compounding rollout errors (Figure~\ref{fig:distribution_shift}). VIGOR equips latent-space MBRL with three interdependent components (AW, DC, and ES) designed to mitigate this failure mode. An extended discussion of MBRL architectures and our backbone selection rationale is provided in Appendix~\ref{sec:appendix_discussions}.

\vspace{-0.13cm}
\subsection{Visual Generalization in Model-Free Reinforcement Learning}
\vspace{-0.13cm}
MFRL approaches address visual robustness through data augmentation~\citep{hansen2021stabilizing,huang2022spectrum,song2024simple}, self-supervised representation learning~\citep{hansen2021generalization,bertoin2022look}, and pretrained visual encoders~\citep{yuan2022pre,nair2022r3m}. These methods operate at the \emph{single-step} level: the encoder maps observations directly to Q-values or policy outputs without recursive rollouts. In MBRL, by contrast, encoder perturbations propagate through multi-step latent rollouts and compound across the planning horizon, motivating VIGOR's two-level design. A detailed survey is provided in Appendix~\ref{sec:appendix_related_mfrl}.

\vspace{-0.13cm}
\subsection{Visual Generalization in Model-Based Reinforcement Learning}
\vspace{-0.13cm}
\textit{Invariant MBRL} methods~\citep{wang2022denoised,zhu2023repo,sun2024learning,zhou2025self} learn representations that discard task-irrelevant visual factors but typically rely on exposure to distractions during training or on \textit{test-time adaptation}. VIGOR targets the stricter \textit{zero-shot} setting where agents are trained on clean observations and no adaptation data are available. 
Dr.~G~\citep{ha2023dream} is the most directly comparable prior work, as it explicitly targets zero-shot generalization in latent-space MBRL. It extends Dreamer~\citep{hafner2019dream} with dual contrastive objectives and an inverse-dynamics auxiliary loss. VIGOR differs in mechanism: rather than contrastive negatives and auxiliary supervision, it enforces dynamics-level consistency~(DC) through direct latent regression, supported by asymmetric weak-to-strong augmentation~(AW) and encoder-level stabilization~(ES). Built on TD-MPC2~\citep{hansen2023td}, VIGOR targets the same zero-shot setting through a fundamentally different inductive bias. A detailed survey is provided in Appendix~\ref{sec:appendix_related_mbrl}.

\vspace{-0.10cm}
\section{VIGOR}\label{sec:VIGOR}
\vspace{-0.15cm}
VIGOR addresses the two-level vulnerability identified in Section~\ref{sec:introduction}: visual distractions first perturb encoder outputs, and these perturbations then compound through recursive latent rollouts. The framework comprises three interdependent components, each targeting a different level of the world-model pipeline: input augmentation, latent dynamics, and encoder representation. Figure~\ref{fig:VIGOR_architecture} provides an overview. The three components depend on each other: AW produces the paired latents on which DC operates; DC stabilizes dynamics but can cause encoder drift; ES counteracts this drift. Because AW modifies only the input pipeline while DC and ES act through auxiliary losses, VIGOR can be integrated into latent-space MBRL frameworks without modifying the underlying planner or optimization procedure. Throughout the main paper, we instantiate VIGOR on TD-MPC2~\citep{hansen2023td}; a second instantiation on DreamerV3~\citep{hafner2023mastering} is described in Appendix~\ref{sec:vigor_on_dreamerv3}.

\vspace{-0.13cm}
\subsection{Asymmetric Weak-to-Strong Augmentation (AW)} \label{sec:weak_strong}
\vspace{-0.13cm}
We distinguish between \textit{weak} augmentations (minor perturbations that preserve most visual information) and \textit{strong} augmentations that impose substantial visual changes. We refer to the sequential composition $\tau^s \circ \tau^w$ as \textit{weak-to-strong} augmentation, emphasizing application order rather than a progressive curriculum. We adopt \textit{random-shift}~\citep{yarats2021mastering} as $\tau^w$ and \textit{random-overlay}~\citep{hansen2021generalization} as $\tau^s$.

\begin{figure}[t!]
\centering
\includegraphics[width=0.95\textwidth]{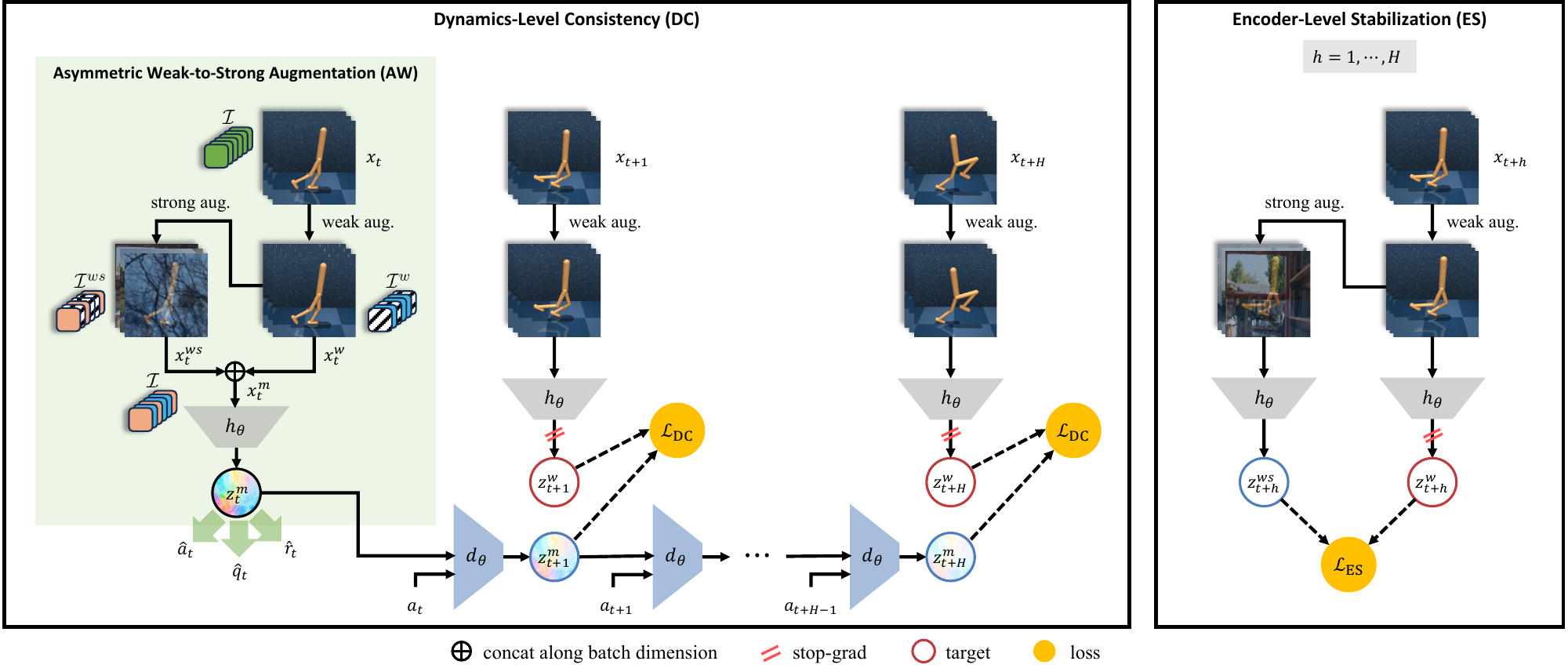}
\captionsetup{font=small}
\caption{\textbf{Overview of VIGOR.} Left: AW partitions a mini-batch into weak-only ($\mathcal{I}^{w}$) and weak-to-strong ($\mathcal{I}^{ws}$) subsets, producing paired views $x_t^w$ and $x_t^{ws}$ that are concatenated along the batch dimension as $x_t^m = x_t^w \oplus x_t^{ws}$ and encoded by the shared encoder $h_\theta$ into the mixed representation $z_t^m$. DC rolls $z_t^m$ forward through the dynamics model over horizon $H$ and aligns each prediction with the weakly augmented target $z_{t+h}^w$ for $h = 1, \dots, H$, enforcing augmentation-invariant transition predictions. Right: ES aligns weak-only and weak-to-strong encodings of the same frame, preventing encoder drift under DC's cross-augmentation supervision.}
\label{fig:VIGOR_architecture}
\end{figure}

AW produces paired \textit{weak-only} and \textit{weak-to-strong} samples within a single mini-batch, providing the structural foundation for DC while preserving the original batch size. As illustrated in Figure~\ref{fig:VIGOR_architecture}, observations are processed through two augmentation branches: the \textit{weak-only} branch applies $\tau^w$ alone, while the \textit{weak-to-strong} branch applies $\tau^s \circ \tau^w$. The two branches are then concatenated along the batch dimension and jointly encoded by the shared encoder $h_\theta$ into the mixed representation $z_t^m$, which is rolled forward by the dynamics model.

\textbf{Batch split.}\quad
Given a mini-batch of $N$ samples with index set $\mathcal{I} = \{1, \dots, N\}$, we partition $\mathcal{I}$ uniformly at random into two disjoint subsets:
\begin{align} 
\mathcal{I} = \mathcal{I}^w \cup \mathcal{I}^{ws}, \quad 
\mathcal{I}^w \cap \mathcal{I}^{ws} = \emptyset, \quad 
\zeta = |\mathcal{I}^w| / |\mathcal{I}^{ws}|. 
\end{align} 
Samples in $\mathcal{I}^w$ are routed through the weak-only branch, while samples in $\mathcal{I}^{ws}$ are routed through the weak-to-strong branch. We set $\zeta = 1$ throughout our experiments, splitting the batch evenly.

\textbf{Augmented representations.}\quad
At the first time step of each trajectory segment of horizon $H$, we construct weak-only and weak-to-strong observation stacks for the two subsets:
\begin{align}
  x_t^w = \tau^w\bigl(\{x_{t,n}\}_{n\in\mathcal{I}^w}, \upsilon^w \bigr), \quad
  x_t^{ws} = \tau^s\bigl( \tau^w\bigl( \{x_{t,n}\}_{n\in\mathcal{I}^{ws}}, \upsilon^w \bigr), \upsilon^s \bigr), 
\end{align}
where $x_{t,n}$ denotes the observation of the $n$-th sample at time $t$, and $\upsilon^w \sim \Upsilon^w$ and $\upsilon^s \sim \Upsilon^s$ parameterize the respective augmentations. The mixed observation $x_t^m = x_t^w \oplus x_t^{ws}$, where $\oplus$ denotes batch-dimension concatenation, is encoded into $z_t^m = h_\theta(x_t^m)$ in a single forward pass. The mixed representation $z_t^m$ initializes the latent rollout, whereas weakly augmented views of the subsequent observations $x_{t+1:t+H}$ provide the DC targets and, paired with their weak-to-strong counterparts, the ES pairs (Sections~\ref{sec:latent_consistency}--\ref{sec:enc_reg}). Further analyses of AW are provided in Appendix~\ref{sec:appendix_augmentation_ablation}.

\vspace{-0.13cm}
\subsection{Dynamics-Level Consistency (DC)}\label{sec:latent_consistency}
\vspace{-0.13cm}
The key insight is that task-relevant dynamics should remain invariant to visual distractions: the world model should produce consistent next-state predictions regardless of the visual perturbation applied to its input. The DC loss enforces this invariance at the dynamics level, a constraint specific to latent-space MBRL, where the encoder output feeds a recursive transition model.

For reference, the standard dynamics loss in latent-space MBRL~\citep{hansen2023td} is:
\begin{align}
  \mathcal{L}_{\text{dyn}}(\theta)
  = \bigl\lVert d_\theta(z_t, a_t)
    - \operatorname{sg}\bigl(h_\theta(x_{t+1})\bigr)
    \bigr\rVert_2^{2},
  \label{eq:dyn_loss}
\end{align}
where $z_t = h_\theta(x_t)$ is the encoded observation, $d_\theta$ is the dynamics model, and $\operatorname{sg}(\cdot)$ denotes the stop-gradient operator.

DC modifies this objective in two ways: at the first time step of each trajectory segment, the dynamics model receives the mixed representation $z_t^m$ from AW; the target is the encoding of a \emph{weakly augmented} successor. The single-step DC loss is:
\begin{align}
  \mathcal{L}_{\text{DC}}(\theta)
  = \bigl\lVert d_\theta(\hat{z}_t^m, a_t)
    - \operatorname{sg}(z_{t+1}^w)
    \bigr\rVert_2^{2},
  \quad \text{where } z_{t+1}^w = h_\theta\bigl(\tau^w(x_{t+1}, \upsilon^w)\bigr).
  \label{eq:dc_loss}
\end{align}
The mixed representation $z_t^m$ is injected only at the initial step of each trajectory segment of horizon $H$ ($\hat{z}_t^m = z_t^m$), after which the dynamics model rolls forward recursively: $\hat{z}_{t+h}^m = d_\theta(\hat{z}_{t+h-1}^m, a_{t+h-1})$ for $h = 1, \dots, H$. At every step, the rollout prediction $\hat{z}_{t+h}^m$ is aligned with the weakly augmented target $z_{t+h}^w$, propagating the initial perturbation through the entire horizon while requiring augmentation-invariant transitions.

The weak target $z_{t+h}^w$ provides a stable supervisory signal. Applying $\tau^w$ universally, rather than maintaining a clean path, is a deliberate choice: consistent with prior work on visual RL~\citep{yarats2021image,yarats2021mastering}, weak augmentation improves learning and provides a stable yet slightly diversified anchor for the subsequent objectives. The model must produce consistent predictions from the perturbed mixed input, thereby enforcing \emph{augmentation-invariant} latent dynamics.

\vspace{-0.13cm}
\subsection{Encoder-Level Stabilization (ES)}\label{sec:enc_reg}
\vspace{-0.13cm}
While DC enforces invariance at the dynamics level, the encoder may still produce inconsistent representations across augmentations. In VIGOR, a single shared encoder feeds the dynamics model, Q-value networks, and model predictive control (MPC) planner. The cross-augmentation supervision from DC can distort this shared representation, causing encoder drift that propagates through multi-step rollouts. ES adapts the alignment objective of SODA~\citep{hansen2021generalization}, originally proposed for MFRL, to a different role in VIGOR: counteracting the encoder drift induced by DC's cross-augmentation supervision, a failure mode specific to MBRL, where the encoder feeds recursive latent rollouts.

At each time step, the ES loss minimizes the $\ell_2$ distance between $\ell_2$-normalized weak-only and weak-to-strong encodings of the same frame:
\begin{align}
  \mathcal{L}_{\text{ES}}(\theta)=
  \Bigl\|
    \frac{z_t^{ws}}{\lVert z_t^{ws}\rVert_2}
    -
    \frac{\operatorname{sg}(z_t^{w})}
         {\lVert\operatorname{sg}(z_t^{w})\rVert_2}
  \Bigr\|_2^{2}.
  \label{eq:es_loss}
\end{align}
ES thus anchors weak-to-strong encodings to their weak-only counterparts, keeping the shared representation stable while DC constrains the dynamics. A detailed discussion of the structural role of ES is provided in Appendix~\ref{sec:appendix_impl_method}.

\vspace{-0.13cm}
\subsection{Overall Training Objective}\label{sec:objective}
\vspace{-0.13cm}
VIGOR modifies the TD-MPC2~\citep{hansen2023td} learning objective by replacing its latent consistency loss (Eq.~\eqref{eq:dyn_loss}) with $\mathcal{L}_{\text{DC}}$ and adding $\mathcal{L}_{\text{ES}}$. For a trajectory segment $\Gamma = (x_i, a_i, r_i, x_{i+1})_{i=t}^{t+H}$ sampled from replay buffer $\mathcal{B}$, the overall objective is:
\begin{equation}
\label{eq:VIGOR_obj}
\resizebox{0.93\textwidth}{!}{$
\mathcal{L}_{\text{total}}(\theta)=
\mathbb{E}_{\Gamma\sim\mathcal{B}}
\Biggl[
\textcolor{darkred}{\mathbb{E}_{\upsilon^{w}\sim\Upsilon^w,\upsilon^{s}\sim\Upsilon^s}}
\Bigl[
\sum_{i=t}^{t+H}\!
\lambda^{\,i-t}
\underbrace{
\underbrace{\mathcal{L}_{\text{TD-MPC2}}
(\theta;\mathcal{L}_{\text{rew}},\mathcal{L}_{Q},
\textcolor{darkred}{\mathcal{L}_{\text{DC}}},\Gamma_i)}_{\text{world-model losses}}
\;+\;
\textcolor{darkred}{\mathcal{L}_{\text{ES}}(\theta;\upsilon^{w},\upsilon^{s})}}_{\textcolor{darkred}{\mathcal{L}_{\text{VIGOR}}(\theta;\upsilon^{w},\upsilon^{s})}}
\Bigr]
\Biggr],
$}
\end{equation}
where $\Gamma_i = (x_i, a_i, r_i, x_{i+1})$ denotes the per-step transition tuple within the segment, and $\mathcal{L}_{\text{TD-MPC2}}$ denotes the base world-model objective comprising the reward prediction loss ($\mathcal{L}_{\text{rew}}$) and the Q-value loss ($\mathcal{L}_Q$), with the DC loss ($\mathcal{L}_{\text{DC}}$) in place of the standard latent consistency loss. The temporal coefficient $\lambda \in (0, 1]$ downweights losses at later horizon steps. Here, $\mathcal{L}_{\text{VIGOR}}(\theta;\upsilon^{w},\upsilon^{s})$ denotes the per-step VIGOR loss. The terms highlighted in \textcolor{darkred}{red} are the only additions to the original TD-MPC2 objective. The training procedure is summarized in Algorithm~\ref{algo:VIGOR} (Appendix~\ref{app:alg}); implementation details and hyperparameters are given in Appendix~\ref{sec:appendix_implementation_details}; and a detailed discussion of each component appears in Appendix~\ref{sec:appendix_impl_method}.

\vspace{-0.10cm}
\section{Experiments}\label{sec:experiments}
\vspace{-0.15cm}
We evaluate VIGOR's zero-shot generalization and sample efficiency against state-of-the-art MFRL and MBRL methods on six continuous control tasks from DMC~\citep{tassa2018deepmind} and four manipulation tasks from Robosuite~\citep{zhu2020robosuite}. All agents are trained in clean environments and evaluated \textit{zero-shot} under unseen visual distractions (Figure~\ref{fig:train_eval}; representative evaluation scenes are catalogued in Appendix~\ref{sec:appendix_eval_scenes}). Our experiments address the following questions:
\begin{itemize}
    \vspace{-0.10cm}
    \item[\textbf{Q1.}] Does VIGOR achieve superior zero-shot generalization while preserving sample efficiency?
    \item[\textbf{Q2.}] How does AW contribute to the efficiency--robustness trade-off, and why is its batch-splitting design necessary for MBRL?
    \item[\textbf{Q3.}] How do DC and ES individually contribute, and how do they interact?
    \item[\textbf{Q4.}] How do specific design choices (augmentation scheduling, augmentation family, and auxiliary objective) affect generalization?
    \item[\textbf{Q5.}] Does VIGOR preserve latent-space consistency at both the encoder and dynamics levels under unseen visual shifts?
\end{itemize}

\begin{figure}[t!]
\centering
\includegraphics[width=0.85\columnwidth]{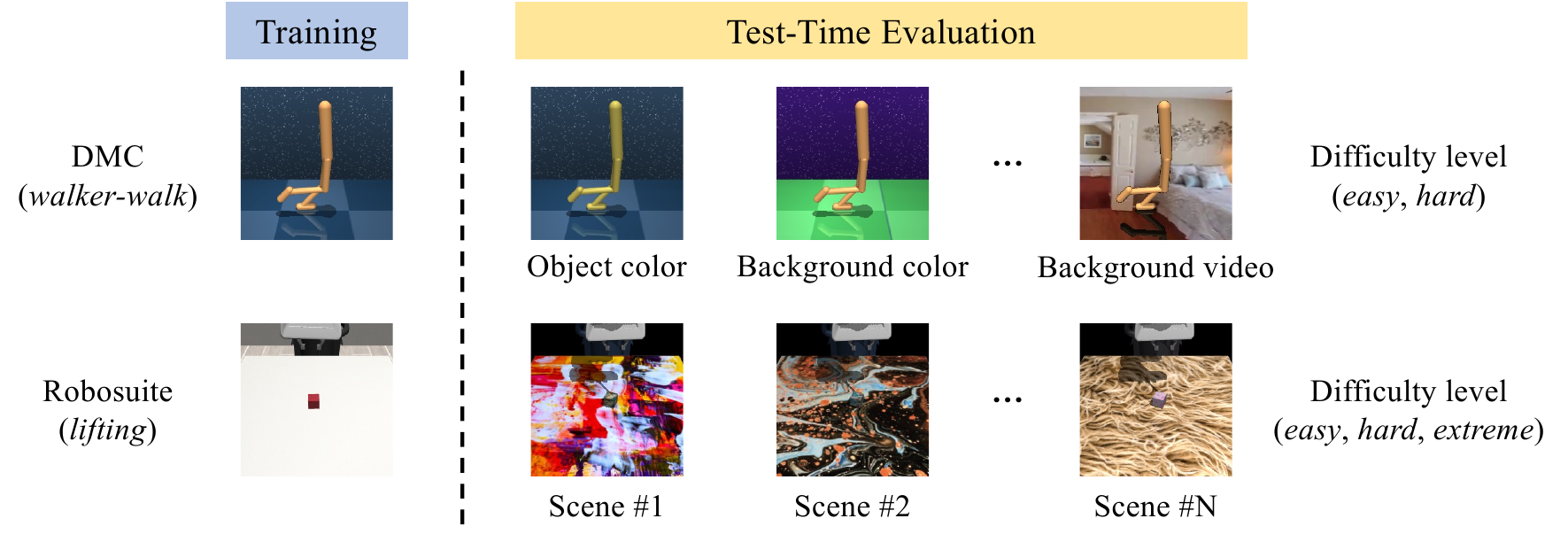}
\captionsetup{font=small}
\caption{\textbf{Illustrative training--evaluation setup.} Agents are trained on clean observations (leftmost column) and evaluated \textit{zero-shot} under unseen visual distractions. Shown are representative examples from DMC (top, \textit{walker-walk}) and Robosuite (bottom, \textit{lifting}) across diverse distraction types, scenes, and difficulty levels.}
\label{fig:train_eval}
\end{figure}

\vspace{-0.13cm}
\subsection{Experimental Setup}\label{sec:exp_setup}
\vspace{-0.13cm}
\textbf{Environments.}\quad
We benchmark VIGOR on six DMC tasks (\textit{walker-walk}, \textit{finger-spin}, \textit{cheetah-run}, \textit{walker-stand}, \textit{reacher-easy}, and \textit{cartpole-swingup}) and four Robosuite manipulation tasks (\textit{door-opening}, \textit{nut-assembly}, \textit{peg-in-hole}, and \textit{lifting}). Details are provided in Appendix~\ref{sec:appendix_evaluation_details}.

\textbf{Baselines.}\quad
We compare VIGOR against eight baselines spanning both paradigms: four MFRL methods (DrQ-v2~\citep{yarats2021mastering}, SVEA~\citep{hansen2021stabilizing}, SRM~\citep{huang2022spectrum}, SimGRL~\citep{song2024simple}) and four MBRL methods (TD-MPC2~\citep{hansen2023td}, DreamerV3~\citep{hafner2023mastering}, HRSSM~\citep{sun2024learning}, Dr.~G~\citep{ha2023dream}). Baseline descriptions are provided in Appendix~\ref{sec:appendix_baselines}.

\textbf{Distractions.}\quad
We follow the RL-ViGen zero-shot protocol~\citep{yuan2024rl} and evaluate all methods under seven visual distraction types, substantially broader than the two types (\textit{background-color}, \textit{background-video}) used in most prior work~\citep{hansen2021generalization,huang2022spectrum,bertoin2022look,ha2023dream}. Details are provided in Appendix~\ref{sec:appendix_distractions}.

\begin{table*}[t!]
\captionsetup{font=small}
\caption{\textbf{Zero-shot generalization scores.} Mean and interquartile mean (IQM) of episode returns aggregated across all tasks, distraction types, and difficulty levels. Values in brackets denote 95\% stratified bootstrap confidence intervals. Cyan-shaded cells with boldface indicate the best score per row.}
\label{tab:exp_gen}
\vspace{-0.15cm}
\centering
\begin{small}
\begin{sc}
\renewcommand{\arraystretch}{1.5}
\resizebox{1.0\textwidth}{!}{
\begin{tabular}{ll|cccc|cccc|c}
\toprule
 & & \multicolumn{4}{c|}{MFRL} & \multicolumn{5}{c}{MBRL} \\
\cmidrule(lr){3-6}\cmidrule(lr){7-11}
Env & Metric & DrQ-v2 & SVEA & SRM & SimGRL & TD-MPC2 & DreamerV3 & HRSSM & Dr.~G & \textbf{VIGOR~(Ours)} \\
\midrule
\multirow[c]{3.5}{*}{DMC}
 & \multirow[c]{1.7}{*}{Mean} & $626.1$ & $726.2$ & $752.0$ & $710.2$ & $599.1$ & $635.3$ & $663.1$ & $623.7$ & \cellcolor{cyan!20}{$\mathbf{777.6}$} \\[-4pt]
 & & {\scriptsize $[618.7, 633.6]$} & {\scriptsize $[721.2, 731.2]$} & {\scriptsize $[747.2, 756.7]$} & {\scriptsize $[706.4, 713.8]$} & {\scriptsize $[592.2, 606.1]$} & {\scriptsize $[629.3, 641.3]$} & {\scriptsize $[657.8, 668.3]$} & {\scriptsize $[616.9, 630.4]$} & \cellcolor{cyan!20}{\scriptsize $\mathbf{[773.6, 781.5]}$} \\
 & \multirow[c]{1.7}{*}{IQM} & $739.5$ & $815.5$ & $834.9$ & $807.9$ & $672.2$ & $738.3$ & $749.1$ & $704.8$ & \cellcolor{cyan!20}{$\mathbf{856.9}$} \\[-4pt]
 & & {\scriptsize $[727.1, 751.6]$} & {\scriptsize $[811.6, 819.4]$} & {\scriptsize $[831.3, 838.5]$} & {\scriptsize $[805.5, 810.2]$} & {\scriptsize $[661.1, 683.3]$} & {\scriptsize $[728.6, 747.6]$} & {\scriptsize $[742.3, 755.9]$} & {\scriptsize $[694.3, 714.9]$} & \cellcolor{cyan!20}{\scriptsize $\mathbf{[853.2, 860.6]}$} \\
\midrule
\multirow[c]{3.6}{*}{Robosuite}
 & \multirow[c]{1.7}{*}{Mean} & $41.0$ & $82.8$ & $91.8$ & $40.3$ & $40.2$ & $64.3$ & $45.1$ & $54.0$ & \cellcolor{cyan!20}{$\mathbf{131.8}$} \\[-4pt]
 & & {\scriptsize $[39.3, 42.8]$} & {\scriptsize $[78.7, 87.0]$} & {\scriptsize $[87.2, 96.6]$} & {\scriptsize $[38.3, 42.6]$} & {\scriptsize $[39.2, 41.2]$} & {\scriptsize $[62.5, 66.3]$} & {\scriptsize $[44.3, 45.9]$} & {\scriptsize $[52.0, 56.3]$} & \cellcolor{cyan!20}{\scriptsize $\mathbf{[124.9, 138.8]}$} \\
 & \multirow[c]{1.7}{*}{IQM} & $15.7$ & $38.7$ & $50.5$ & $7.2$ & $1.7$ & $30.1$ & $6.2$ & $3.6$ & \cellcolor{cyan!20}{$\mathbf{58.7}$} \\[-4pt]
 & & {\scriptsize $[14.2, 17.2]$} & {\scriptsize $[33.9, 43.9]$} & {\scriptsize $[45.5, 55.6]$} & {\scriptsize $[5.5, 9.1]$} & {\scriptsize $[1.4, 2.0]$} & {\scriptsize $[27.2, 33.0]$} & {\scriptsize $[5.7, 6.8]$} & {\scriptsize $[2.4, 5.1]$} & \cellcolor{cyan!20}{\scriptsize $\mathbf{[50.8, 67.2]}$} \\
\bottomrule
\end{tabular}
}
\vspace{-0.10cm}
\end{sc}
\end{small}
\end{table*}

\vspace{-0.13cm}
\subsection{Experimental Results}\label{sec:results}
\vspace{-0.13cm}
To address \textbf{(Q1)}, we evaluate VIGOR along two axes: \textit{zero-shot generalization} and \textit{sample efficiency}. All results are averaged over ten random seeds.

\textbf{Zero-shot generalization.}\quad
All agents are trained in a clean setting and evaluated \textit{zero-shot} under unseen visual distractions across multiple difficulty levels (\textit{easy} and \textit{hard} in DMC; \textit{easy}, \textit{hard}, and \textit{extreme} in Robosuite). Table~\ref{tab:exp_gen} reports mean and interquartile mean (IQM)~\citep{agarwal2021deep} aggregated across all tasks, distraction types, and difficulty levels; detailed breakdowns are provided in Appendix~\ref{sec:appendix_zsg}.

VIGOR achieves the strongest zero-shot generalization on both benchmarks. On DMC, it leads on both metrics (mean: $777.6$; IQM: $856.9$), surpassing the second-best baseline (SRM) by $3.4\%$ in mean and $2.6\%$ in IQM. The gap widens substantially on Robosuite, where most baselines degrade under severe distribution shift: VIGOR reaches mean $131.8$ and IQM $58.7$, exceeding the second-best baseline (SRM) by $43.6\%$ in mean and $16.2\%$ in IQM. Notably, our backbone TD-MPC2 ranks lowest among all baselines on both benchmarks and both metrics, indicating that latent-space MBRL without explicit consistency mechanisms is particularly fragile under visual shifts. VIGOR's strong performance suggests that jointly enforcing encoder-level stabilization and dynamics-level consistency enables structured rollouts under unseen distractions.

\begin{figure}[t!]
\centering
\includegraphics[width=0.90\columnwidth]{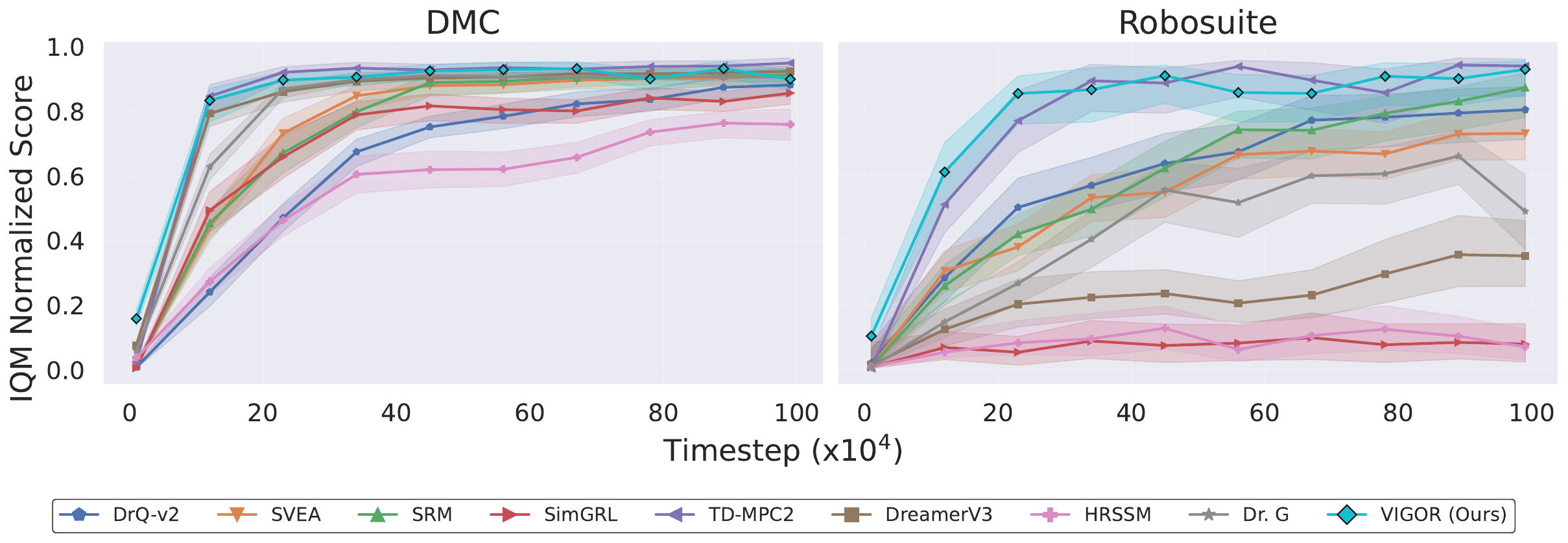}
\captionsetup{font=small}
\caption{\textbf{Sample efficiency.} Learning curves on DMC and Robosuite, reported as IQM normalized scores across methods. Shaded regions denote 95\% stratified bootstrap confidence intervals.}
\vspace{-0.14cm}
\label{fig:sample_efficiency_curves}
\end{figure}

\textbf{Sample efficiency.}\quad
Figure~\ref{fig:sample_efficiency_curves} reports aggregated IQM normalized scores; per-task curves and statistical analyses are provided in Appendix~\ref{sec:appendix_se}. On DMC, VIGOR closely matches TD-MPC2's rapid convergence while requiring fewer samples than all other baselines. On Robosuite, the efficiency gap widens further: TD-MPC2 and VIGOR converge at comparable rates, while other baselines fall behind in both speed and asymptotic returns. These results confirm that VIGOR preserves the strong sample efficiency of its TD-MPC2 backbone while substantially improving robustness---a combination that prior methods achieve only in part. We corroborate this under a tighter budget on DMC 100K (Appendix~\ref{sec:appendix_dmc_100k}) and report wall-clock cost in Appendix~\ref{sec:rebuttal_wall-clock}.

\textbf{Beyond the TD-MPC2 backbone.}\quad
To examine whether these benefits are specific to TD-MPC2, we additionally instantiate VIGOR on DreamerV3, an RSSM-based world model with stochastic latents and actor--critic learning in imagination (Appendix~\ref{sec:vigor_on_dreamerv3}). Without backbone-specific tuning, DreamerV3+VIGOR raises the mean zero-shot return of DreamerV3 from $635.3$ to $767.8$ on DMC and from $64.3$ to $98.1$ on Robosuite, exceeding the mean return of every baseline in Table~\ref{tab:exp_gen} on both benchmarks, although it remains below the default TD-MPC2 instantiation. VIGOR's gains are thus not limited to the TD-MPC2 backbone.

\vspace{-0.13cm}
\subsection{Ablation Studies}\label{sec:core_components}
\vspace{-0.13cm}

We conduct ablation studies addressing \textbf{(Q2)} and \textbf{(Q3)}, isolating the contribution of each component to VIGOR's efficiency--robustness balance. Unless otherwise noted, the ablations below use two tasks per benchmark (\textit{cartpole-swingup} and \textit{reacher-easy}; \textit{door-opening} and \textit{peg-in-hole}).

\begin{figure}[t!]
\centering
\includegraphics[width=0.90\textwidth]{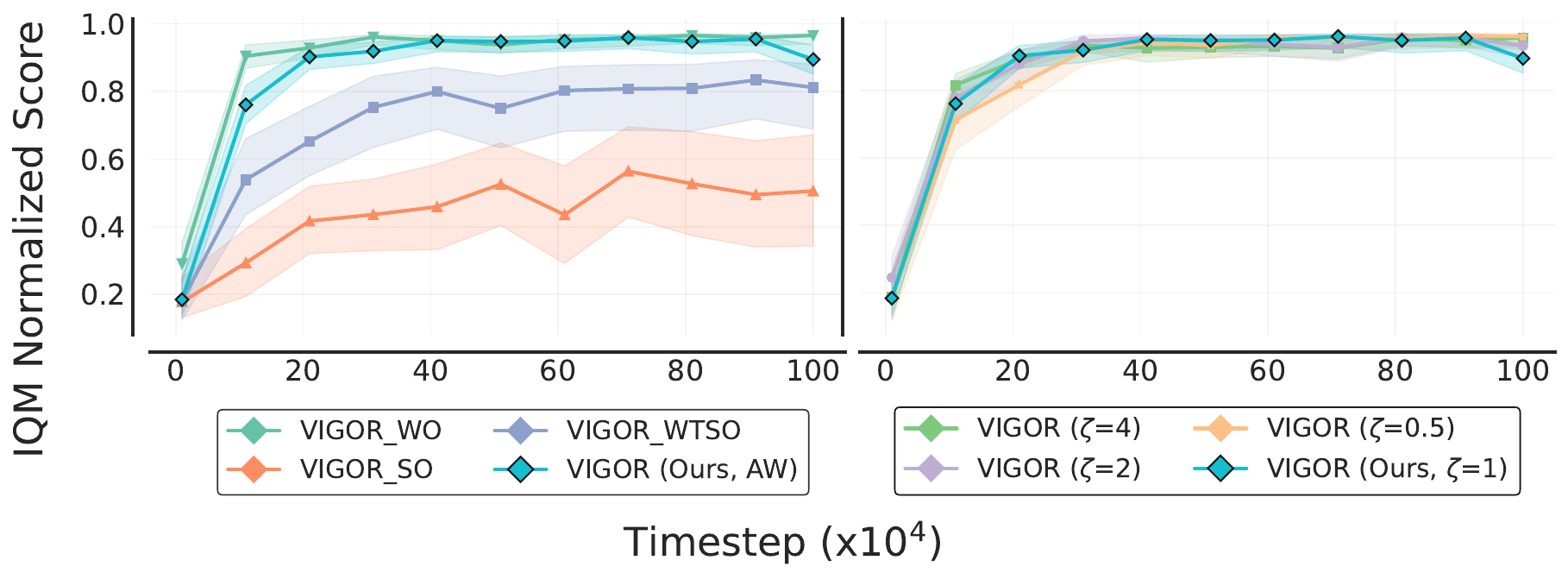}
\captionsetup{font=small}
\caption{\textbf{AW ablation: sample efficiency.} IQM normalized scores as a function of environment steps. \texttt{WO}:~weak-only; \texttt{SO}:~strong-only; \texttt{WTSO}:~weak-to-strong-only; $\zeta$: $|\mathcal{I}^w|/|\mathcal{I}^{ws}|$.}
\vspace{-0.10cm}
\label{fig:ablation_mixed_augmentations}
\end{figure}

\textbf{AW strategy (Q2).}\quad
We compare AW against three variants: weak-only (\texttt{WO}), strong-only (\texttt{SO}), and weak-to-strong-only (\texttt{WTSO}), which applies both transforms to every sample without batch splitting. We also sweep the mixture ratio $\zeta = |\mathcal{I}^w|/|\mathcal{I}^{ws}|$ to probe the sensitivity of the weak-only to weak-to-strong balance. Table~\ref{tab:ablation_mixed} and Figure~\ref{fig:ablation_mixed_augmentations} report zero-shot generalization and sample efficiency, respectively; further analyses are provided in Appendix~\ref{sec:appendix_augmentation_ablation}.

\begin{table}[t!]
\captionsetup{font=small}
\caption{\textbf{AW ablation: zero-shot generalization.} Mean episode returns ($\pm$ std) averaged across distraction types at each difficulty level. Percentages denote performance relative to VIGOR (AW, $\zeta{=}1$).}
\label{tab:ablation_mixed}
\centering
\begin{small}
\begin{sc}
\resizebox{1.00\textwidth}{!}{
\renewcommand{\arraystretch}{1.5}
\begin{tabular}{ll|ccccccc}
\toprule
Env & Difficulty & weak-only & strong-only & weak-to-strong-only & VIGOR $(\zeta{=}4)$ & VIGOR $(\zeta{=}2)$ & VIGOR $(\zeta{=}0.5)$ & \textbf{VIGOR (Ours; AW, $\zeta{=}1$)} \\
\midrule
\multirow{3.3}{*}{DMC}
& easy & $773.9 \pm 246.0$ & $423.8 \pm 421.9$ & $550.6 \pm 377.6$ & $854.4 \pm 176.5$ & $820.8 \pm 192.0$ & $780.9 \pm 235.2$ & \cellcolor{cyan!15}{$\mathbf{863.9 \pm 155.4}$} \\
& hard & $606.6 \pm 326.8$ & $374.9 \pm 401.0$ & $481.3 \pm 358.3$ & $736.3 \pm 274.3$ & $735.8 \pm 270.6$ & $683.0 \pm 302.1$ & \cellcolor{cyan!15}{$\mathbf{781.1 \pm 261.2}$} \\
\cmidrule(lr){2-9}
& Average & $690.2 \pm 301.0\ (83\%)$ & $399.4 \pm 412.2\ (48\%)$ & $516.0 \pm 369.6\ (62\%)$ & $795.3 \pm 238.0\ (96\%)$ & $778.3 \pm 238.4\ (94\%)$ & $731.9 \pm 275.1\ (88\%)$ & \cellcolor{cyan!30}{$\mathbf{822.5 \pm 218.9}$} \\
\midrule
\multirow{4.4}{*}{Robosuite}
& easy & $134.2 \pm 143.8$ & $187.1 \pm 182.2$ & $297.6 \pm 155.2$ & $276.2 \pm 195.0$ & $313.2 \pm 165.7$ & $329.6 \pm 187.3$ & \cellcolor{cyan!15}{$\mathbf{332.0 \pm 186.2}$} \\
& hard & $112.6 \pm 118.0$ & $137.3 \pm 137.6$ & $138.1 \pm 131.3$ & $151.8 \pm 156.4$ & $144.7 \pm 144.5$ & $172.2 \pm 157.4$ & \cellcolor{cyan!15}{$\mathbf{236.7 \pm 186.5}$} \\
& extreme & $93.4 \pm 96.9$ & $121.3 \pm 122.6$ & $130.1 \pm 120.7$ & $109.0 \pm 110.6$ & $117.1 \pm 111.7$ & $125.1 \pm 123.2$ & \cellcolor{cyan!15}{$\mathbf{146.0 \pm 143.0}$} \\
\cmidrule(lr){2-9}
& Average & $113.4 \pm 122.0\ (47\%)$ & $148.6 \pm 152.0\ (62\%)$ & $188.6 \pm 156.6\ (79\%)$ & $179.0 \pm 172.8\ (75\%)$ & $191.7 \pm 166.5\ (80\%)$ & $209.0 \pm 180.5\ (87\%)$ & \cellcolor{cyan!30}{$\mathbf{238.3 \pm 188.9}$} \\
\bottomrule
\end{tabular}}
\vspace{-0.10cm}
\end{sc}
\end{small}
\end{table}

The results reveal complementary failure modes that illuminate why AW's specific design is necessary for MBRL. 
On DMC, \texttt{SO} degrades to $48\%$ of VIGOR, confirming that applying strong augmentations to an entire batch triggers the two-level vulnerability: perturbed encoder outputs compound through recursive rollouts, destabilizing world-model learning. \texttt{WTSO} partially recovers ($62\%$) but remains far below AW, indicating that persistent strong exposure without a stable weak-only anchor disrupts optimization. \texttt{WO} converges fast (Figure~\ref{fig:ablation_mixed_augmentations},~left), but under-generalizes ($83\%$), as weak shifts alone are insufficient for harder distractions.
On Robosuite, where distribution shifts are more severe, the ranking inverts: \texttt{WO} becomes the weakest ($47\%$), \texttt{SO} improves relatively ($62\%$), and \texttt{WTSO} reaches $79\%$ yet still falls short of AW. AW is the only strategy that succeeds on both benchmarks, because its asymmetric splitting simultaneously provides optimization stability and perturbation diversity.

The $\zeta$-sweep reveals a complementary cross-benchmark pattern: on DMC, increasing the weak-only fraction improves performance ($\zeta{=}4$: $96\%$ $>$ $\zeta{=}2$: $94\%$ $>$ $\zeta{=}0.5$: $88\%$), while on Robosuite the trend reverses ($\zeta{=}0.5$: $87\%$ $>$ $\zeta{=}2$: $80\%$ $>$ $\zeta{=}4$: $75\%$). Only the balanced mixture ($\zeta{=}1$) achieves the best performance on both, confirming that this ratio balances stability and generalization coverage.

\textbf{DC and ES contributions (Q3).}\quad
Table~\ref{tab:ablation_components} isolates the effect of each loss. Removing DC reduces generalization to $89\%$ of full VIGOR on DMC and to $46\%$ on Robosuite; removing ES reduces it to $93\%$ and $88\%$, respectively. Removing both losses jointly reduces performance further ($76\%$ on DMC; $33\%$ on Robosuite).

The asymmetry between DC and ES is more pronounced under severe distribution shifts. On DMC, both components contribute modestly and similarly ($89\%$ vs.\ $93\%$). On Robosuite, the gap widens sharply: DC alone (w/o~ES) retains $88\%$, whereas ES alone (w/o~DC) retains only $46\%$. DC constrains the recursive prediction pathway: without it, the dynamics model produces inconsistent rollouts under visual shifts. ES alone cannot compensate, because encoder-level alignment does not prevent the dynamics model from amplifying residual discrepancies across multi-step rollouts. Conversely, the $12\%$ gap between DC alone and full VIGOR on Robosuite shows that ES's encoder-level alignment is also needed for DC's dynamics-level constraint to function effectively. The two losses are thus complementary: DC anchors transition predictions, while ES anchors the representations on which DC operates. We further visualize these failure modes in Appendix~\ref{sec:appendix_embedding_vis_ablation}.

\begin{table}[t!]
\captionsetup{font=small}
\caption{\textbf{Ablation on DC and ES.} Mean episode returns ($\pm$ std) averaged across distraction types at each difficulty level. Percentages denote performance relative to full VIGOR.}
\label{tab:ablation_components}
\centering
\begin{small}
\begin{sc}
\resizebox{0.75\columnwidth}{!}{
\renewcommand{\arraystretch}{1.5}
\begin{tabular}{ll|cccc}
\toprule
Env & Difficulty & \makecell[l]{w/o~DC \&~ES} & w/o~DC & w/o~ES & \textbf{VIGOR (Ours)} \\
\midrule
\multirow{3.3}{*}{\makecell[l]{DMC}}
& easy & $730.7 \pm 296.0$ & $816.4 \pm 232.2$ & $826.7 \pm 205.0$ & \cellcolor{cyan!15}{$\mathbf{863.9 \pm 155.4}$} \\
& hard & $522.3 \pm 357.2$ & $652.1 \pm 328.1$ & $713.8 \pm 300.5$ & \cellcolor{cyan!15}{$\mathbf{781.1 \pm 261.3}$} \\
\cmidrule(lr){2-6}
& Average & $626.5 \pm 344.2\ (76\%)$ & $734.3 \pm 295.8\ (89\%)$ & $770.2 \pm 263.3\ (93\%)$ & \cellcolor{cyan!30}{$\mathbf{822.5 \pm 218.9}$} \\
\midrule
\multirow{4.4}{*}{\makecell[l]{Robosuite}}
& easy & $74.2 \pm 78.6$ & $128.9 \pm 143.3$ & $311.8 \pm 184.9$ & \cellcolor{cyan!15}{$\mathbf{332.0 \pm 186.4}$} \\
& hard & $83.2 \pm 85.8$ & $108.5 \pm 113.5$ & $193.5 \pm 178.6$ & \cellcolor{cyan!15}{$\mathbf{236.7 \pm 186.7}$} \\
& extreme & $80.2 \pm 82.5$ & $95.9 \pm 99.9$ & $127.6 \pm 120.0$ & \cellcolor{cyan!15}{$\mathbf{146.0 \pm 143.2}$} \\
\cmidrule(lr){2-6}
& Average & $79.2 \pm 82.3\ (33\%)$ & $111.1 \pm 120.8\ (46\%)$ & $211.0 \pm 180.4\ (88\%)$ & \cellcolor{cyan!30}{$\mathbf{238.3 \pm 189.0}$} \\
\bottomrule
\end{tabular}}
\vspace{-0.10cm}
\end{sc}
\end{small}
\end{table}

\textbf{Summary.}\quad
VIGOR's gains stem from the interplay of all three components. AW provides the efficiency--robustness balance through asymmetric batch splitting; DC and ES are jointly necessary for reliable generalization, with DC playing the dominant role under severe distribution shifts by directly constraining the recursive dynamics pathway.

\begin{figure}[t!]
    \centering
    \includegraphics[width=0.75\columnwidth]{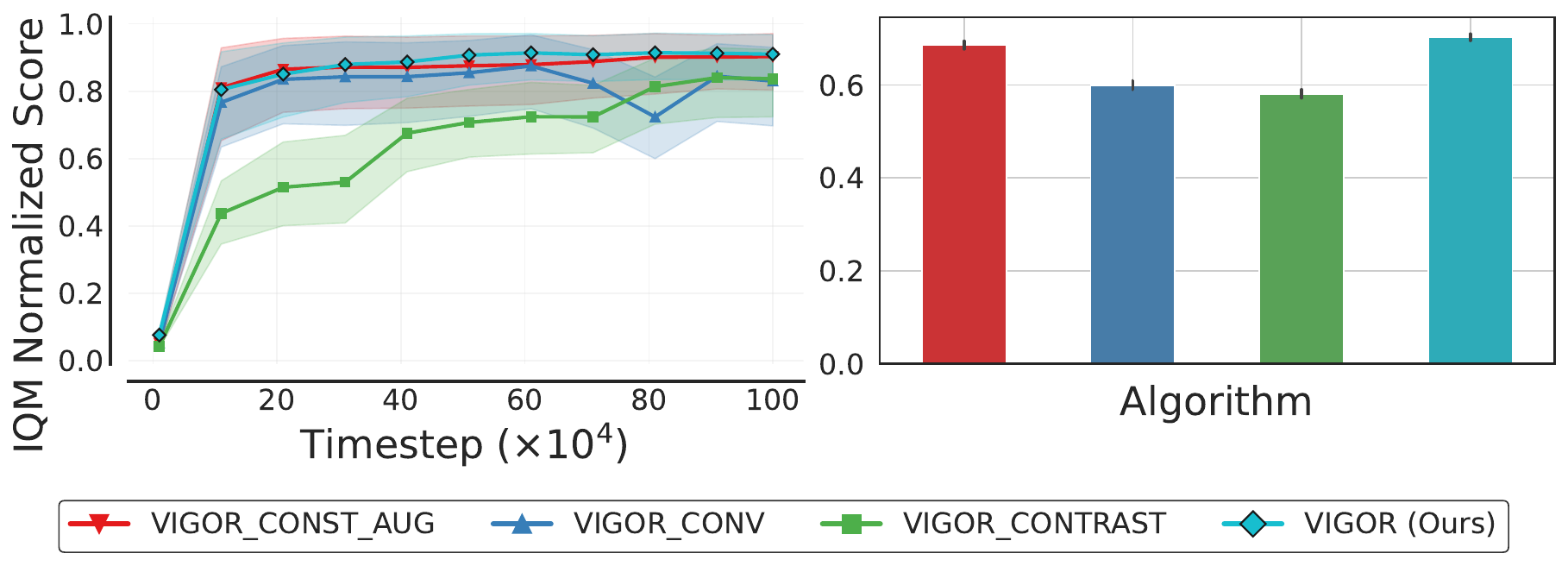}
    \captionsetup{font=small}
    \caption{\textbf{Ablation on design choices.} Left: Sample efficiency. Right: Zero-shot generalization.}
    \vspace{-0.10cm}
    \label{fig:design_choices}
\end{figure}

\vspace{-0.13cm}
\subsection{Ablation on Key Design Choices}\label{sec:design_choices}
\vspace{-0.13cm}
We analyze \textbf{(Q4)} by ablating three design choices, each isolated through a single-variable variant: \texttt{VIGOR\_CONST\_AUG} (fixed augmentation parameters across the horizon), \texttt{VIGOR\_CONV} (\textit{random-conv} replacing \textit{random-overlay}), and \texttt{VIGOR\_CONTRAST} (contrastive ES objective replacing the alignment-based one).
Figure~\ref{fig:design_choices} summarizes results on \textit{finger-spin}, \textit{cheetah-run}, and \textit{walker-walk}. VIGOR outperforms all variants on both axes, indicating that (i)~dynamic augmentation resampling is important for stable rollouts; (ii)~\textit{random-overlay} provides stronger robustness than convolution-based perturbations; and (iii)~alignment-based ES complements AW and DC more effectively than contrastive ES. Formal definitions and detailed results are provided in Appendix~\ref{sec:appendix_design_choices}.

\begin{figure}[t!]
\centering
\includegraphics[width=0.75\columnwidth]{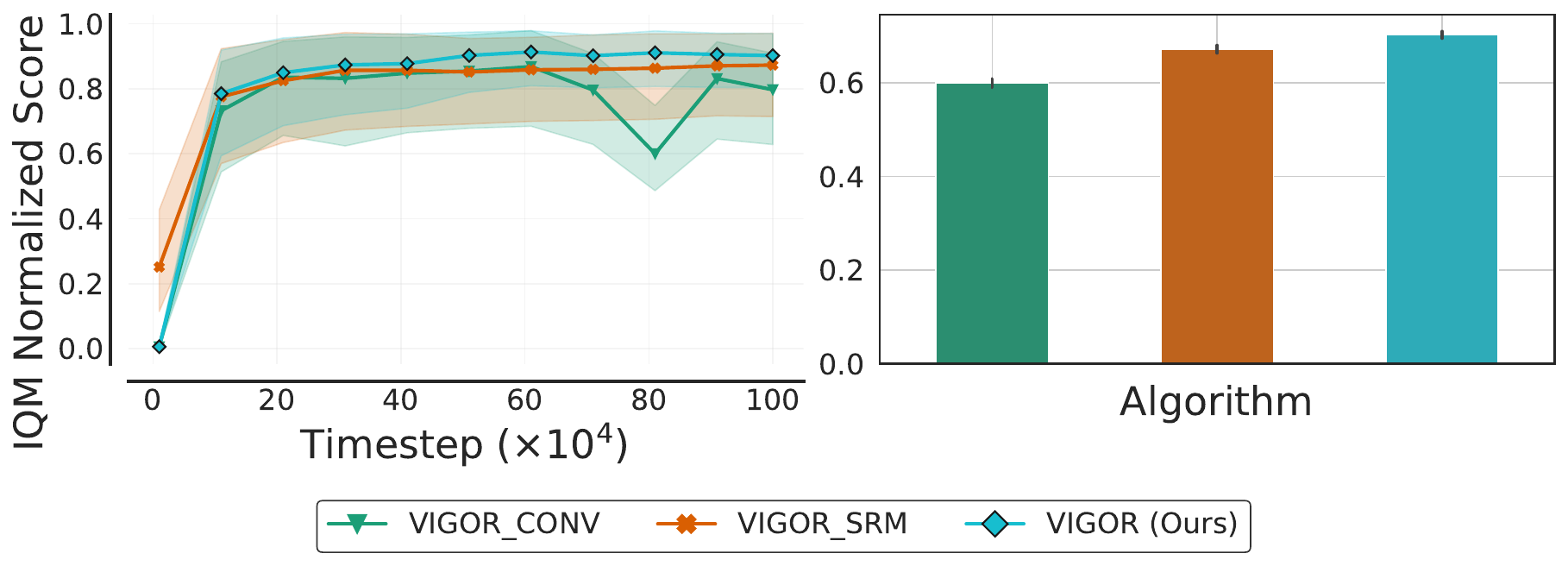}
\captionsetup{font=small}
\caption{\textbf{Augmentation-agnostic robustness.} Three VIGOR variants trained with three different strong augmentations (\textit{random-conv}, \textit{SRM}, and \textit{random-overlay}) all maintain strong zero-shot generalization.}
\vspace{-0.10cm}
\label{fig:aug_family_results}
\end{figure}

\textbf{Augmentation-agnostic robustness.}\quad 
A central question is whether VIGOR's robustness depends on the specific augmentation choice or on latent-space consistency. We replace \textit{random-overlay} with \textit{random-conv} (textural perturbations) and \textit{SRM} (spectral masking), two families that produce qualitatively distinct perturbations (Appendix~\ref{sec:appendix_augmentation_ablation}, Figure~\ref{fig:aug_family_diff}). As shown in Figure~\ref{fig:aug_family_results}, all three variants maintain strong zero-shot on the same three tasks, indicating that VIGOR's robustness is driven by latent-space consistency rather than by the choice of augmentation.

\vspace{-0.13cm}
\subsection{Latent-Space Consistency Analysis}\label{sec:embedding_vis}
\begin{figure}[t!]
    \centering
    \includegraphics[width=0.95\columnwidth]{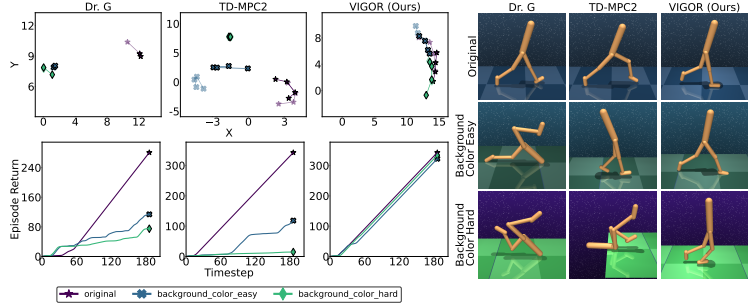}
    \captionsetup{font=small}
    \caption{\textbf{Latent-space consistency under visual perturbations.} Left: UMAP projections of latent rollouts (top) and episode returns (bottom) on \textit{walker-walk} with \textit{background-color} distractions. Right: Environment snapshots across evaluation settings. In the UMAP plots, markers denote difficulty levels ({\color[RGB]{100,50,140}$\bigstar$}:~\textit{original}, {\color[RGB]{40,80,215}$\bigtimes$}:~\textit{easy}, {\color[RGB]{34,180,55}$\Diamond$}:~\textit{hard}); darker trajectories show rollouts starting from $z_t$ ($z_{t:t+H}$), and lighter trajectories show rollouts starting from the preceding step $z_{t-1}$ ($z_{t-1:t+H-1}$), providing temporal context. VIGOR's latent rollouts across difficulty levels remain compact and overlap closely, while TD-MPC2~\citep{hansen2023td} and Dr.~G~\citep{ha2023dream} produce scattered initial encodings or divergent rollouts; the corresponding return curves are shown below each panel.}
    \vspace{-0.10cm}
    \label{fig:visualization_embeddings}
\end{figure}

\vspace{-0.13cm}
To address \textbf{(Q5)}, we compare VIGOR against TD-MPC2 (its backbone) and Dr.~G (the most directly comparable prior method) by rolling out each agent's world model from identical initial inputs and projecting the predicted latent states into 2D via UMAP~\citep{mcinnes2018umap} (Figure~\ref{fig:visualization_embeddings}). 

VIGOR maintains a compact latent manifold across difficulty levels, consistently aligning perturbed trajectories with their clean counterparts. This consistency holds at both levels of the vulnerability identified in Figure~\ref{fig:distribution_shift}: the initial encoder outputs under visual distractions remain close to their clean counterparts (encoder level), and this alignment is preserved throughout the rollout horizon (dynamics level). The resulting structural stability translates into stable returns. TD-MPC2 and Dr.~G exhibit scattered or divergent rollouts, leading to degraded performance, despite Dr.~G's contrastive objectives being designed for generalization. This comparison suggests that direct dynamics-level regression~(DC) based on asymmetric batch-splitting~(AW), combined with encoder stabilization~(ES), provides a more effective inductive bias for MBRL robustness than contrastive objectives with inverse-dynamics supervision, because it jointly constrains both levels of the prediction pathway that drives planning. Additional visualizations are provided in Appendix~\ref{sec:appendix_embedding_vis}; we further verify that this consistency holds under deeper planning horizons up to $H{=}10$ (Appendix~\ref{sec:rebuttal_horizons}).

\vspace{-0.15cm}
\section{Conclusion}\label{sec:conclusion}
\vspace{-0.20cm}
We introduced \textbf{VIGOR}, a framework that achieves robust zero-shot visual generalization in latent-space MBRL while preserving the strong sample efficiency of its TD-MPC2 backbone. VIGOR addresses the two-level vulnerability of latent-space MBRL: visual distractions perturb encoder outputs, and these perturbations compound through recursive rollouts. The framework integrates three interdependent components (AW, DC, and ES) to mitigate this failure mode. Across challenging OOD benchmarks, VIGOR surpasses the second-best baseline by $3.4\%$ on DMC and $43.6\%$ on Robosuite. Ablation studies confirm that all three components are jointly necessary and that VIGOR's robustness is augmentation-agnostic: latent-space consistency, not any particular augmentation choice, is what drives generalization. A second instantiation on DreamerV3 (Appendix~\ref{sec:vigor_on_dreamerv3}) also improves zero-shot generalization over its backbone on both benchmarks, suggesting that these benefits are not limited to the TD-MPC2 backbone.

\textbf{Limitations and future work.}\quad
VIGOR targets appearance-based perturbations and does not address geometric distortions; evaluations under unseen camera rotations (Appendix~\ref{sec:rebuttal_camera-viewpoints}) reveal that such shifts remain challenging, motivating geometry-aware representations as a primary future direction. Beyond visual shifts, extending VIGOR to dynamics- and task-level shifts~\citep{seo2020trajectory,beukman2023dynamics} is a natural next step, and validating it in real-robot deployments would assess its practical applicability.


\bibliographystyle{plainnat} 
\bibliography{reference}

\newpage
\appendix
\startcontents[app]
\setcounter{tocdepth}{2}
\clearpage
\phantomsection
\section*{Appendix Contents}
\printcontents[app]{l}{1}{\setlength{\parskip}{0pt}}
\newpage

\section{Impact Statement}\label{sec:impact}
This work presents VIGOR, an MBRL framework that achieves zero-shot visual generalization to unseen distractions while preserving sample efficiency. By integrating AW, DC, and ES, VIGOR addresses a critical bottleneck in deploying visual RL agents to environments with visual distribution shifts.

The primary impact is methodological: VIGOR provides a robust approach to stabilize both encoder representations and latent dynamics under distribution shifts. These advances contribute to the broader goal of developing RL systems that are less reliant on task-specific data collection and more resilient to the visual variability in real-world environments.

As our evaluations are conducted within simulated benchmarks (DMC and Robosuite), this work does not pose immediate direct societal or ethical risks. However, as these techniques move toward real-world deployment in robotics and autonomous systems, we emphasize that reliability and safety become paramount. Future applications in safety-critical domains should undergo rigorous validation and incorporate safeguards to prevent catastrophic decision errors arising from misaligned latent representations or unexpected environmental shifts.

\section{Extended Related Work}\label{sec:appendix_related_work} 
\subsection{Visual Generalization in MFRL}\label{sec:appendix_related_mfrl} 
This subsection provides a detailed survey of MFRL visual generalization methods, complementing the summary in Section~\ref{sec:related_work}. Baseline selection criteria are detailed in Appendix~\ref{sec:appendix_baselines}; methods not selected as baselines are discussed below for context.

\paragraph{Data-augmentation approaches.} 
Data augmentation has emerged as a simple yet effective strategy for visual robustness in MFRL. RAD~\citep{laskin2020reinforcement} systematically evaluates a suite of image transformations for RL, demonstrating that augmentations such as random crop and color jitter improve generalization. DrQ~\citep{yarats2021image} and DrQ-v2~\citep{yarats2021mastering} integrate random-shift augmentation into off-policy actor--critic training, achieving strong sample efficiency on pixel-based control. SVEA~\citep{hansen2021stabilizing} addresses the instability of augmented Q-learning by maintaining a weighted mixture of augmented and clean Q-targets, stabilizing training under diverse perturbations. SRM~\citep{huang2022spectrum} applies spectral perturbations in the frequency domain to improve robustness to spatial corruptions. Network Randomization~\citep{lee2020network} passes observations through a randomly initialized convolutional layer, providing an implicit form of style augmentation; this principle inspires our \textit{random-conv} ablation in Section~\ref{sec:design_choices}. SimGRL~\citep{song2024simple} tackles imbalanced saliency via feature-level frame stacking and mitigates observational overfitting via a shifted random-overlay augmentation, which is applied uniformly to every sample without batch partitioning. DG2A~\citep{lee2025diffusion} leverages a pretrained diffusion model (Stable Diffusion~\citep{rombach2022high}) for adaptive data augmentation, generating augmented images guided by saliency maps.

\paragraph{Representation-learning approaches.} 
A complementary direction aims to learn visual representations that are inherently robust to task-irrelevant variations. CURL~\citep{laskin2020curl} applies contrastive learning to augmented observations, encouraging the encoder to capture task-relevant features. SODA~\citep{hansen2021generalization} aligns the encoding of an augmented observation with that of its non-augmented counterpart through a soft alignment loss, promoting augmentation-invariant features without contrastive negatives. VIGOR's ES component adapts SODA's alignment objective to a structurally different role: in MFRL, alignment promotes augmentation-invariant features at the encoder, whereas in VIGOR it stabilizes the encoder against the cross-augmentation supervision induced by DC's dynamics-level training (Section~\ref{sec:enc_reg}). SGQN~\citep{bertoin2022look} combines saliency-guided attention masks with self-supervised objectives, encouraging the encoder to focus on action-relevant regions. SMG~\citep{zhang2024focus} exploits an auxiliary image reconstruction task via separated model branches for encoders and decoders, forcing the encoder to extract task-relevant representations from foreground pixels. ALDA~\citep{batra2025zero} disentangles the latent representation and projects OOD observations back to the training distribution using associative memory. ROUSER~\citep{yang2025learning} learns robust action-value representations based on the information bottleneck framework, leveraging one-step rewards and compressed reward representations.

\paragraph{Pretrained-encoder approaches.} 
Recent work leverages visual representations pretrained on large-scale datasets. PIE-G~\citep{yuan2022pre} uses frozen pretrained encoders to provide generalization-friendly features, bypassing the need for augmentation-based robustness. R3M~\citep{nair2022r3m} pretrains visual representations on egocentric video data using time-contrastive learning for downstream robotic manipulation. MoVie~\citep{yang2024movie} adapts pretrained vision models for RL via multi-view representation learning. These methods are orthogonal to VIGOR and could potentially be combined with our framework to further improve robustness.

\paragraph{Gradient- and optimization-based approaches.} 
CG2A~\citep{liu2023improving} identifies conflicting gradients between augmented views and aggregates them via conflict-aware gradient surgery, improving generalization stability. PAD~\citep{hansen2021deployment} uses a self-supervised auxiliary task at test time to adapt the encoder to new visual conditions, an approach that requires online adaptation and thus falls outside the zero-shot setting targeted by VIGOR.

\paragraph{Relationship to VIGOR.} 
All MFRL methods above address robustness at the single-step level: the encoder feeds into a Q-value or policy without recursive dynamics. As identified in Section~\ref{sec:introduction}, MBRL exhibits a two-level vulnerability: visual distractions first perturb encoder outputs, and these perturbations then compound through recursive latent rollouts. This structural difference means that augmentation or regularization applied to the encoder alone is insufficient: the dynamics model must also produce consistent predictions under distribution shifts. VIGOR addresses both levels: ES stabilizes the encoder (level one), while DC constrains the recursive dynamics pathway (level two), with AW providing the paired latent structure that enables both constraints. The closest MFRL counterpart, SimGRL, uses random-overlay augmentation but applies it uniformly to every sample without batch splitting, leaving both the dynamics-level vulnerability and cross-augmentation supervision unaddressed.

\subsection{Extended Discussion of MBRL Visual Generalization}\label{sec:appendix_related_mbrl} 

\paragraph{Invariant MBRL.} 
Several methods aim to learn distraction-invariant representations within the MBRL framework. DBC~\citep{zhang2021learning} uses bisimulation metrics to learn representations that group visually different but behaviorally equivalent states. Denoised MDP~\citep{wang2022denoised} identifies and removes noise variables from the latent representation. RePo~\citep{zhu2023repo} regularizes the encoder via a reconstruction-based objective that selectively preserves task-relevant information. HRSSM~\citep{sun2024learning} learns a bisimulation-based hybrid RSSM in DreamerV3~\citep{hafner2023mastering} to extract endogenous, task-relevant latent representations without pixel-level reconstruction. SD~\citep{kim2025make} uses segmentation masks as a structural prior for robust representation learning, building upon the DreamerV3 architecture. While these methods improve robustness, several require test-time adaptation or additional environment interactions in the target domain, limiting their applicability in the zero-shot setting targeted by VIGOR.

\paragraph{Dr.~G and its relationship to VIGOR.} 
Dr.~G~\citep{ha2023dream} is the most directly comparable prior work, as it explicitly targets zero-shot visual generalization in MBRL. Built on Dreamer~\citep{hafner2019dream}, Dr.~G applies weak and strong augmentations to create two independent views of every sample and trains dual contrastive losses (reality--reality at the encoder level and dream--reality at the dynamics level) alongside a recurrent inverse-dynamics objective.

VIGOR differs from Dr.~G in three key aspects: 
(i)~\textit{Efficiency}: AW partitions a single batch into weak-only and weak-to-strong subsets via asymmetric splitting, where the weak-only subset directly serves as the stable anchor for both DC's transition target and ES's alignment target. This anchor-and-augment structure requires only a single encoder pass per sample. Dr.~G, by contrast, generates two fully independent views per sample to construct contrastive pairs, doubling the encoder computation. 
(ii)~\textit{Mechanism}: Both methods address visual robustness in the latent-space MBRL pipeline, but through different means. Dr.~G uses contrastive losses requiring negative samples and an inverse-dynamics auxiliary loss. VIGOR enforces consistency through direct latent regression~(DC) based on asymmetric batch splitting~(AW), combined with encoder stabilization~(ES), without contrastive negatives or auxiliary supervision. 
(iii)~\textit{Backbone}: VIGOR builds on TD-MPC2's decoder-free, MPC-based architecture. Dr.~G builds on Dreamer's recurrent world model (RSSM) but deliberately omits the decoder, relying instead on contrastive and inverse-dynamics objectives to learn latent representations.

Our experiments show that VIGOR achieves stronger zero-shot generalization with higher sample efficiency (Section~\ref{sec:results}), suggesting that direct dynamics-level regression combined with encoder stabilization provides a more effective inductive bias than contrastive learning for this setting. This advantage in zero-shot generalization persists when VIGOR is instantiated on an RSSM-based backbone (DreamerV3+VIGOR; Appendix~\ref{sec:vigor_on_dreamerv3}, Table~\ref{tab:dreamer_vigor_tasks}), indicating that it does not stem from the TD-MPC2 backbone alone.

\section{Training Procedure}\label{app:alg}
This section details the training procedure of VIGOR. Since VIGOR leaves the planning mechanism unchanged, we focus on world model learning and omit planner updates in Algorithm~\ref{algo:VIGOR}.

For a trajectory segment $\Gamma = (x_i, a_i, r_i, x_{i+1})_{i=t}^{t+H}$ sampled from replay buffer $\mathcal{B}$, the reward and Q-value losses are defined as:
\begin{equation}
    \mathcal{L}_{\text{rew}}(\theta) = \operatorname{CE}\bigl(\hat{r}_t, r_t\bigr), \qquad 
    \mathcal{L}_Q (\theta) = \operatorname{CE}\bigl(\hat{q}_t, q_t\bigr),
\end{equation}
where $\hat{r}_t = R_\theta(z_t, a_t)$ and $\hat{q}_t = Q_\theta(z_t, a_t)$ are the predicted reward and Q-value, respectively. The TD target is $q_t = r_t + \gamma\,\bar{Q}_\theta \bigl(z_{t+1}, \pi_\theta(z_{t+1})\bigr)$, where $\gamma$ is the discount factor, $\bar{Q}_\theta$ is an exponential moving average of the Q-value, $z_t = h_\theta(x_t)$ is the latent representation, and $\pi_\theta$ is a policy trained via entropy maximization. $\operatorname{CE}$ denotes the cross-entropy loss. For details on the world model architecture and planning procedure, we refer readers to~\citet{hansen2023td}.

\definecolor{highlightcolor}{rgb}{0.9,0.9,0.7}

\begin{algorithm}[h]
\caption{World model learning in VIGOR}\label{algo:VIGOR}
\begin{algorithmic}
\STATE {\bfseries Input:} Replay buffer $\mathcal{B}$; Horizon $H$; Weak and strong augmentation functions $\tau^w,\tau^s$; Temporal coefficient $\lambda$; Learning rate $\eta$; Momentum coefficient $\kappa$
\WHILE{not converged}
    \FOR{gradient step $t_g=1,2,\dots,T_g$}
        \STATE $\Gamma=(x_i,a_i,r_i,x_{i+1})_{i=t}^{t+H}\sim\mathcal{B}$ \hfill \COMMENT{Sample a trajectory segment from the replay buffer}
        \STATE $L\leftarrow 0$ \hfill \COMMENT{Initialize cumulative loss}
        \FOR{$i=t,t+1,\dots,t+H$}
            \STATE \colorbox{yellow!30}{$\upsilon^w\sim\Upsilon^w,\upsilon^s\sim\Upsilon^s$} \hfill \COMMENT{Sample augmentation parameters}
            \STATE \colorbox{yellow!30}{$L\leftarrow L+\lambda^{i-t}\mathcal{L}_{\text{VIGOR}}(\theta;\upsilon^w,\upsilon^s)$} \hfill \COMMENT{Calculate the per-step VIGOR loss (Eq.~\eqref{eq:VIGOR_obj})}
        \ENDFOR
        \STATE $\theta \leftarrow \theta - \eta\frac{1}{H+1}\nabla_\theta L$ \hfill \COMMENT{Update online network parameters}
        \STATE $\theta^- \leftarrow \kappa \theta^- + (1-\kappa) \theta$ \hfill \COMMENT{Update target network parameters}
    \ENDFOR
\ENDWHILE
\end{algorithmic}
\end{algorithm}

As shown in Algorithm~\ref{algo:VIGOR}, integrating VIGOR into a latent-space MBRL framework is straightforward: only the steps highlighted in \colorbox{yellow!30}{yellow} need to be added. In the main paper, we instantiate and validate VIGOR within the TD-MPC2 framework, without modifying its planner or optimization procedures. Appendix~\ref{sec:vigor_on_dreamerv3} presents a second instantiation on DreamerV3, an RSSM-based world model, together with the corresponding training procedure (Algorithm~\ref{algo:dreamer_vigor}).

\section{Discussion of Method Components}\label{sec:appendix_impl_method}
This section discusses why VIGOR's three components (AW, DC, and ES) require careful adaptation for the MBRL setting, how they are implemented, and how they interact. While related ideas have appeared in MFRL, directly applying them to latent-space MBRL is non-trivial due to the two-level vulnerability identified in Section~\ref{sec:introduction}: visual distractions first perturb encoder outputs, and these perturbations then compound through recursive world-model rollouts. We highlight the resulting design challenges and clarify how each component addresses them.

\paragraph{AW.}
The asymmetric batch-splitting design of AW, together with the rationale for routing every sample through the weak augmentation $\tau^w$ before applying $\tau^s$ to half the batch, is described in Section~\ref{sec:weak_strong}. Two consequences are worth highlighting in the context of the two-level vulnerability. First, the paired weak-only and weak-to-strong latents produced within a single mini-batch are precisely the supervision signal consumed by DC; without this pairing, DC would require a second encoder pass per sample, doubling encoder computation. Second, by applying strong augmentation only to a subset, AW balances two opposing failure modes observed in our ablations (Section~\ref{sec:core_components}; Appendix~\ref{sec:appendix_augmentation_ablation}): weak-only ablations exhibit insufficient robustness due to limited augmentation diversity, while strong-only and weak-to-strong-only ablations exhibit optimization instability under heavy perturbation.

\paragraph{DC.}
DC directly addresses the dynamics level of the two-level vulnerability. Perturbation-induced drift in the latent state corrupts the entire synthetic trajectory produced by the world model, leading to erroneous planning decisions. This problem is amplified relative to MFRL because both Q-values and transition dynamics are conditioned on encoder outputs, so a single perturbed latent propagates errors across the full rollout horizon.

To mitigate this, VIGOR enforces latent consistency by anchoring the dynamics model to weakly augmented successor states. While the dynamics model is trained on perturbed mixed latents ($z_t^m$) generated by AW, the DC objective regresses these predictions toward targets produced by weak augmentations, which serve as nearly noise-free anchors. This ensures that the world model produces consistent next-state predictions regardless of input perturbations, preventing the compounding of rollout errors over long horizons. By aligning predictions from OOD inputs with their in-distribution counterparts, DC stabilizes the forward dynamics and supports robust planning under visual distribution shifts.

\paragraph{ES.}
ES addresses the encoder level of the two-level vulnerability. While ES adapts the alignment objective of SODA~\citep{hansen2021generalization}, originally proposed for MFRL representation learning, its role in VIGOR is tailored to a specific failure mode in latent-space MBRL: encoder drift induced by DC's cross-augmentation supervision. A single shared encoder produces the latent representations that concurrently feed the dynamics model, the reward predictor, Q-value networks, and the MPC planner. The cross-augmentation supervision imposed by DC, which trains the dynamics on mixed latents ($z_t^m$), can inadvertently distort the encoder's representation. Without explicit constraints, encoder outputs tend to drift under distribution shifts, causing the dynamics-level compounding that DC is designed to prevent.

To mitigate this, VIGOR introduces an auxiliary ES loss that aligns weak-only and weak-to-strong encodings of the same frame. In VIGOR, ES acts as a structural stabilizer for the shared encoder, enabling DC to operate on consistent representations without distorting the latent geometry. The encoder thereby preserves task-relevant features under both clean and perturbed inputs, ensuring that initial latent states remain grounded for long-horizon planning in unseen environments.

\paragraph{Summary.}
VIGOR's effectiveness arises from the synergy between its three components, each targeting a different level of the world-model pipeline: AW establishes an asymmetric batching structure that provides paired latents without additional overhead; DC stabilizes the forward dynamics by anchoring predictions from perturbed mixed latents to weakly augmented targets, mitigating compounding rollout errors; and ES prevents the shared encoder from drifting under cross-augmentation pressure, ensuring that DC's dynamics-level constraint operates on stable representations. The three components are complementary: our ablations (Sections~\ref{sec:core_components}--\ref{sec:design_choices}; Appendix~\ref{sec:appendix_augmentation_ablation}--\ref{sec:appendix_design_choices}) demonstrate that removing any single component results in significant performance degradation, with the largest drops observed on Robosuite manipulation tasks under severe distribution shifts.

\clearpage
\section{Implementation Details}\label{sec:appendix_implementation_details}

\subsection{Code Base}
We evaluate VIGOR against four MFRL baselines (DrQ-v2~\citep{yarats2021mastering}, SVEA~\citep{hansen2021stabilizing}, SRM~\citep{huang2022spectrum}, and SimGRL~\citep{song2024simple}) and four MBRL baselines (TD-MPC2~\citep{hansen2023td}, DreamerV3~\citep{hafner2023mastering}, HRSSM~\citep{sun2024learning}, and Dr.~G~\citep{ha2023dream}).
We use the reference implementations from RL-ViGen~\citep{yuan2024rl}\footnote{\url{https://github.com/gemcollector/RL-ViGen}} for DrQ-v2, SVEA, and SRM, and the official implementations from the respective authors for SimGRL\footnote{\url{https://github.com/W-Song11/SimGRL-Code}}, DreamerV3\footnote{\url{https://github.com/danijar/dreamerv3}}, HRSSM\footnote{\url{https://github.com/bit1029public/HRSSM}}, and Dr.~G\footnote{\url{https://github.com/JeongsooHa/DrG}}. VIGOR is built on the official TD-MPC2 repository\footnote{\url{https://github.com/nicklashansen/tdmpc2}}, with modifications restricted to the additional components introduced in Section~\ref{sec:VIGOR}.

\subsection{Computational Resources}
All experiments were conducted on a single NVIDIA A5000 GPU. On average, a complete training and evaluation run took approximately 48 hours for the DMC tasks and 72 hours for the Robosuite tasks.

\subsection{Hyperparameters}
To ensure a fair comparison, we reuse common settings across all baselines whenever possible and follow each method's official configuration for method-specific values. Table~\ref{tab:common_hyperparams} reports hyperparameters across baselines (with method-specific values listed where they differ), and Table~\ref{tab:VIGOR_hyperparams} reports the full hyperparameter set for VIGOR. VIGOR inherits TD-MPC2 defaults~\citep{hansen2023td} unless otherwise noted; VIGOR-specific parameters are highlighted in \colorbox{yellow!30}{yellow}.

\begin{table}[ht]
\captionsetup{font=small}
\caption{\textbf{Hyperparameters across baselines.} Settings used in DMC and Robosuite experiments. Common values apply to all methods; method-specific values are listed per column where they differ. ``N/A'' denotes hyperparameters that do not apply to a given method (\eg Dreamer-family-specific state dimensions).}
\label{tab:common_hyperparams}
\centering
\resizebox{\textwidth}{!}{
\renewcommand{\arraystretch}{1.5}
\setlength{\aboverulesep}{0pt}
\setlength{\belowrulesep}{0pt}
\begin{tabular}{l|cccccccc}
\toprule
Hyperparameter &  DrQ-v2 & SVEA & SRM & \multicolumn{1}{c|}{SimGRL} & TD-MPC2 & DreamerV3 & HRSSM & Dr.~G \\
\midrule
\midrule
Discount factor $\gamma$ & \multicolumn{5}{c|}{0.99} & \multicolumn{2}{c|}{0.997} & 0.99 \\ 
\cmidrule(lr){2-9}
Replay buffer size $B$ & \multicolumn{8}{c}{Unlimited (same as $T_g$)} \\ 
\cmidrule(lr){2-9}
Action repeats & \multicolumn{3}{c|}{2} & Task-specific & \multicolumn{1}{|c|}{2} & 1 & \multicolumn{1}{|c|}{2} & Task-specific \\ 
\cmidrule(lr){2-9}
Frame stack $k$ & \multicolumn{5}{c|}{3} & \multicolumn{3}{c}{1} \\ 
\cmidrule(lr){2-9}
Pixel RGB image space & \multicolumn{4}{c|}{$o_t \in \mathcal{O}^{84\times84\times3}$} & \multicolumn{4}{c}{$o_t \in \mathcal{O}^{64\times64\times3}$} \\ 
\cmidrule(lr){2-9}
Maximum episode length & \multicolumn{8}{c}{$1{,}000$ (DMC); $500$ (Robosuite)} \\ 
\cmidrule(lr){2-9}
Batch size & \multicolumn{3}{c|}{\makecell[c]{512 (\textit{walker-walk}, \textit{walker-stand});\\256 (otherwise)}} & 128 & \multicolumn{1}{|c|}{256} & \multicolumn{2}{c|}{16} & 50 \\ 
\cmidrule(lr){2-9}
Total gradient steps $T_g$ & \multicolumn{8}{c}{1M} \\ 
\cmidrule(lr){2-9}
Total seeding steps & \multicolumn{3}{c|}{$4{,}000$} & $1{,}000$ & \multicolumn{1}{|c|}{\makecell[c]{$2{,}500$ (DMC);\\$1{,}250$ (Robosuite)}} & $1{,}024$ & \multicolumn{1}{|c|}{$2{,}500$} & \multicolumn{1}{c}{\makecell[c]{$2{,}500$ (DMC);\\$1{,}250$ (Robosuite)}} \\ 
\cmidrule(lr){2-9}
Periodic evaluation steps during training & \multicolumn{8}{c}{10k} \\ 
\cmidrule(lr){2-9}
$N$-step TD target & \multicolumn{3}{c|}{3} & \multicolumn{5}{c}{1} \\ 
\cmidrule(lr){2-9}
MLP hidden layer dimension & \multicolumn{4}{c|}{$1{,}024$} & 512 & \multicolumn{1}{|c|}{$1{,}024$} & 512 & \multicolumn{1}{|c}{200} \\ 
\cmidrule(lr){2-9}
Latent dimension & \multicolumn{3}{c|}{50} & 100 & \multicolumn{1}{|c|}{512} & $1{,}024$ & \multicolumn{1}{|c|}{512} & $1{,}024$ \\ 
\cmidrule(lr){2-9}
\makecell[l]{Deterministic state dimension \\(for Dreamer family)} & \multicolumn{5}{c|}{N/A} & $8{,}192$ & \multicolumn{1}{|c|}{512} & 200 \\ 
\cmidrule(lr){2-9}
\makecell[l]{Stochastic state dimension/categories \\(for Dreamer family)} & \multicolumn{5}{c|}{N/A} & $2{,}048$ & \multicolumn{1}{|c|}{$1{,}024$} & 30 \\ 
\cmidrule(lr){2-9}
Activation function & \multicolumn{4}{c|}{ReLU} & LayerNorm + Mish & \multicolumn{1}{|c|}{RMSNorm + SiLU} & LayerNorm + SiLU & \multicolumn{1}{|c}{ReLU + ELU} \\ 
\cmidrule(lr){2-9}
Target network EMA weight & \multicolumn{3}{c|}{1e-2} & \makecell[c]{1e-2 (Critic);\\5e-2 (Encoder)} & \multicolumn{1}{|c|}{1e-2} & \multicolumn{2}{c|}{2e-2} & 5e-2 \\ 
\bottomrule
\end{tabular}}
\end{table}

\begin{table}[H]
\captionsetup{font=small}
\caption{\textbf{VIGOR hyperparameters.} Full set of hyperparameters used in VIGOR, which builds upon TD-MPC2 defaults. VIGOR-specific parameters are highlighted in \colorbox{yellow!30}{yellow}.}
\label{tab:VIGOR_hyperparams}
\centering
\resizebox{0.75\textwidth}{!}{
\renewcommand{\arraystretch}{1.3}
\begin{tabular}{l|c}
\toprule
\textbf{Hyperparameter} & \textbf{Value} \\
\midrule
\underline{MPC planning} & \\
Planning Horizon $H$ & 3 \\
Std. range & $\sigma\in[0.05, 2]$ \\
Population size & 512 \\
Elite fraction & 64 \\
Iterations & 6 \\
Policy prior samples & 24 \\
Sampling temperature & 0.5 \\ 
\\
\underline{Model learning} &  \\
Temporal coefficient $\lambda$ & 0.5 \\
Reward loss coefficient $c_1$ & 0.1 \\
Q-value loss coefficient $c_2$ & 0.1 \\
Latent consistency loss coefficient $c_3$ & 20 \\ 
\\
\underline{Optimization} &  \\
Learning rate $\eta$ & 3e-4 \\
Periodic target network ($\theta^-$) update steps & 1 \\
Momentum coefficient for Q-network ($\kappa$) & 0.99 \\
Optimizer & Adam ($\beta_1=0.9$,\ $\beta_2=0.999$) \\
Exploration schedule (std) & Linear  (0.5,\ 0.05,\ 25,000\ steps) \\
Planning horizon schedule & Linear (1,\ 5,\ 25,000\ steps) \\ 
\\
\underline{VIGOR-specific hyperparameters} & \\
\cellcolor{yellow!30}{Weak augmentation $\tau^w$} & \cellcolor{yellow!30}{\textit{random-shift}: $\texttt{padding}\quad p=4$} \\
\cellcolor{yellow!30}{Strong augmentation $\tau^s$} & \cellcolor{yellow!30}{\textit{random-overlay}:
$
\left\{
\begin{aligned}
    & \texttt{linear interpolation}\quad\delta=0.5 \\
    & \texttt{Image dataset}\quad\mathcal{D}=\texttt{Places}~\text{\citep{zhou2017places}}
\end{aligned}
\right.
$} \\
\cellcolor{yellow!30}{Augmentation ratio $\zeta$} & \cellcolor{yellow!30}{1.0 (half weak-only, half weak-to-strong)}\\
\bottomrule
\end{tabular}}
\end{table}

\section{Evaluation Details}\label{sec:appendix_evaluation_details}
We follow the zero-shot evaluation protocol of \citet{yuan2024rl} and define seven distraction categories for DMC, each evaluated at two difficulty levels (\textit{easy} and \textit{hard}), resulting in a total of fourteen DMC evaluation settings. For Robosuite, we consider three difficulty levels (\textit{easy}, \textit{hard}, and \textit{extreme}). These settings induce distribution shifts in appearance factors such as color, lighting, background, and camera position, providing a rigorous testbed for zero-shot visual generalization.

\subsection{Environments and Tasks}
\begin{figure}[ht]
    \centering
    \includegraphics[width=\textwidth]{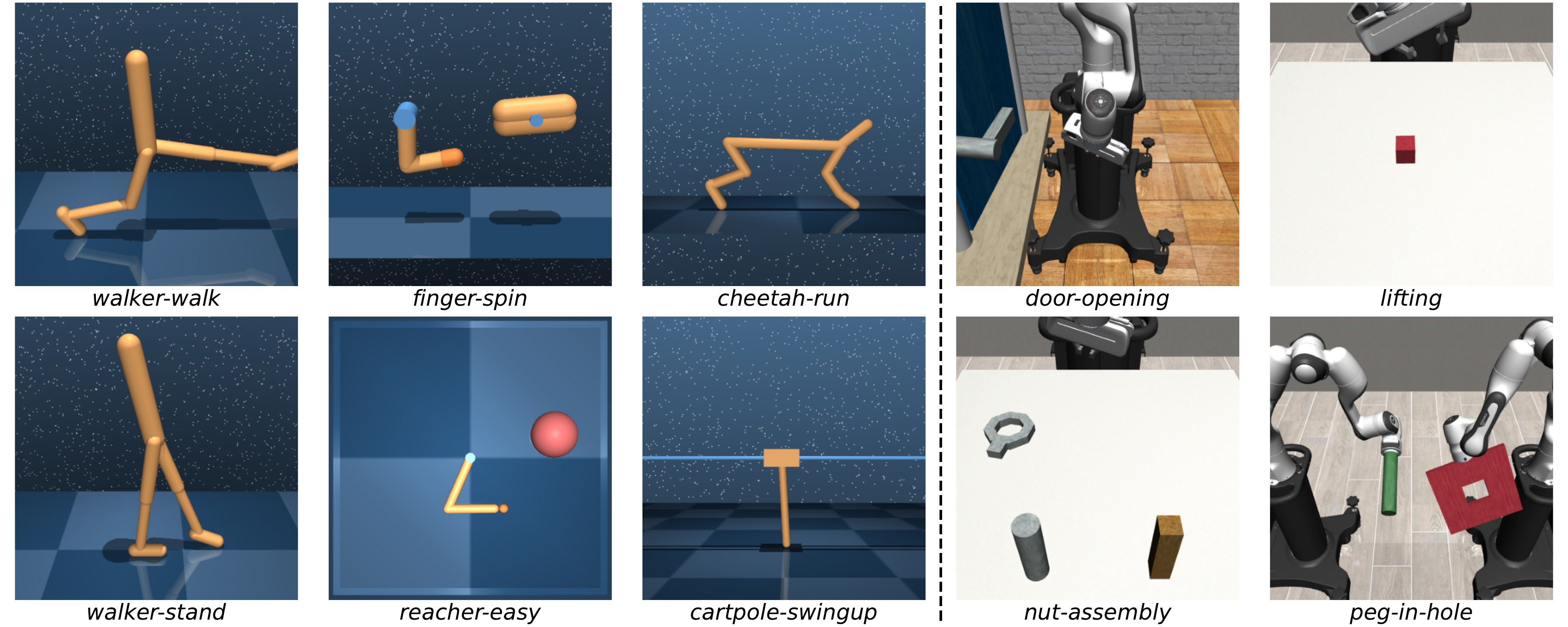}
    \captionsetup{font=small}
    \caption{\textbf{Environments and tasks.} Six visuomotor control tasks from DMC~\citep{tassa2018deepmind} and four robotic manipulation tasks from Robosuite~\citep{zhu2020robosuite}.}
    \label{fig:envs}
\end{figure}

We benchmark VIGOR on two widely used continuous-control suites: DMC and Robosuite. Representative environments are illustrated in Figure~\ref{fig:envs}.

In \textbf{DMC}~\citep{tassa2018deepmind}, we consider six visuomotor control tasks: \textit{cartpole-swingup}, where the agent must swing up and balance a cartpole; \textit{finger-spin}, which requires continuous spinning of a planar finger; \textit{walker-walk}, which trains a bipedal walker to move forward stably; \textit{walker-stand}, which focuses on maintaining balance in an upright posture; \textit{cheetah-run}, where the agent learns high-speed forward locomotion; and \textit{reacher-easy}, which involves controlling a 2-DoF arm to reach a target location.

For \textbf{Robosuite}~\citep{zhu2020robosuite}, we evaluate four robotic manipulation tasks: \textit{door-opening}, where the robot arm must open a hinged door; \textit{nut-assembly}, which requires placing a round nut onto a peg; \textit{peg-in-hole}, where a peg must be inserted into a narrow slot; and \textit{lifting}, which involves grasping and lifting a cube.

\subsection{Baseline Descriptions}\label{sec:appendix_baselines}
This subsection provides detailed descriptions of the baselines used in our experiments, complementing the summary in Section~\ref{sec:exp_setup}. 

\paragraph{Selection criteria.}
We select baselines according to the following criteria: (i)~The method must provide an official codebase in its original paper or in a recognized benchmarking framework (\eg RL-ViGen~\citep{yuan2024rl}). We exclude methods whose scores can only be reproduced from unofficial or re-implemented code, as this may introduce unreliable or biased results. (ii)~The method must explicitly target visual robustness or zero-shot visual generalization in RL. (iii)~Reproduced scores must be broadly consistent with those reported in the original publication; methods with large unexplained discrepancies are excluded.

We additionally include three strong baselines (DrQ-v2, TD-MPC2, and DreamerV3) that do not explicitly target zero-shot visual generalization but serve essential reference roles: DrQ-v2 as a widely adopted MFRL baseline for sample efficiency, TD-MPC2 as VIGOR's backbone, and DreamerV3 as a state-of-the-art reconstruction-based MBRL method. Including these methods enables direct assessment of how much zero-shot robustness VIGOR adds over its backbone and how general-purpose methods perform under distribution shifts without explicit robustness mechanisms.

\paragraph{MFRL baselines.} 
(i)~DrQ-v2~\citep{yarats2021mastering} combines random-shift augmentation with an off-policy actor--critic architecture, achieving strong sample efficiency on pixel-based continuous control. (ii)~SVEA~\citep{hansen2021stabilizing} stabilizes off-policy Q-learning by maintaining a weighted mixture of augmented and clean Q-targets, preventing the value function from being corrupted by augmentation noise. (iii)~SRM~\citep{huang2022spectrum} applies spectrum-based augmentations in the frequency domain to enhance robustness to spatial corruptions. (iv)~SimGRL~\citep{song2024simple} combines random-overlay with random-shift augmentation, injecting task-irrelevant background perturbations to mitigate observational overfitting.

\paragraph{MBRL baselines.} 
(v)~TD-MPC2~\citep{hansen2023td} achieves strong performance via decoder-free latent-space planning with temporal-difference learning. It serves as VIGOR's backbone. (vi)~DreamerV3~\citep{hafner2023mastering} employs latent imagination with pixel-level reconstruction and achieves competitive performance across diverse control tasks. (vii)~HRSSM~\citep{sun2024learning} extends DreamerV3 with a bisimulation-based hybrid RSSM to extract task-relevant latent representations without pixel-level reconstruction. (viii)~Dr.~G~\citep{ha2023dream} extends Dreamer~\citep{hafner2019dream} with dual contrastive objectives and a recurrent inverse-dynamics loss, explicitly targeting zero-shot visual generalization. It is the most directly comparable prior method.

\subsection{Visual Distraction Categories}\label{sec:appendix_distractions}
We consider seven types of distractions. For DMC, each type is evaluated separately at two difficulty levels (\textit{easy} and \textit{hard}); for Robosuite, the RL-ViGen evaluation splits (\textit{easy}, \textit{hard}, and \textit{extreme}) combine these factors within predefined scenes (Appendix~\ref{sec:appendix_eval_scenes}). The seven types are defined as follows:
\begin{itemize}
    \item \textit{object-color}: Change the body color of the agent.
    \begin{itemize}
        \item Uniformly sample the parameters of the color (\ie \textit{(r,g,b)}) from the pre-defined distribution for each difficulty.
    \end{itemize}
    \item \textit{background-color}: Change the background color of the environment (\eg terrain grid or sky).
    \begin{itemize}
        \item Uniformly sample the parameters of the color (\ie \textit{(r,g,b)}) from the pre-defined distribution for each difficulty.
    \end{itemize}
    \item \textit{background-video}: Overlay the background with the randomly sampled natural video.
    \begin{itemize}
        \item Sample a random video with the same width and height as the original image from a set of natural videos~\citep{stone2021distracting}.
        \item Overlay the video only to the background sky for \textit{easy} and to all backgrounds, including the ground terrain other than the agent, for \textit{hard} difficulty.
    \end{itemize}
    \item \textit{cam-pos}: Change the position of the tracking camera's focus by randomly adding noise offset.
    \begin{itemize}
        \item Let the initial position of the tracking camera's focus be $X_{\text{cam}}=(c_x, c_y, c_z)$ in Euclidean space.
        \item Sample a random offset $\delta\in\mathbb{R}^3$ from the uniform distribution with different bounds: $\mathcal{U}(-0.08,0.08)$ for \textit{easy} and $\mathcal{U}(-0.15,0.15)$ for \textit{hard} difficulty.
        \item Inject the offset to the initial position of the camera; $X_{\text{cam}}=(c_x+\delta_x, c_y+\delta_y, c_z+\delta_z)$.
    \end{itemize}
    \item \textit{light-position}: Change the position and orientation of the tracking light of the agent.
    \begin{itemize}
        \item Following the approach used in~\citep{stone2021distracting}, the tracking light's coordinate is parameterized as the spherical coordinate; $(\phi,\vartheta,\rho)$ where $\phi$ is azimuth, $\vartheta$ is inclination, and $\rho$ is the radius of the sphere.
        \item Sample $\phi$ from the normal distribution $\mathcal{N}(\pi/6,1)$ for \textit{easy} and $\mathcal{N}(\pi/3,1)$ for \textit{hard} difficulty.
        \item Sample $\vartheta\sim\mathcal{N}(2\pi,1)$ and transform the initial pose of the tracking light $X_{\text{light}}$ to $(\phi,\vartheta,\rho)$, where $\rho = \lVert X_{\text{light}} \rVert$ is the radius (Euclidean norm of the initial position).
    \end{itemize}
    \item \textit{light-color}: Change the color of the tracking light of the agent.
    \begin{itemize}
        \item Uniformly sample the parameters of the color (\ie \textit{(r,g,b)}) from the pre-defined distribution for each difficulty.
    \end{itemize}
    \item \textit{moving-light}: Rotate the tracking light around the agent. 
    \begin{itemize}
        \item As in \textit{light-position}, the spherical coordinate of the tracking light is randomly initialized as $(\phi,\vartheta,\rho)$.
        \item Let the speed of azimuth rotation as $\Delta_\phi=\pi/200$ for \textit{easy} and $\Delta_\phi=\pi/100$ for \textit{hard} difficulty.
        \item Rotate the tracking light counterclockwise along the azimuth axis at every time step; $(\phi,\vartheta,\rho)\leftarrow(\phi,\vartheta,\rho)+(\Delta_\phi,0,0)$.
    \end{itemize}
\end{itemize}

\subsection{Evaluation Protocol}
For DMC, each of the seven distraction types listed above includes two difficulty levels: \textit{easy} and \textit{hard}. Robosuite follows the predefined evaluation splits (\textit{eval‑easy}, \textit{eval‑hard}, and \textit{eval‑extreme}) as introduced in RL‑ViGen~\citep{yuan2024rl}.

\subsection{Evaluation Scenes}\label{sec:appendix_eval_scenes}
\begin{figure}[ht]
    \centering
    \captionsetup[sub]{font=small}
    \subfigure[Evaluation scenes in DMC]{\centering\includegraphics[height=0.33\textwidth]{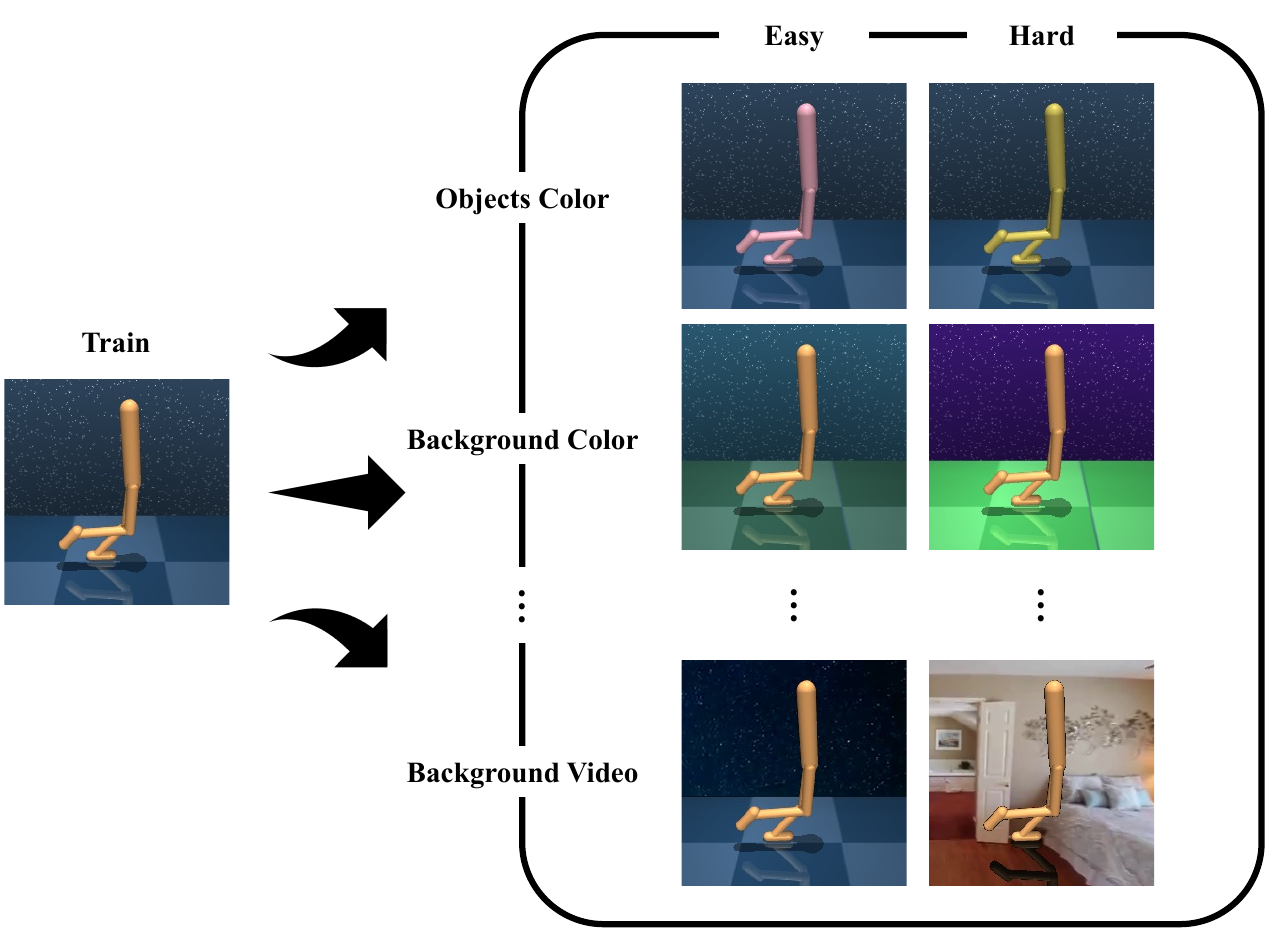}\label{fig:eval_set_dmc}}
    ~~\subfigure[Evaluation scenes in Robosuite]{\centering\includegraphics[height=0.33\textwidth]{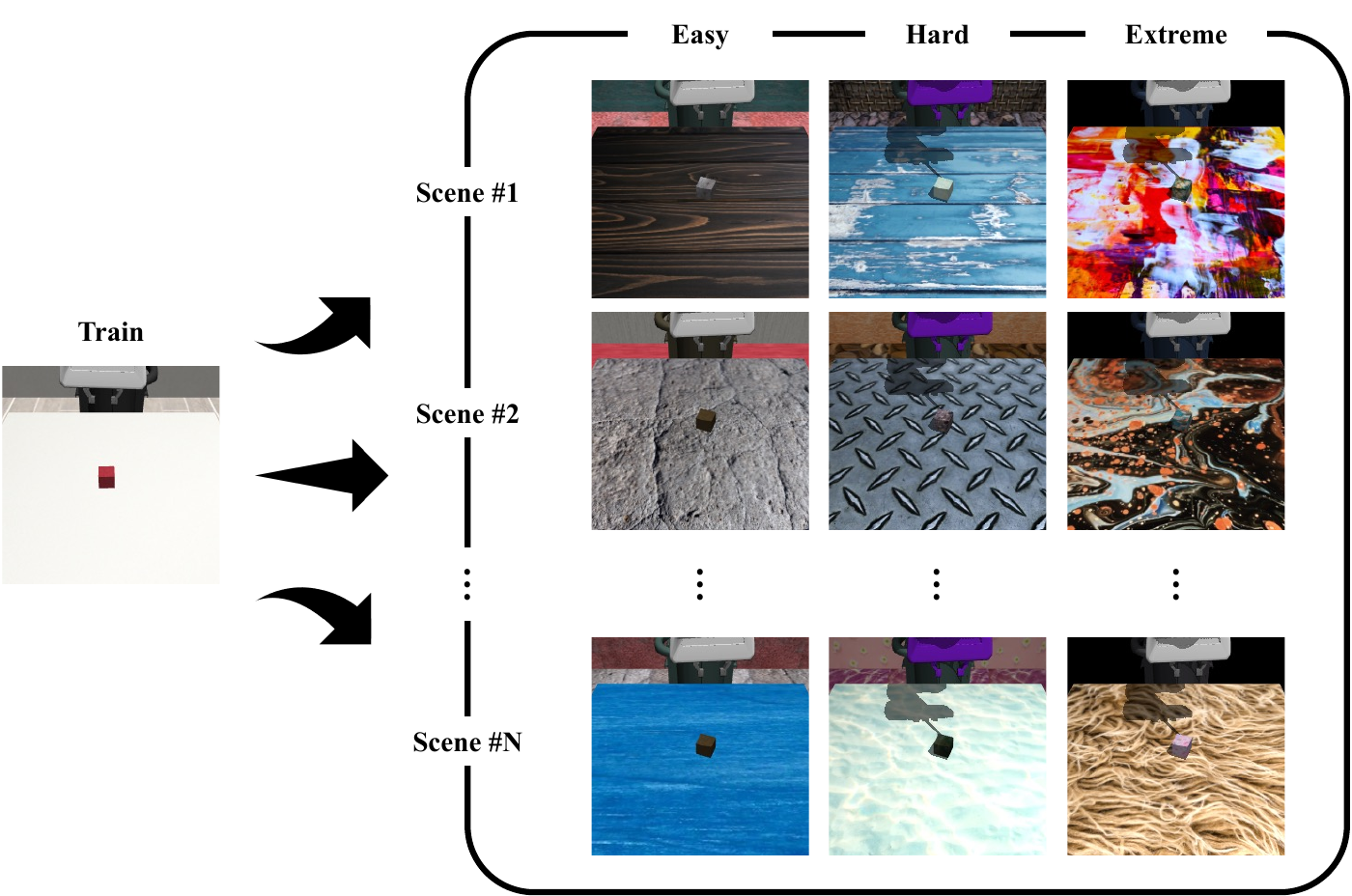}\label{fig:eval_set_robosuite}}
    \captionsetup{font=small}
    \caption{\textbf{Evaluation scenes.} Agents are trained in clean environments and evaluated \textit{zero-shot} on seven distraction types for DMC (two difficulty levels: \textit{easy} and \textit{hard}) and ten distraction scenes for Robosuite (three difficulty levels: \textit{easy}, \textit{hard}, and \textit{extreme}).}
    \label{fig:eval_sets}
\end{figure}

\begin{figure}[ht]
    \centering
    \includegraphics[width=\textwidth]{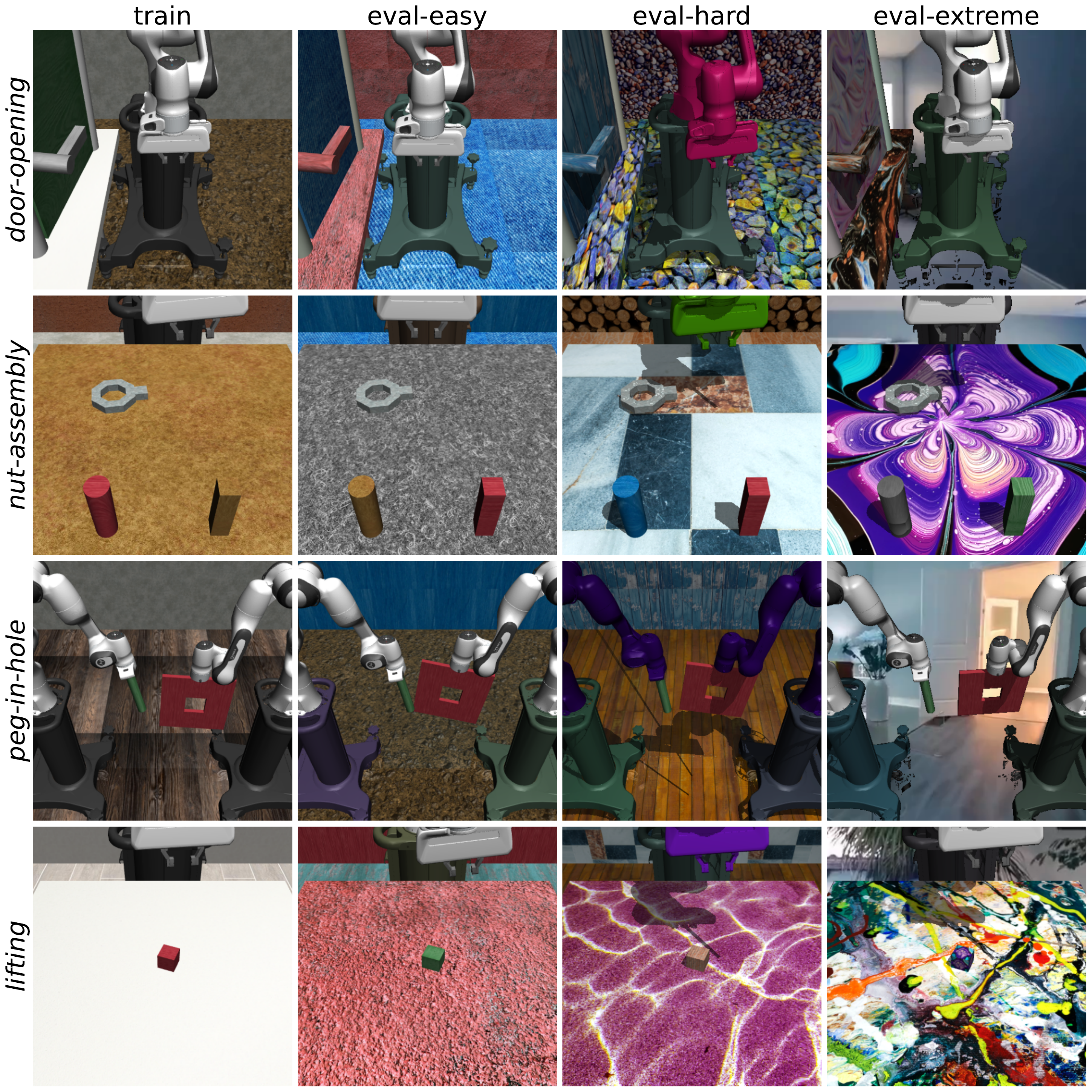}
    \captionsetup{font=small}
    \caption{\textbf{Visual generalization evaluation in Robosuite.} Visual generalization setup in Robosuite, which includes four robotic manipulation tasks: \textit{door-opening}, \textit{nut-assembly}, \textit{peg-in-hole}, and \textit{lifting}. Each column represents a different level of visual distraction, from left to right: \textit{train}, \textit{eval-easy}, \textit{eval-hard}, and \textit{eval-extreme}. Each row shows a different task.}
    \label{fig:eval_envs}
\end{figure}

Representative evaluation scenes for DMC and Robosuite are shown in Figures~\ref{fig:eval_set_dmc} and~\ref{fig:eval_set_robosuite}, respectively. Agents are trained in the default \textit{clean} setting and then evaluated \textit{zero-shot} under diverse unseen distractions. Figure~\ref{fig:eval_envs} illustrates the four Robosuite tasks (\textit{door-opening}, \textit{nut-assembly}, \textit{peg-in-hole}, and \textit{lifting}) under increasingly severe visual distractions, from \textit{eval-easy} to \textit{eval-extreme}.

\clearpage
\section{Supplementary Results}\label{sec:appendix_supp_results}
This appendix complements Section~\ref{sec:results} with additional figures, tables, and analyses. 

\subsection{Zero-Shot Generalization}\label{sec:appendix_zsg}
\begin{figure}[ht]
\centering
\includegraphics[width=\textwidth]{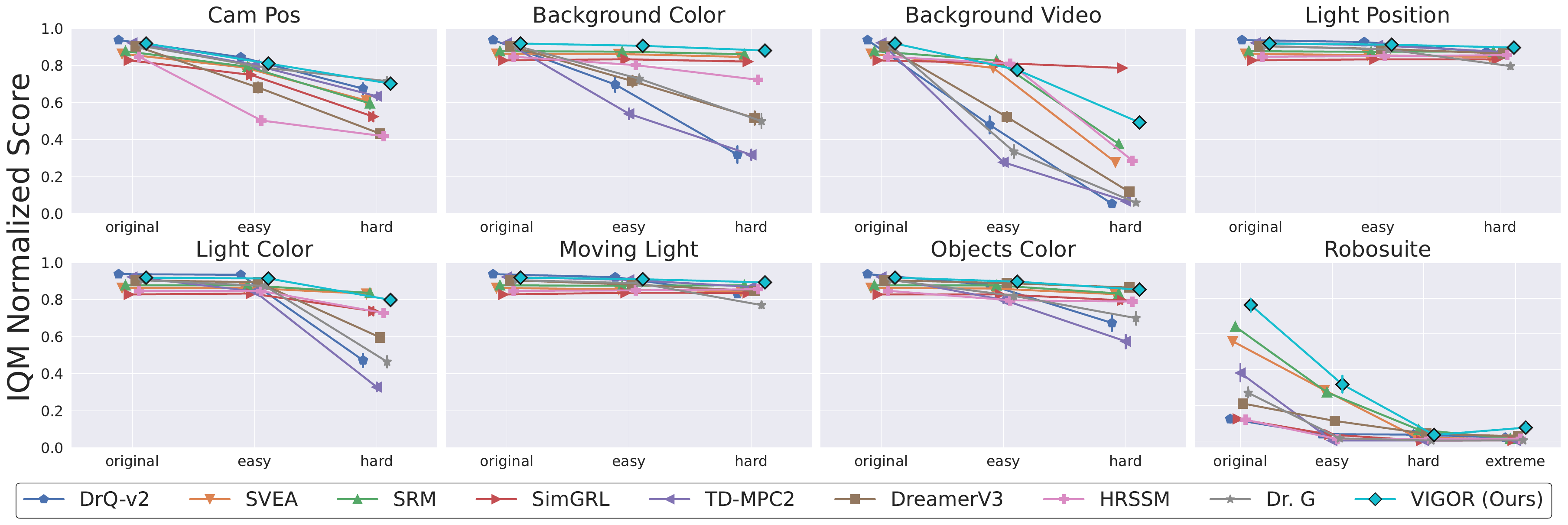}
\captionsetup{font=small}
\caption{\textbf{Zero-shot generalization across distraction types and difficulty levels.} Each subplot reports IQM normalized scores for a single distraction type across difficulty levels. DMC: seven distraction types $\times$ three settings (\textit{original}, \textit{easy}, \textit{hard}). Robosuite: four settings (\textit{original}, \textit{easy}, \textit{hard}, \textit{extreme}).}
\label{fig:eval_results_over_eval_types}
\end{figure}

\begin{figure}[ht]
\centering
\includegraphics[width=\textwidth]{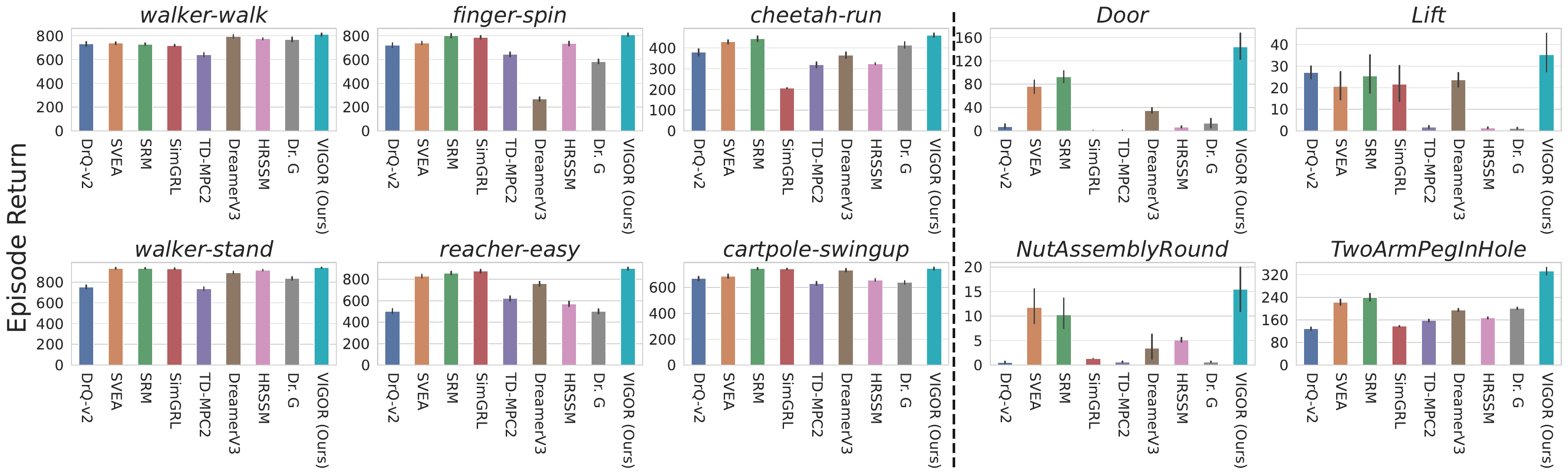}
\captionsetup{font=small}
\caption{\textbf{Zero-shot generalization across tasks.} Mean episode returns across six DMC tasks and four Robosuite tasks, averaged over all distraction types and difficulty levels. Error bars denote standard deviations.}
\label{fig:eval_results_over_envs}
\end{figure}

This subsection provides a detailed breakdown of our zero-shot generalization results, complementing the aggregate scores in Section~\ref{sec:results}. Full results across distraction types and tasks are reported in Figures~\ref{fig:eval_results_over_eval_types} and~\ref{fig:eval_results_over_envs}, with per-task averages summarized in Table~\ref{tab:visual_generalization}.

\paragraph{Performance across distraction types and difficulty levels.}
Figure~\ref{fig:eval_results_over_eval_types} reports zero-shot generalization on DMC across fourteen conditions (seven distraction types $\times$ two difficulty levels: \textit{easy} and \textit{hard}) and on Robosuite across three difficulty levels (\textit{easy}, \textit{hard}, and \textit{extreme}); each panel additionally shows the \textit{original} (distraction-free) condition as a reference. Across all methods, returns tend to degrade monotonically as distraction difficulty increases (\textit{original}~$\rightarrow$~\textit{easy}~$\rightarrow$~\textit{hard}~$\rightarrow$~\textit{extreme}), reflecting the inherent challenge of distribution shifts in visual control. However, the magnitude of this degradation differs substantially across methods.

On DMC, baseline behavior diverges sharply across distraction types. TD-MPC2, despite being one of the strongest latent-space MBRL methods under clean settings, exhibits sharp drops under hard distractions on \textit{background-color} (${\approx}0.32$), \textit{background-video} (${\approx}0.05$), and \textit{light-color} (${\approx}0.32$), highlighting its sensitivity to severe visual shifts. DreamerV3 and HRSSM follow similar trajectories, with HRSSM degrading even under \textit{easy} \textit{cam-pos} perturbations. SimGRL maintains relatively stable performance across most distractions and achieves the highest score on \textit{background-video} hard (${\approx}0.79$), reflecting its training strategy that injects diverse background perturbations. In contrast, VIGOR sustains high returns across nearly all distraction types and difficulty levels, ranking first or within a narrow margin of the best method on every distraction type except \textit{background-video}. On Robosuite, where most baselines collapse to near-zero performance under hard and extreme shifts, VIGOR maintains the highest returns at every difficulty level, demonstrating its robustness under severe distribution shifts.

\begin{table}[ht]
\captionsetup{font=small}
\caption{\textbf{Zero-shot generalization scores.} Mean episode returns ($\pm$ std) averaged over all distraction types and difficulty levels for each task. Boldface denotes the best score per task.}
\label{tab:visual_generalization}
\centering
\begin{small}
\begin{sc}
\resizebox{\textwidth}{!}{
\renewcommand{\arraystretch}{1.5}
\begin{tabular}{ll|cccc|ccccc}
\toprule
Env & Task & DrQ-v2 & SVEA & SRM & SimGRL & TD-MPC2 & DreamerV3 & HRSSM & Dr.~G & VIGOR (Ours) \\
\midrule
\midrule
\multirow[c]{7}{*}{DMC} & walker-walk & $731.5 \pm 344.4$ & $738.5 \pm 177.0$ & $730.2 \pm 153.6$ & $719.3 \pm 128.5$ & $642.2 \pm 344.8$ & $795.5 \pm 279.0$ & $775.3 \pm 153.0$ & $769.8 \pm 331.9$ & \cellcolor{cyan!15}{$\mathbf{811.3 \pm 201.7}$} \\
 & finger-spin & $721.6 \pm 373.0$ & $739.8 \pm 246.4$ & $801.5 \pm 319.8$ & $787.9 \pm 242.5$ & $643.9 \pm 336.7$ & $268.2 \pm 301.6$ & $734.6 \pm 319.1$ & $584.2 \pm 321.9$ & \cellcolor{cyan!15}{$\mathbf{809.1 \pm 212.2}$} \\
 & cheetah-run & $379.3 \pm 295.5$ & $429.1 \pm 179.1$ & $443.7 \pm 228.9$ & $206.1 \pm 25.6$ & $318.4 \pm 243.4$ & $364.3 \pm 281.2$ & $323.8 \pm 85.7$ & $413.7 \pm 286.7$ & \cellcolor{cyan!15}{$\mathbf{461.6 \pm 176.7}$} \\
 & walker-stand & $754.3 \pm 313.7$ & $933.5 \pm 132.0$ & $932.7 \pm 105.0$ & $929.5 \pm 110.3$ & $737.1 \pm 320.5$ & $891.9 \pm 233.4$ & $915.7 \pm 116.0$ & $834.6 \pm 274.9$ & \cellcolor{cyan!15}{$\mathbf{938.6 \pm 72.9}$} \\
 & reacher-easy & $501.8 \pm 449.9$ & $829.8 \pm 322.1$ & $858.2 \pm 291.8$ & $875.5 \pm 286.8$ & $621.9 \pm 422.5$ & $758.5 \pm 345.7$ & $571.2 \pm 427.5$ & $500.9 \pm 388.8$ & \cellcolor{cyan!15}{$\mathbf{899.0 \pm 232.3}$} \\
 & cartpole-swingup & $668.3 \pm 289.2$ & $686.5 \pm 281.8$ & $745.7 \pm 143.4$ & $742.7 \pm 87.5$ & $631.1 \pm 241.7$ & $733.5 \pm 212.5$ & $657.8 \pm 192.1$ & $638.7 \pm 274.7$ & \cellcolor{cyan!15}{$\mathbf{746.0 \pm 173.8}$} \\
\cmidrule(lr){2-11}
 & Average & $626.1 \pm 375.1$ & $726.2 \pm 279.4$ & $752.0 \pm 269.8$ & $710.2 \pm 292.9$ & $599.1 \pm 349.8$ & $635.3 \pm 363.2$ & $663.1 \pm 308.4$ & $623.7 \pm 347.5$ & \cellcolor{cyan!30}{$\mathbf{777.6 \pm 241.5}$} \\
\midrule
\multirow[c]{5}{*}{Robosuite} & door-opening & $7.6 \pm 32.2$ & $76.1 \pm 98.1$ & $92.5 \pm 90.9$ & $1.0 \pm 0.4$ & $1.0 \pm 1.0$ & $35.0 \pm 41.5$ & $6.8 \pm 12.7$ & $13.2 \pm 69.0$ & \cellcolor{cyan!15}{$\mathbf{143.9 \pm 201.6}$} \\
 & lifting & $27.0 \pm 25.1$ & $20.6 \pm 58.9$ & $25.5 \pm 78.7$ & $21.6 \pm 74.1$ & $1.8 \pm 5.0$ & $23.7 \pm 27.7$ & $1.5 \pm 2.7$ & $1.3 \pm 2.5$ & \cellcolor{cyan!15}{$\mathbf{35.4 \pm 74.9}$} \\
 & nut-assembly & $0.6 \pm 1.6$ & $11.8 \pm 31.7$ & $10.2 \pm 26.8$ & $1.3 \pm 0.0$ & $0.7 \pm 1.1$ & $3.4 \pm 21.7$ & $5.2 \pm 3.8$ & $0.7 \pm 1.0$ & \cellcolor{cyan!15}{$\mathbf{15.5 \pm 39.3}$} \\
 & peg-in-hole & $128.8 \pm 45.0$ & $222.6 \pm 86.6$ & $239.1 \pm 114.2$ & $137.3 \pm 12.9$ & $157.4 \pm 35.9$ & $195.2 \pm 39.1$ & $166.9 \pm 24.1$ & $200.8 \pm 28.7$ & \cellcolor{cyan!15}{$\mathbf{332.6 \pm 114.2}$} \\
\cmidrule(lr){2-11}
 & Average & $41.0 \pm 59.9$ & $82.8 \pm 111.9$ & $91.8 \pm 123.4$ & $40.3 \pm 68.0$ & $40.2 \pm 70.1$ & $64.3 \pm 83.5$ & $45.1 \pm 71.7$ & $54.0 \pm 92.8$ & \cellcolor{cyan!30}{$\mathbf{131.8 \pm 176.1}$} \\
\bottomrule
\end{tabular}
}
\end{sc}
\end{small}
\end{table}

\paragraph{Task-level performance breakdown.}
Table~\ref{tab:visual_generalization} and Figure~\ref{fig:eval_results_over_envs} provide a detailed breakdown of zero-shot generalization performance across all tasks in DMC and Robosuite. VIGOR ranks first among all baselines on every individual task in both benchmarks, demonstrating that its advantage is not driven by a few favorable cases but by uniformly strong performance across the task set.

On DMC, VIGOR achieves the highest average return ($777.6$), surpassing the second-best baseline (SRM, $752.0$) by $3.4\%$. At the task level, VIGOR attains the strongest scores on all six tasks, exceeding the respective second-best methods by $2.0\%$ on \textit{walker-walk} ($811.3$ vs.\ DreamerV3, $795.5$), $1.0\%$ on \textit{finger-spin} ($809.1$ vs.\ SRM, $801.5$), $4.0\%$ on \textit{cheetah-run} ($461.6$ vs.\ SRM, $443.7$), $0.5\%$ on \textit{walker-stand} ($938.6$ vs.\ SVEA, $933.5$), $2.7\%$ on \textit{reacher-easy} ($899.0$ vs.\ SimGRL, $875.5$), and $0.0\%$ on \textit{cartpole-swingup} ($746.0$ vs.\ SRM, $745.7$).

On Robosuite, the performance gains are substantially more pronounced. VIGOR achieves the highest average return ($131.8$), surpassing the second-best baseline (SRM, $91.8$) by $43.6\%$, and attains the best scores on all four tasks: \textit{door-opening} ($143.9$, $+55.6\%$ over SRM, $92.5$), \textit{lifting} ($35.4$, $+31.1\%$ over DrQ-v2, $27.0$), \textit{nut-assembly} ($15.5$, $+31.4\%$ over SVEA, $11.8$), and \textit{peg-in-hole} ($332.6$, $+39.1\%$ over SRM, $239.1$). Many baselines degrade to near-random performance in this setting, whereas VIGOR maintains robustness under visually challenging manipulation conditions.

\paragraph{Summary.}
These findings confirm that VIGOR's three components (AW, DC, and ES) are essential for mitigating severe distribution shifts. By jointly promoting augmentation-invariant latent dynamics, stabilizing transition predictions under perturbed inputs, and preventing encoder drift, VIGOR achieves the highest zero-shot generalization on every task across both benchmarks, not only in aggregate but also at the individual task level. This consistent dominance corroborates that conventional latent-space MBRL is fragile to OOD perturbations, whereas VIGOR's latent-space consistency enables robust generalization without test-time adaptation.

\subsection{Sample Efficiency}\label{sec:appendix_se}
This subsection provides a detailed per-task comparison of sample efficiency, complementing the aggregate results in Section~\ref{sec:results}.

\begin{figure}[ht]
\centering
\includegraphics[width=\textwidth]{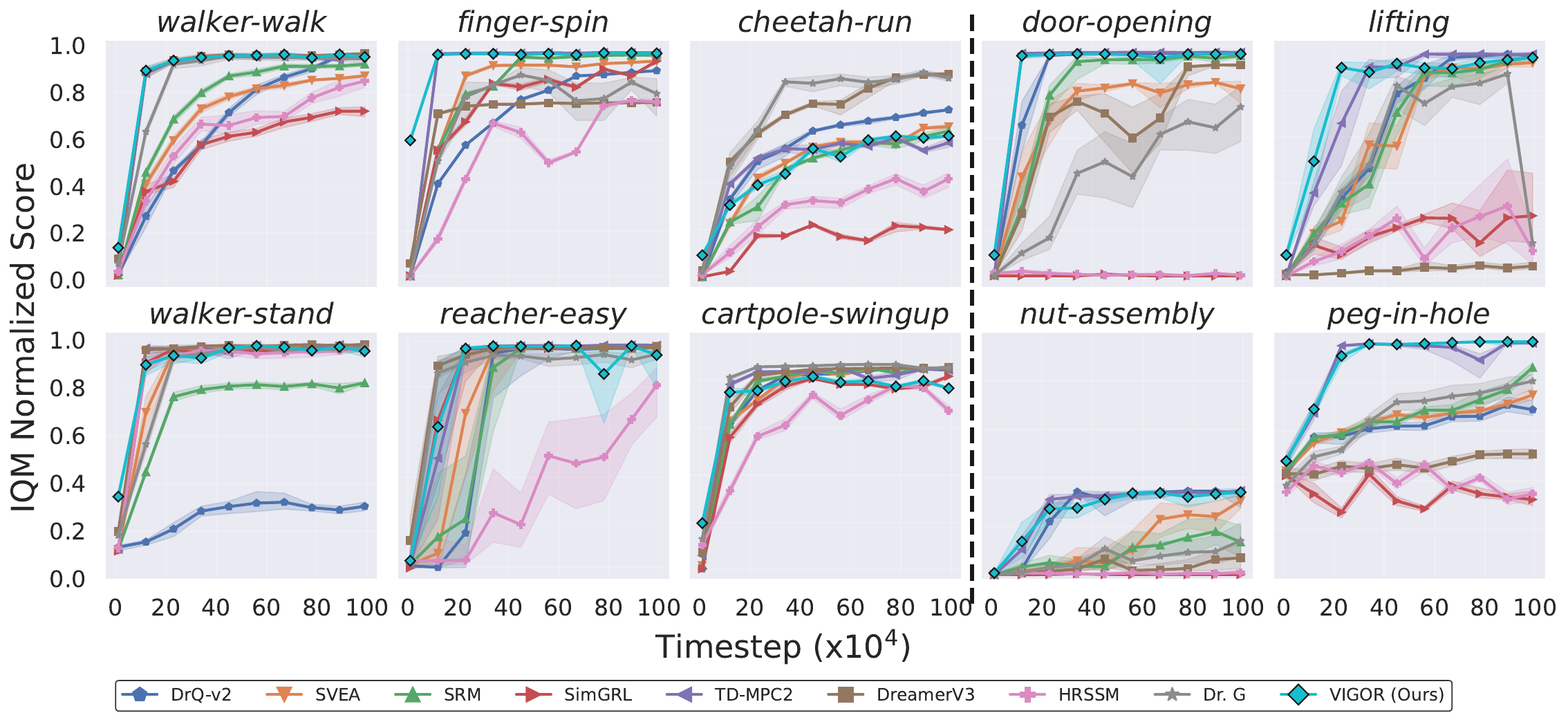}
\captionsetup{font=small}
\caption{\textbf{Per-task sample efficiency.} Learning curves reported as IQM normalized scores across methods. Shaded regions denote 95\% stratified bootstrap confidence intervals.}
\label{fig:eval_during_train_results_over_envs}
\end{figure}

\paragraph{Training curves.}
Figure~\ref{fig:eval_during_train_results_over_envs} presents per-task learning curves for DMC and Robosuite, reporting IQM normalized scores across all methods. While no single method dominates across all tasks, VIGOR consistently ranks among the top performers. Notably, VIGOR maintains this level of efficiency despite being trained with more challenging objectives that incorporate visual perturbations, confirming that robustness does not come at the cost of efficiency.

\paragraph{Evaluation protocol.}
To quantify sample efficiency, we adopt the evaluation protocol of \citet{mai2022sample}. We define TD-MPC2 as the oracle baseline, since it serves as VIGOR's backbone and provides the most direct reference for assessing the efficiency cost of the added robustness components. We then measure the number of environment episodes required for each non-oracle method to reach $25\%$, $50\%$, and $75\%$ of TD-MPC2's asymptotic return. If an agent fails to reach a threshold within the training budget, we assign the maximum episode count as a penalty ($1{,}000$ for DMC and $2{,}000$ for Robosuite).

\paragraph{Quantitative results.}
As shown in Table~\ref{tab:sample_efficiency}, VIGOR reaches every threshold in the fewest episodes among all non-oracle baselines on both benchmarks, demonstrating that the added robustness components incur minimal cost in sample efficiency relative to the TD-MPC2 backbone.

On DMC, VIGOR reaches the $25\%$, $50\%$, and $75\%$ thresholds in $395$, $412$, and $422$ episodes, respectively. The second-best baseline at each threshold is Dr.~G ($415$ at $25\%$; $540$ at $50\%$) or DreamerV3 ($553$ at $75\%$); VIGOR requires $4.8\%$ to $23.7\%$ fewer episodes than the second-best baseline at each threshold. On Robosuite, the efficiency advantage is more pronounced: VIGOR requires $655$, $720$, and $740$ episodes to reach the three thresholds, while the second-best baseline requires $1{,}240$ (SRM), $1{,}340$ (SRM), and $1{,}365$ (DrQ-v2) episodes, corresponding to roughly $1.8$--$1.9\times$ more interactions across all thresholds. Many baselines fail to reach the higher thresholds within the training budget, indicating that VIGOR's sample-efficiency advantage amplifies under visually complex manipulation tasks. These results confirm that AW, DC, and ES preserve the strong sample efficiency of the TD-MPC2 backbone while enabling robustness to unseen distractions.

\begin{table}[ht]
\captionsetup{font=small}
\caption{\textbf{Sample efficiency.} Mean number of episodes (lower is better) required to reach $25\%$, $50\%$, and $75\%$ of the oracle baseline (TD-MPC2), following the protocol of \citet{mai2022sample}. Boldface denotes the best score per row (excluding the oracle).}
\label{tab:sample_efficiency}
\centering
\begin{small}
\begin{sc}
\resizebox{1.0\textwidth}{!}{
\begin{tabular}{lc|cccc|cccc}
\toprule
Env & Percentile & DrQ-v2 & SVEA & SRM & SimGRL & DreamerV3 & HRSSM & Dr.~G & VIGOR (Ours) \\
\midrule
\midrule
\multirow[c]{3}{*}{DMC}
 & $25\%$ & $772$ & $493$ & $623$ & $625$ & $437$ & $802$ & $415$ & \cellcolor{cyan!20}{$\mathbf{395}$} \\
 & $50\%$ & $883$ & $605$ & $702$ & $642$ & $550$ & $833$ & $540$ & \cellcolor{cyan!20}{$\mathbf{412}$} \\
 & $75\%$ & $895$ & $640$ & $722$ & $657$ & $553$ & $847$ & $617$ & \cellcolor{cyan!20}{$\mathbf{422}$} \\
\midrule
\multirow[c]{3}{*}{Robosuite}
 & $25\%$ & $1{,}335$ & $1{,}450$ & $1{,}240$ & $1{,}985$ & $1{,}840$ & $2{,}000$ & $1{,}685$ & \cellcolor{cyan!20}{$\mathbf{655}$} \\
 & $50\%$ & $1{,}360$ & $1{,}700$ & $1{,}340$ & $1{,}990$ & $1{,}840$ & $2{,}000$ & $1{,}925$ & \cellcolor{cyan!20}{$\mathbf{720}$} \\
 & $75\%$ & $1{,}365$ & $1{,}780$ & $1{,}370$ & $1{,}990$ & $1{,}845$ & $2{,}000$ & $1{,}965$ & \cellcolor{cyan!20}{$\mathbf{740}$} \\
\bottomrule
\end{tabular}
}
\end{sc}
\end{small}
\end{table}

\begin{figure}[ht]
\centering
\includegraphics[width=\textwidth]{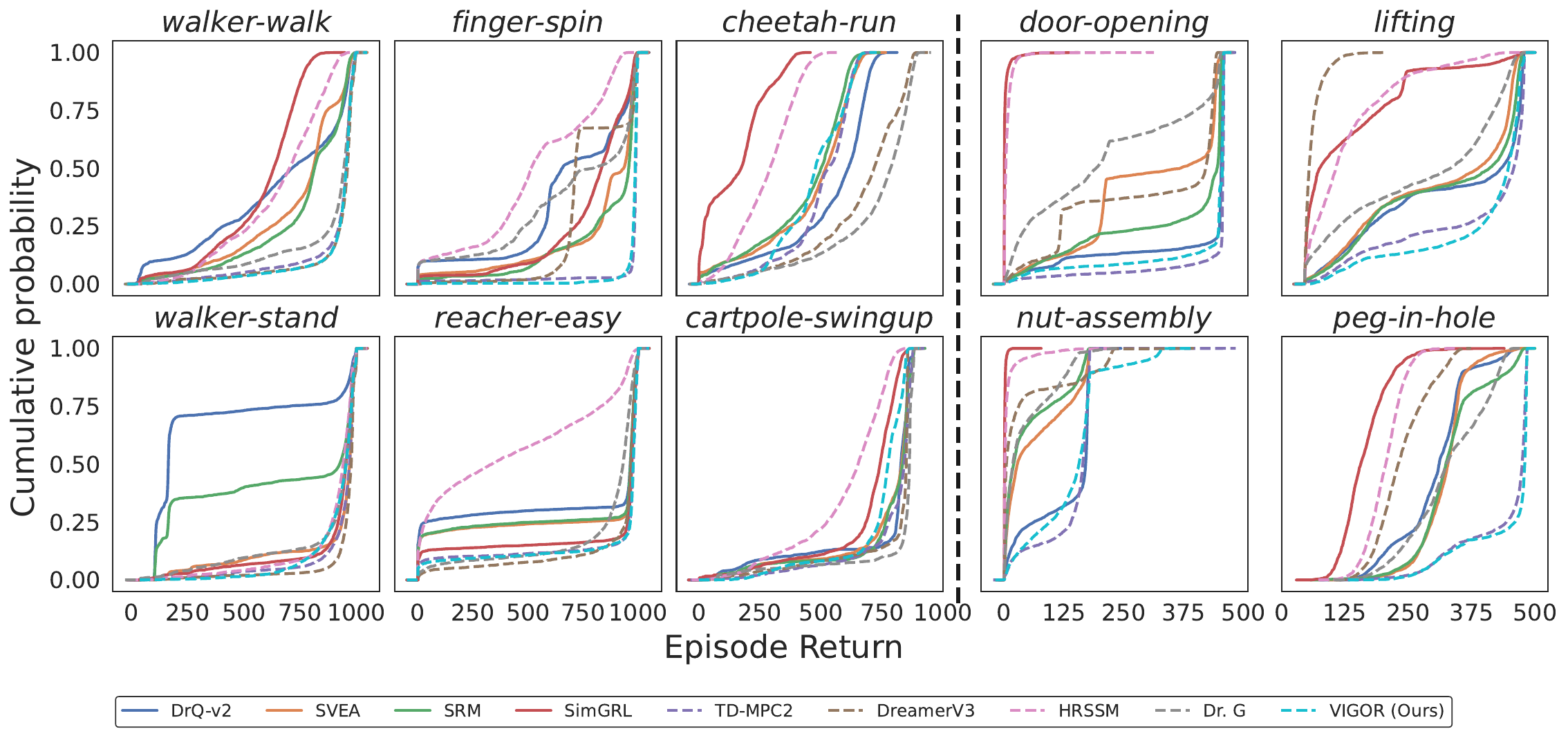}
\captionsetup{font=small}
\caption{\textbf{Statistical comparison of sample efficiency.} ECDFs of episode returns across tasks, computed from training statistics. These distributions complement the threshold-based comparisons in Table~\ref{tab:sample_efficiency}.}
\label{fig:sample_efficiency_over_tasks}
\end{figure}

\paragraph{Statistical analysis.}
Figure~\ref{fig:sample_efficiency_over_tasks} presents empirical cumulative distribution functions (ECDFs) of episode returns across tasks. ECDFs capture the full distribution of training efficiency and illustrate how quickly each method reaches target performance thresholds. VIGOR exhibits ECDF profiles that closely track TD-MPC2 while surpassing all other baselines, further validating its efficiency--robustness balance.

\paragraph{Summary.}
These results demonstrate that AW, DC, and ES preserve the sample efficiency of latent-space MBRL while enabling robust zero-shot generalization, incurring only modest overhead relative to the TD-MPC2 backbone.

\subsection{DMC 100K}\label{sec:appendix_dmc_100k}
The main experiments (Section~\ref{sec:results}) train all methods for 1M gradient steps, a budget sufficient for most methods to reach asymptotic performance. To stress-test VIGOR's efficiency--robustness trade-off under severe data constraints, we conduct an additional evaluation on DMC with only 100K gradient steps (10\% of the standard budget), following prior work~\citep{kaiser2019model,laskin2020curl,schwarzer2020data,ye2021mastering,schwarzer2023bigger,alonso2024diffusion}.

\paragraph{Motivation.} 
Evaluating sample efficiency requires assessing two conditions: (i)~how \textit{quickly} an agent reaches its asymptotic performance, and (ii)~how \textit{high} that asymptote is. Under the full 1M-step budget, most methods converge, making it difficult to distinguish whether strong generalization stems from robust representations or simply from prolonged training. The 100K budget isolates early-phase learning dynamics: methods that build robust latent structures efficiently should generalize well even before convergence, while methods that rely on extended training to compensate for fragile representations will degrade disproportionately. Furthermore, zero-shot generalization performance is bounded by the training-time return, so any efficiency gap directly tightens the robustness ceiling, exacerbating the efficiency--robustness trade-off.

\paragraph{Experimental setup.} 
All methods are trained for 100K gradient steps on six DMC tasks under the same RL-ViGen protocol used in the main experiments (Section~\ref{sec:exp_setup}). Hyperparameters are identical to the 1M-step setting; no method-specific tuning is performed for the reduced budget. We evaluate both sample efficiency (IQM normalized scores as a function of training steps) and zero-shot generalization (IQM normalized scores under unseen visual distractions after 100K steps). All results are averaged over ten random seeds.

\begin{figure}[ht]
\centering
\includegraphics[width=1.0\textwidth]{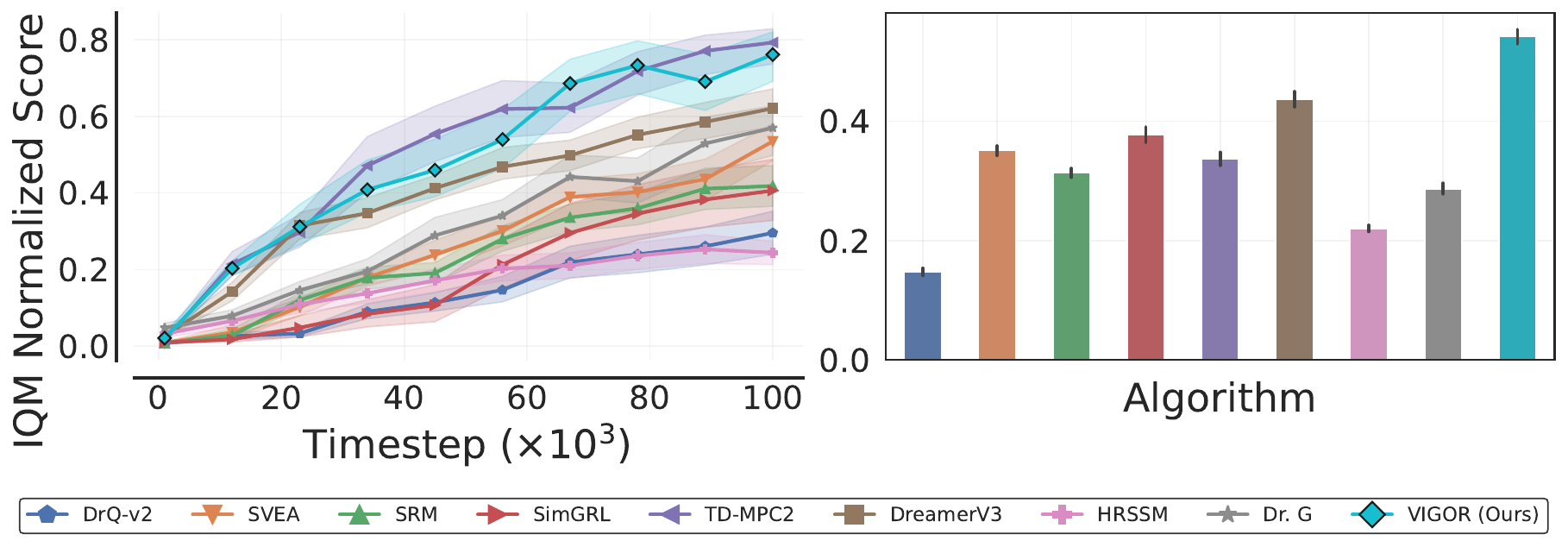}
\captionsetup{font=small}
\caption{\textbf{DMC 100K results.} Left: Sample efficiency (IQM normalized scores) under a 100K-step training budget. Right: Zero-shot generalization (IQM) after 100K steps.}
\label{fig:dmc_100k}
\end{figure}

\paragraph{Results.}
Figure~\ref{fig:dmc_100k} reports sample-efficiency curves (left) and zero-shot generalization (right).
\begin{itemize}
  \item \textbf{Sample efficiency.}\quad TD-MPC2 achieves the highest IQM normalized score (${\approx}0.80$), with VIGOR closely following (${\approx}0.78$). DreamerV3 ranks third (${\approx}0.61$), followed by Dr.~G (${\approx}0.58$) and SVEA (${\approx}0.56$). Two patterns are notable. First, the strongest sample-efficient methods are dominated by MBRL backbones (TD-MPC2, VIGOR, DreamerV3, and Dr.~G), confirming that latent-space world models provide an efficiency advantage even under severe data constraints. However, not all MBRL methods benefit equally: HRSSM (${\approx}0.25$) falls below all MFRL baselines under this tight budget, suggesting that its more complex world-model objective (bisimulation-based hybrid RSSM) is harder to optimize under limited interaction. Second, TD-MPC2 and VIGOR substantially outpace DreamerV3, Dr.~G, and HRSSM, highlighting the efficiency of TD-MPC2's MPC-based planning over recurrent world-model rollouts. VIGOR closely matches TD-MPC2's trajectory throughout training despite the addition of AW, DC, and ES, indicating that these components introduce minimal training overhead (Appendix~\ref{sec:rebuttal_wall-clock}).
  \item \textbf{Zero-shot generalization.}\quad VIGOR achieves the strongest zero-shot generalization, substantially outperforming all baselines. The contrast with the sample-efficiency ranking is striking: TD-MPC2, the most efficient method, drops to mid-tier generalization performance. This confirms that sample efficiency alone does not guarantee robustness: without explicit consistency mechanisms, the representations learned under a constrained budget remain fragile to unseen visual shifts. VIGOR closes this gap: it retains near-TD-MPC2 efficiency while achieving the highest generalization, demonstrating that latent-space consistency is established early in training rather than requiring extended training to converge.
\end{itemize}

\paragraph{Summary.}
These results demonstrate that VIGOR's efficiency--robustness advantage is not an artifact of extended training. Even under a 10$\times$ reduction in training budget, VIGOR achieves the strongest trade-off between sample efficiency and zero-shot generalization, confirming that latent-space consistency is built into the world-model representations from the early stages of training.

\subsection{Wall-Clock Training Efficiency}\label{sec:rebuttal_wall-clock}
To complement the sample-efficiency analysis, we evaluate VIGOR's wall-clock training efficiency, which is important for assessing practical usability.

\paragraph{Experimental setup.}
VIGOR and TD-MPC2 are trained under an identical RL-ViGen protocol on a single NVIDIA A5000 (24\,GB) GPU, using the same dataloaders, environment settings, and hyperparameters described in Appendices~\ref{sec:appendix_implementation_details} and~\ref{sec:appendix_evaluation_details}.

\begin{figure}[ht]
\centering
\includegraphics[width=1.0\textwidth]{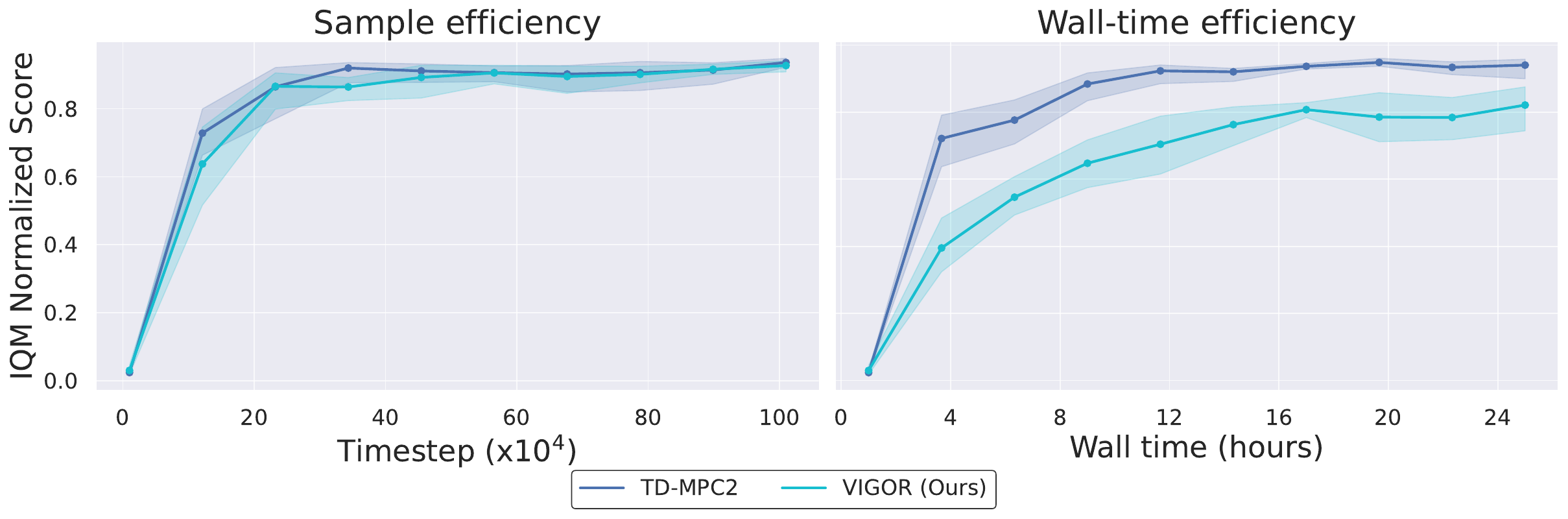}
\captionsetup{font=small}
\caption{\textbf{Sample efficiency and wall-clock training efficiency.} Left: Learning curves as a function of environment steps. Right: Learning curves as a function of wall-clock time.}
\label{fig:wall_time}
\end{figure}

\paragraph{Overall results.}
As shown in Figure~\ref{fig:wall_time} (left), VIGOR closely matches TD-MPC2 in sample efficiency: IQM normalized scores as a function of environment steps are nearly identical, and both methods reach asymptotic performance at similar sample budgets. Figure~\ref{fig:wall_time} (right) analyzes performance as a function of wall-clock time aggregated across all tasks. TD-MPC2 trains faster during the early stages (the gap peaks at $0.37$ around hours 3--4), but the difference narrows as training progresses. Over a 24-hour window, the average normalized-score gap is approximately $0.17$, and it stays within $0.06$--$0.14$ after hour $14$.

\paragraph{Wall-clock overhead analysis.}
The wall-clock overhead of VIGOR is modest and stems primarily from strong augmentations, which introduce additional preprocessing (\eg \textit{background-video} composition and external image loading). VIGOR mitigates this cost through its batch-splitting design: AW partitions a single batch into weak-only and weak-to-strong subsets rather than creating duplicate views, and all augmentations are applied \emph{before} encoding. As a result, the encoder processes the same total number of samples as in the original TD-MPC2 pipeline, and the DC and ES losses add only lightweight computations on top of the existing latent representations. This design explains why VIGOR incurs only a minor slowdown despite incorporating substantially stronger generalization mechanisms.

\begin{figure}[ht]
\centering
\includegraphics[width=1.0\textwidth]{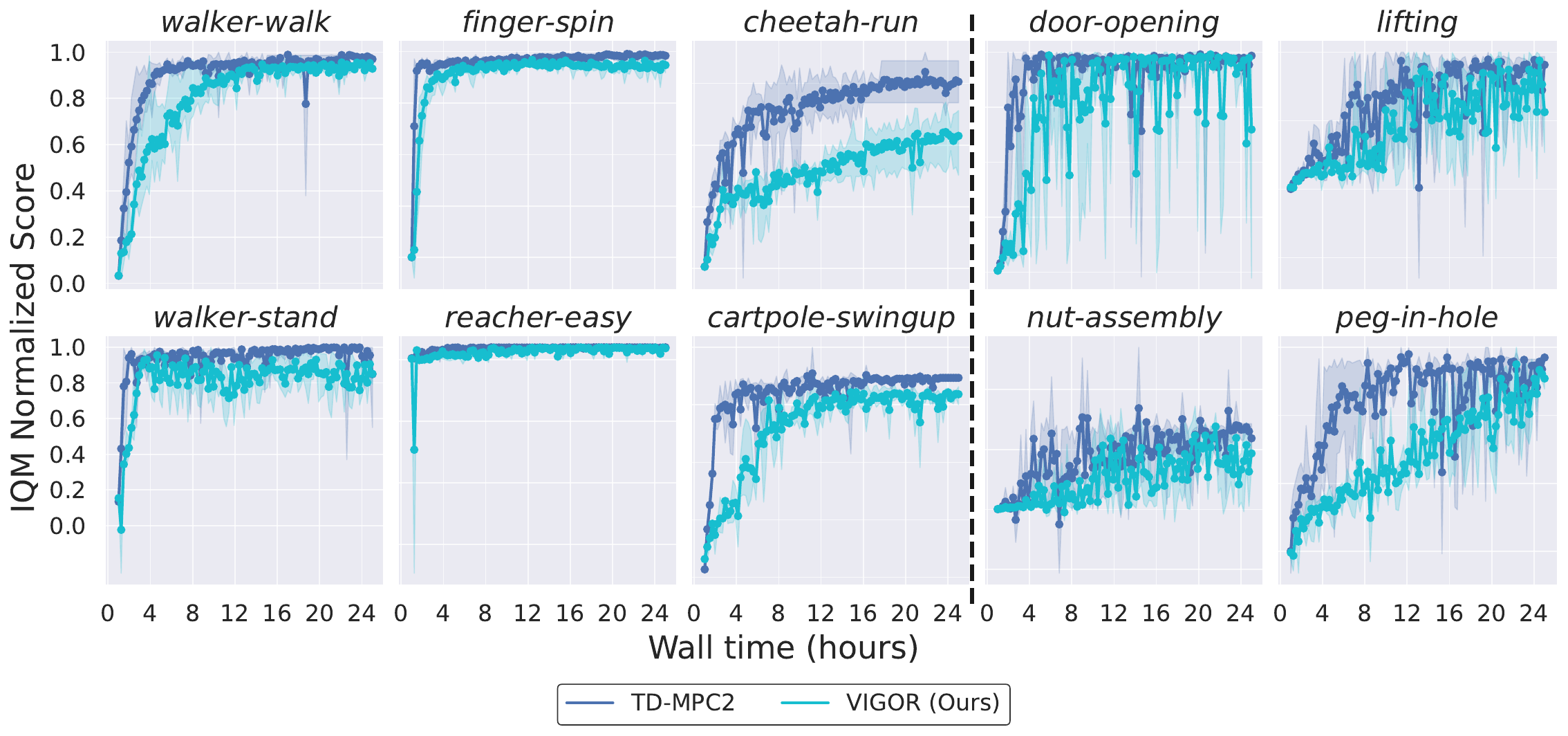}
\captionsetup{font=small}
\caption{\textbf{Per-task wall-clock training efficiency.} IQM normalized scores as a function of wall-clock time for TD-MPC2 and VIGOR across all DMC and Robosuite tasks.}
\label{fig:wall_time_over_tasks}
\end{figure}

\paragraph{Per-task analysis.}
Figure~\ref{fig:wall_time_over_tasks} provides a per-task breakdown across all ten environments. TD-MPC2 maintains a small speed advantage in wall-clock time, but VIGOR remains close throughout training and the difference narrows as both methods converge. Table~\ref{tab:wall_time_efficiency} summarizes the IQM normalized-score differences (TD-MPC2 minus VIGOR) at fixed wall-clock checkpoints. The differences are largest in the first several hours and diminish toward later stages (at hour $14$ the gap drops to $0.06$), confirming that VIGOR's overhead becomes less pronounced over time.

\begin{table}[ht]
\captionsetup{font=small}
\caption{\textbf{Wall-clock training efficiency.} Difference in IQM normalized scores (TD-MPC2 minus VIGOR) at fixed wall-clock checkpoints across all tasks. Values near zero indicate matched training speed; positive values indicate a slight advantage for TD-MPC2.}
\label{tab:wall_time_efficiency}
\centering
\begin{small}
\begin{sc}
\resizebox{0.60\textwidth}{!}{
\begin{tabular} {c|c}
\toprule
Wall-clock time (hours) & IQM normalized score difference \\
\midrule
$1$ & $-0.01$ \\
$2$ & $0.24$ \\
$3$ & $0.37$ \\
$4$ & $0.37$ \\
$5$ & $0.26$ \\
$6$ & $0.20$ \\
$7$ & $0.24$ \\
$8$ & $0.26$ \\
$9$ & $0.19$ \\
$10$ &$0.15$ \\
$11$ &$0.24$ \\
$12$ &$0.12$ \\
$13$ &$0.18$ \\
$\mathbf{14}$ &$\mathbf{0.06}$ \\
$15$ &$0.10$ \\
$16$ &$0.07$ \\
$17$ &$0.08$ \\
$18$ &$0.12$ \\
$19$ &$0.14$ \\
$20$ &$0.13$ \\
$21$ &$0.13$ \\
$22$ &$0.10$ \\
$23$ &$0.12$ \\
$24$ &$0.12$ \\
\midrule
Average & $0.17$ \\
\bottomrule
\end{tabular}}
\end{sc}
\end{small}
\end{table}

\paragraph{Summary.}
VIGOR introduces only a modest wall-clock overhead relative to TD-MPC2, which diminishes as training progresses. These results confirm that VIGOR preserves practical training efficiency while delivering substantially improved zero-shot robustness.

\subsection{Augmentation Ablation Studies}\label{sec:appendix_augmentation_ablation}
This subsection provides supplementary analyses for the augmentation ablations in Sections~\ref{sec:core_components} and~\ref{sec:design_choices}, which respectively present the AW ablation results (zero-shot generalization in Table~\ref{tab:ablation_mixed}; sample efficiency in Figure~\ref{fig:ablation_mixed_augmentations}) and the augmentation-family experiment (Figure~\ref{fig:aug_family_results}). Here, we focus on patterns not discussed in the main text: per-difficulty performance and variance characteristics of each AW variant, the $\zeta$-sweep's cross-benchmark dynamics, sample-efficiency convergence patterns, and a qualitative comparison of the augmentation families used in the augmentation-agnostic experiment.

\paragraph{Per-difficulty and variance analysis of AW variants.}
The main text reports aggregate performance for each variant (Table~\ref{tab:ablation_mixed}). Here, we examine the per-difficulty and variance patterns that reveal \emph{how} each variant fails.
\begin{itemize}
  \item \textbf{DMC.}\quad
    The gap between variants widens from easy to hard. \texttt{SO} achieves $423.8 \pm 421.9$ (easy) and $374.9 \pm 401.0$ (hard), with standard deviations nearly equal to the means throughout, indicating that learning is highly unstable and seed-dependent. \texttt{WTSO} shows a similar pattern ($550.6 \pm 377.6$ easy; $481.3 \pm 358.3$ hard) with large variance, confirming that persistent strong exposure without a stable anchor produces unreliable training. \texttt{WO}, by contrast, maintains lower relative variance ($773.9 \pm 246.0$ easy; $606.6 \pm 326.8$ hard) and converges fast (Figure~\ref{fig:ablation_mixed_augmentations}, left), but its generalization ceiling is bounded by limited perturbation diversity. VIGOR achieves both the highest returns and the lowest variance at both levels ($863.9 \pm 155.4$ easy; $781.1 \pm 261.2$ hard), demonstrating that AW's asymmetric splitting stabilizes optimization while maintaining robustness.

  \item \textbf{Robosuite.}\quad
    The per-difficulty breakdown reveals a sharper separation, with all variants exhibiting monotonic degradation from easy to extreme. \texttt{WO} achieves moderate returns at the easy level ($134.2 \pm 143.8$) but degrades steadily under harder shifts ($112.6 \pm 118.0$ hard; $93.4 \pm 96.9$ extreme), confirming that weak augmentations alone provide insufficient robustness in visually complex manipulation. \texttt{SO} achieves higher easy-level returns ($187.1 \pm 182.2$) but follows the same degradation pattern ($137.3 \pm 137.6$ hard; $121.3 \pm 122.6$ extreme), with standard deviations consistently matching or exceeding the means, indicating highly unstable learning. \texttt{WTSO} improves at the easy level ($297.6 \pm 155.2$) but similarly degrades under harder shifts ($138.1 \pm 131.3$ hard; $130.1 \pm 120.7$ extreme; $79\%$ overall). Only AW maintains the highest returns at every difficulty level ($332.0$ easy; $236.7$ hard; $146.0$ extreme) with the strongest overall performance.
\end{itemize}

The cross-benchmark contrast is instructive: on DMC, \texttt{WO} preserves training efficiency but under-generalizes; on Robosuite, \texttt{WO} fails entirely because the visual complexity of manipulation scenes demands stronger perturbation exposure. Only AW's asymmetric splitting simultaneously provides both signals: stability from the weak-only partition and diversity from the weak-to-strong partition.

\paragraph{Mixture-ratio sensitivity ($\zeta$-sweep).} 
The mixture ratio $\zeta = |\mathcal{I}^w|/|\mathcal{I}^{ws}|$ controls the balance between weak-only and weak-to-strong samples within each batch. Table~\ref{tab:ablation_mixed} shows that $\zeta{=}4$, $\zeta{=}2$, and $\zeta{=}0.5$ all underperform $\zeta{=}1$ on average, but the nature of the degradation differs across benchmarks:
\begin{itemize}
  \item On DMC, increasing the weak-only fraction improves performance ($\zeta{=}4$: $96\%$ $>$ $\zeta{=}2$: $94\%$ $>$ $\zeta{=}0.5$: $88\%$). Per-difficulty data show that this advantage stems primarily from the easy level, where $\zeta{=}4$ ($854.4 \pm 176.5$) nearly matches VIGOR ($863.9 \pm 155.4$). At the hard level, the gap widens ($736.3 \pm 274.3$ vs.\ $781.1 \pm 261.2$), indicating that excessive weak-only bias sacrifices robustness under stronger shifts.
  \item On Robosuite, the trend reverses: $\zeta{=}0.5$ ($87\%$) outperforms $\zeta{=}2$ ($80\%$) and $\zeta{=}4$ ($75\%$). The advantage of $\zeta{=}0.5$ is most pronounced at the hard level ($172.2 \pm 157.4$ vs.\ $144.7 \pm 144.5$ for $\zeta{=}2$ and $151.8 \pm 156.4$ for $\zeta{=}4$), confirming that severe shifts demand stronger perturbation exposure. Nevertheless, $\zeta{=}0.5$ still falls short of $\zeta{=}1$ at every level, indicating that excessive strong exposure impairs optimization even when robustness demands are high.
\end{itemize}

These results confirm that the balanced mixture ($\zeta{=}1$) is a deliberate design choice. The optimal ratio reflects a trade-off between training stability (favoring more weak-only samples) and generalization coverage (favoring more weak-to-strong samples), with $\zeta{=}1$ providing the best balance across both benchmarks.

\paragraph{Sample efficiency of AW variants.} 
Figure~\ref{fig:ablation_mixed_augmentations} reveals convergence patterns that complement the generalization analysis.

For the AW variants (left panel), AW and \texttt{WO} converge at comparable rates, confirming that the weak-only partition in AW preserves training stability. \texttt{SO} exhibits wide confidence intervals throughout training and fails to reach meaningful returns, while \texttt{WTSO} converges but to a substantially lower asymptote.

For the $\zeta$-sweep (right panel), all variants converge to similar asymptotic levels, but $\zeta{=}0.5$ exhibits slower early-phase convergence due to its higher proportion of strongly perturbed samples. These patterns reinforce the main-text finding that AW uniquely balances efficiency and robustness: alternative strategies sacrifice one or the other.

\begin{figure}[ht]
\centerline{\includegraphics[width=1.0\textwidth]{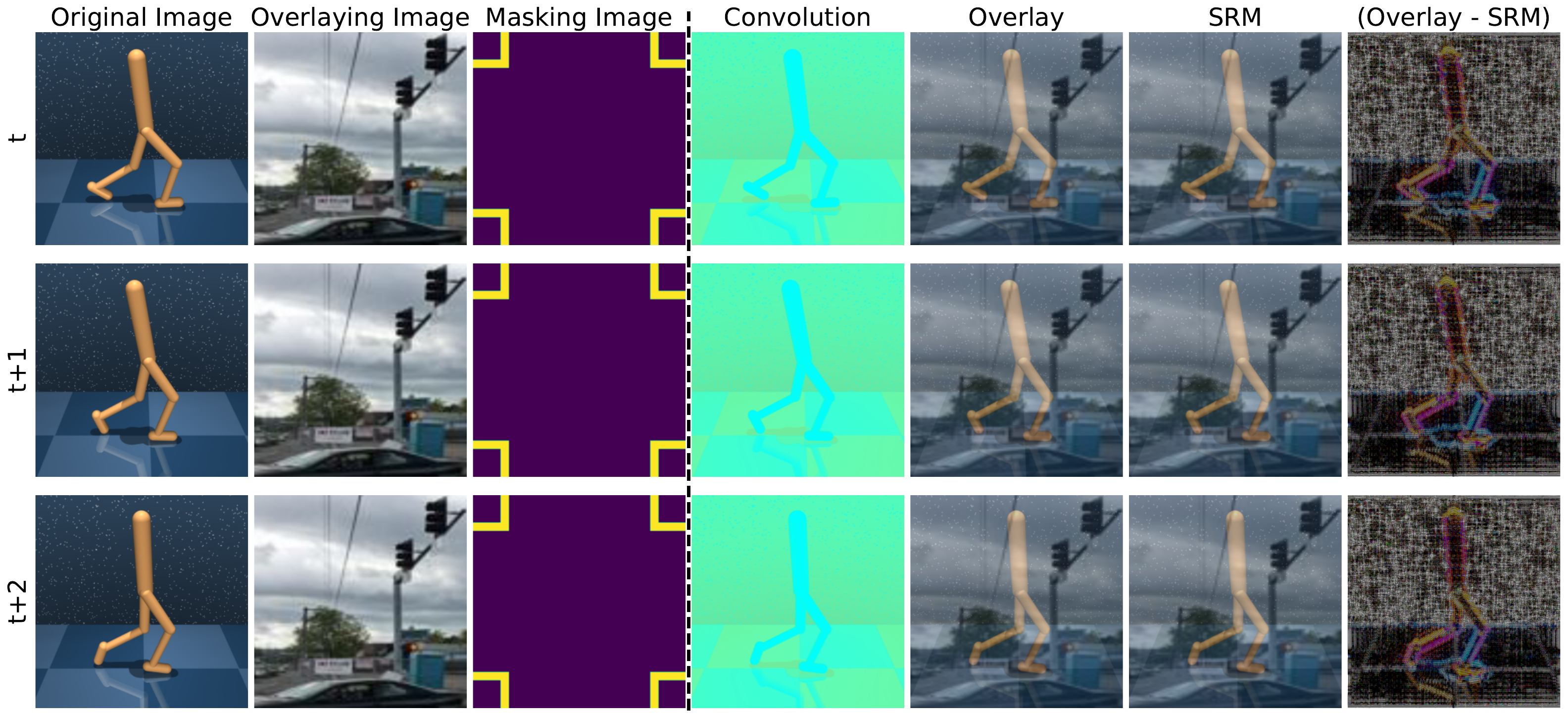}}
\captionsetup{font=small}
\caption{\textbf{Strong augmentation families.} Three augmentation types applied to \textit{walker-walk}: \textit{random-overlay} (contextual background replacement), \textit{random-conv} (local texture and color perturbation), and \textit{SRM} (Fourier-domain frequency masking). The rightmost column shows the absolute pixel-wise difference between \textit{random-overlay} and \textit{SRM}, illustrating their distinct spectral characteristics.}
\label{fig:aug_family_diff}
\end{figure}

\paragraph{Qualitative comparison of augmentation families.} 
The augmentation-agnostic experiment in Section~\ref{sec:design_choices} replaces the default strong augmentation (\textit{random-overlay}) with \textit{random-conv} and \textit{SRM}, two families from distinct perturbation domains. Figure~\ref{fig:aug_family_diff} visualizes the qualitative differences on \textit{walker-walk}:

\begin{itemize}
  \item \textit{Random-overlay} blends a randomly sampled natural image into the background, introducing contextual replacement that alters the global scene appearance while preserving the agent's shape and texture.
  \item \textit{Random-conv} applies a random convolution kernel to each frame, producing spatially smooth color and texture shifts without modifying scene structure. The perturbation is local and frequency-preserving.
  \item \textit{SRM (Spectrum Random Masking)} masks contiguous bands in the Fourier spectrum, removing mid- and high-frequency image components. The resulting perturbation is global and frequency-destructive, qualitatively distinct from both overlay and convolution.
\end{itemize}

The rightmost column of Figure~\ref{fig:aug_family_diff} shows the absolute pixel-wise difference between \textit{random-overlay} and \textit{SRM}, highlighting their distinct spectral characteristics. Despite these fundamental differences, all three variants maintain strong zero-shot generalization (Figure~\ref{fig:aug_family_results}), confirming that latent-space consistency, not the augmentation choice, is the source of robustness.

\paragraph{Summary.} 
These supplementary analyses reinforce three conclusions from the main text. First, AW's asymmetric batch splitting is the only augmentation strategy that preserves both training stability and zero-shot robustness across benchmarks with fundamentally different visual complexities; per-difficulty and variance analysis reveals that alternative strategies fail through either instability (SO, WTSO) or insufficient diversity (WO). Second, the balanced mixture ratio ($\zeta{=}1$) reflects a principled trade-off between stability and generalization coverage, with the optimal direction shifting across benchmarks but $\zeta{=}1$ providing the best compromise at every difficulty level. Third, VIGOR's robustness is augmentation-agnostic: three qualitatively distinct perturbation families (contextual, textural, and spectral) all yield strong generalization, confirming that latent-space consistency, not the augmentation, drives robustness.

\subsection{Latent-Space Consistency Analysis: DC and ES Ablation}\label{sec:appendix_embedding_vis_ablation}
To complement the quantitative ablation in Section~\ref{sec:core_components} (Table~\ref{tab:ablation_components}), we visualize how removing DC and ES affects the structure of latent rollouts under visual perturbations. This analysis isolates the qualitative role of each component beyond aggregate return statistics.

\begin{figure}[ht]
    \centering
    \captionsetup[sub]{font=small}
    \subfigure[\textit{background-color}]{\centering\includegraphics[width=1.0\textwidth]{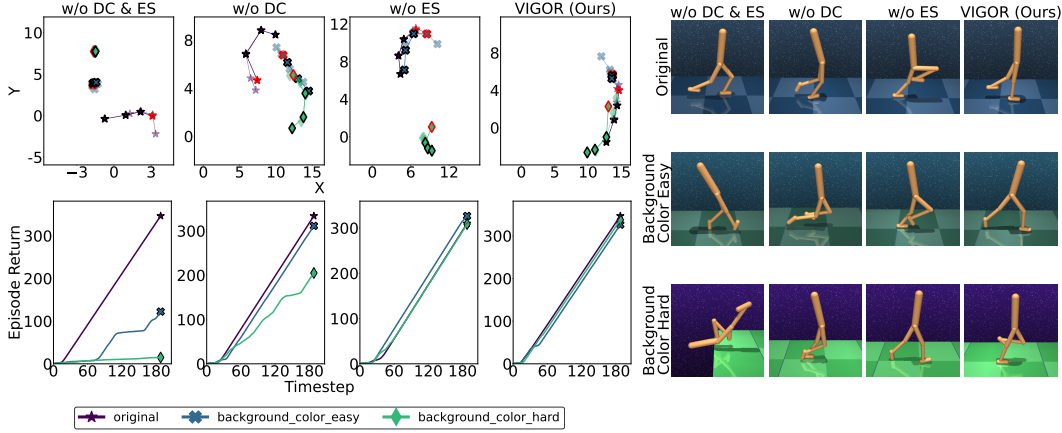}
    \label{fig:appendix_visualization_embeddings_ablation_a}}
    \\
    \subfigure[\textit{object-color}]{\centering\includegraphics[width=1.0\textwidth]{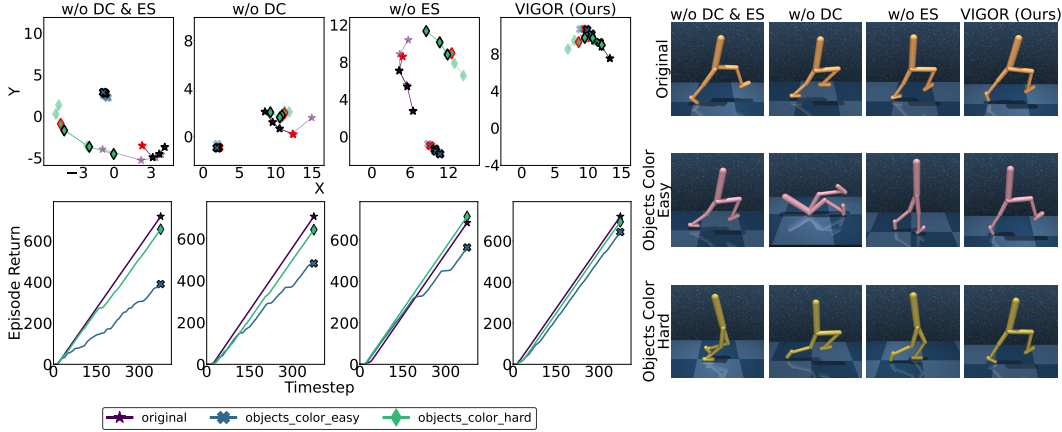}
    \label{fig:appendix_visualization_embeddings_ablation_b}}
    \captionsetup{font=small}
    \caption{\textbf{Latent-space consistency: ablation on DC and ES.} Left: UMAP projections of predicted latent states (top) and episode returns (bottom) for \textit{walker-walk} under \textit{background-color} and \textit{object-color} distractions, comparing VIGOR with three ablated variants (w/o~DC \&~ES, w/o~DC, w/o~ES). Right: Environment snapshots across difficulty levels. Markers denote difficulty ({\color[RGB]{100,50,140}$\bigstar$}:~\textit{original}, {\color[RGB]{40,80,215}$\bigtimes$}:~\textit{easy}, {\color[RGB]{34,180,55}$\Diamond$}:~\textit{hard}); arrows indicate temporal progression. Red-edged markers indicate the initial latent state $z_t$ encoded from the observation; subsequent markers correspond to rolled-out states $z_{t+1:t+H}$ produced by the dynamics model.}
    \label{fig:appendix_visualization_embeddings_ablation}
\end{figure}

\paragraph{Experimental setup.}
We compare the full VIGOR with three ablated variants (w/o~DC \&~ES, w/o~DC, and w/o~ES) on \textit{walker-walk} under \textit{background-color} and \textit{object-color} distractions. For each variant, we roll out the learned world model from identical initial inputs and project the resulting latent states into 2D via UMAP~\citep{mcinnes2018umap}: the initial state $z_t$ is encoded directly from the observation, while the subsequent states $z_{t+1:t+H}$ are produced by the dynamics model. To make this distinction visible, we display $z_t$ with red-edged markers in Figure~\ref{fig:appendix_visualization_embeddings_ablation}. Episode returns and environment snapshots are shown alongside.

\paragraph{Results.}
Figure~\ref{fig:appendix_visualization_embeddings_ablation} reveals distinct failure modes for each ablation that align with the quantitative patterns in Table~\ref{tab:ablation_components}.
\begin{itemize}
  \item \textbf{w/o~DC \&~ES.} Without either consistency mechanism, latent trajectories scatter widely even at the encoder level: the initial states $z_t$ under \textit{easy} and \textit{hard} settings are far from those under \textit{original}, and the rollouts diverge sharply. The return curves drop substantially under perturbations, with episode failures visible in the snapshots.

  \item \textbf{w/o~DC.} ES alone aligns the encoder representations to some extent, but the dynamics rollouts diverge across difficulty levels. Although the initial states $z_t$ are partially clustered, the dynamics model produces inconsistent multi-step predictions, illustrating that encoder-level alignment alone cannot prevent the dynamics from amplifying residual discrepancies across recursive rollouts.

  \item \textbf{w/o~ES.} DC alone yields more compact rollout trajectories than the previous two variants, but the latent geometry is shifted across difficulty levels: the initial states $z_t$ under perturbations are visibly displaced from their clean counterparts. While returns are higher than the previous variants, the misaligned encoder representations limit the benefit of DC's dynamics-level constraint.

  \item \textbf{VIGOR (full).} Only the full model maintains both encoder-level alignment (initial states $z_t$ across difficulty levels are tightly clustered) and dynamics-level consistency (subsequent rollouts $z_{t+1:t+H}$ remain aligned across the full horizon). This two-level stability translates directly into stable returns and successful task execution across all evaluated difficulty levels.
\end{itemize}

\paragraph{Summary.}
These visualizations corroborate the quantitative ablation (Table~\ref{tab:ablation_components}): ES alone leaves the rollouts divergent, DC alone leaves the initial encodings displaced, and only their combination yields the two-level stability that underlies VIGOR's robust returns.

\newpage
\subsection{Ablation Study on Design Choices}\label{sec:appendix_design_choices}
This subsection provides formal definitions of the ablated variants introduced in Section~\ref{sec:design_choices} and detailed results across distraction types, tasks, and efficiency metrics.

\paragraph{Dynamic vs.\ consistent augmentations.} 
VIGOR applies augmentations at every step of the rollout horizon. For a trajectory segment $x_{t:t+H}$, the augmented views are:
\begin{align} 
x_i^{w} = \tau^{w}(x_i, \upsilon_i^{w}), \quad
x_i^{ws} = \tau^{s}(\tau^{w}(x_i, \upsilon_i^{w}), \upsilon_i^{s}), \quad
i = t, \dots, t{+}H. 
\end{align} 
If $\upsilon_i^{w} = \upsilon_t^{w}$ and $\upsilon_i^{s} = \upsilon_t^{s}$ for all $i$, the scheme is \textit{consistent}: a single set of augmentation parameters is sampled and reused across all rollout steps. Otherwise, the scheme is \textit{dynamic}: new parameters are sampled independently at every step. VIGOR uses the dynamic setting by default; the consistent variant is \texttt{VIGOR\_CONST\_AUG}.

\begin{figure}[ht]
\centerline{\includegraphics[width=0.80\textwidth]{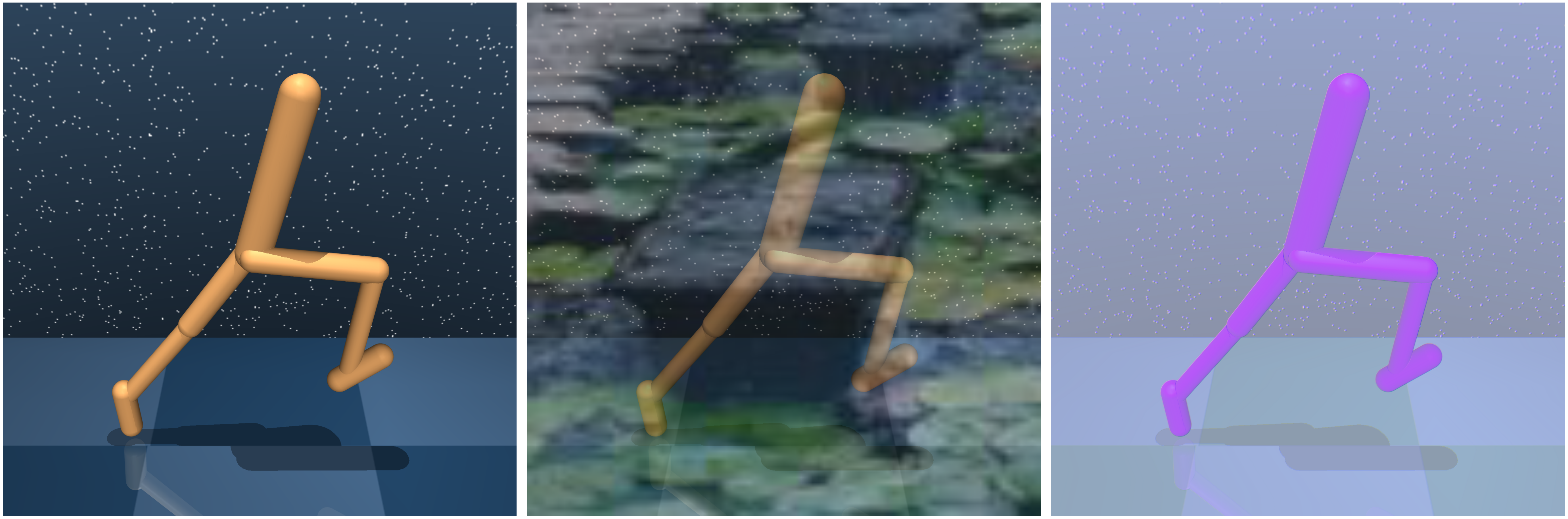}}
\captionsetup{font=small}
\caption{\textbf{Strong augmentation examples.} Visualization of augmentation effects in the \textit{walker-walk} task. Left: Original image. Center: \textit{random-overlay} used in VIGOR. Right: \textit{random-conv} used in \texttt{VIGOR\_CONV}.}
\label{fig:example_strong_aug}
\end{figure}

\paragraph{Choice of strong augmentation.} 
VIGOR uses \textit{random-overlay} by default. We compare against \textit{random-conv}: 
\begin{align} 
\tau^{s,\text{overlay}}(o, \tilde{o}) = (1{-}\delta)\,o + \delta\,\tilde{o}, \quad 
\tau^{s,\text{conv}}(o, w) = \mathrm{CONV}(o, w), 
\end{align}
where $o$ is the input image, $\tilde{o} \sim \mathcal{D}$ is a task-irrelevant overlay, $\delta = 0.5$, and $w \sim \mathcal{N}(0,1)$ is a random convolution kernel. Both transforms are applied to every frame in the stacked observation $x_t = \{o_t, o_{t-1}, \dots, o_{t-k+1}\}$. We adopt \textit{random-overlay} because it better approximates natural visual distractions at deployment, whereas \textit{random-conv} applies synthetic pixel-level perturbations. The variant with \textit{random-conv} is \texttt{VIGOR\_CONV}. Figure~\ref{fig:example_strong_aug} visualizes their qualitative differences.

\paragraph{Choice of auxiliary objective for ES.} 
VIGOR uses an alignment-based ES objective (adapted from SODA~\citep{hansen2021generalization}) by default. As an alternative, we consider a contrastive ES objective (adapted from CURL~\citep{laskin2020curl}):
\begin{align} 
  \mathcal{L}_{\text{contrast}}
  = -\log 
    \frac{\exp(q^{\top} W k_{+})} 
         {\exp(q^{\top} W k_{+}) 
          + \sum_{j=1}^{N} \exp(q^{\top} W k_{j})}, 
\end{align} 
where $q = h_\theta(\tau^w(x_t, \upsilon^q))$, $k_+ = h_{\theta^-}(\tau^w(x_t, \upsilon^k))$, $N$ is the mini-batch size (Section~\ref{sec:weak_strong}), and $W$ is a learnable bilinear projection. The variant replacing alignment-based ES with contrastive ES is \texttt{VIGOR\_CONTRAST}.

\begin{figure}[ht]
\centering
\includegraphics[width=1.0\textwidth]{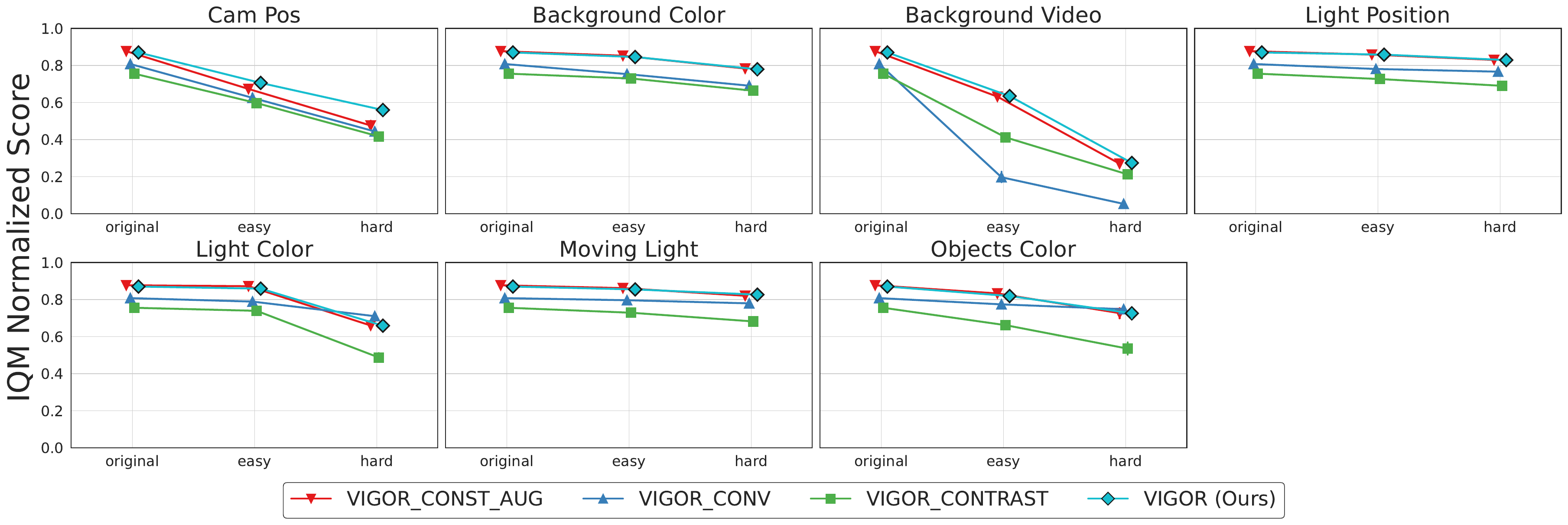}
\captionsetup{font=small}
\caption{\textbf{Design-choice ablation: zero-shot generalization across distraction types.} Layout follows Figure~\ref{fig:eval_results_over_eval_types}, comparing VIGOR against its ablated variants. VIGOR shows more stable performance across difficulty levels.}
\label{fig:eval_results_over_eval_types_VIGOR}
\end{figure}

\paragraph{Zero-shot generalization.}
Figure~\ref{fig:eval_results_over_eval_types_VIGOR} reports zero-shot generalization across all distraction types. VIGOR exhibits smoother and more monotonic degradation (\textit{original}~$\rightarrow$~\textit{easy}~$\rightarrow$~\textit{hard}) compared with the ablated variants, indicating stronger robustness to distribution shifts.

Among the variants, \texttt{VIGOR\_CONV} performs competitively on color-related distractions (\textit{background-color}, \textit{light-color}, \textit{object-color}), consistent with the color-invariance bias of random convolution. However, its performance drops sharply on \textit{background-video}, where perturbations alter spatial and contextual cues rather than color. \texttt{VIGOR\_CONTRAST} consistently underperforms across most distraction types.

\begin{figure}[ht]
    \centering
    \includegraphics[width=1.0\textwidth]{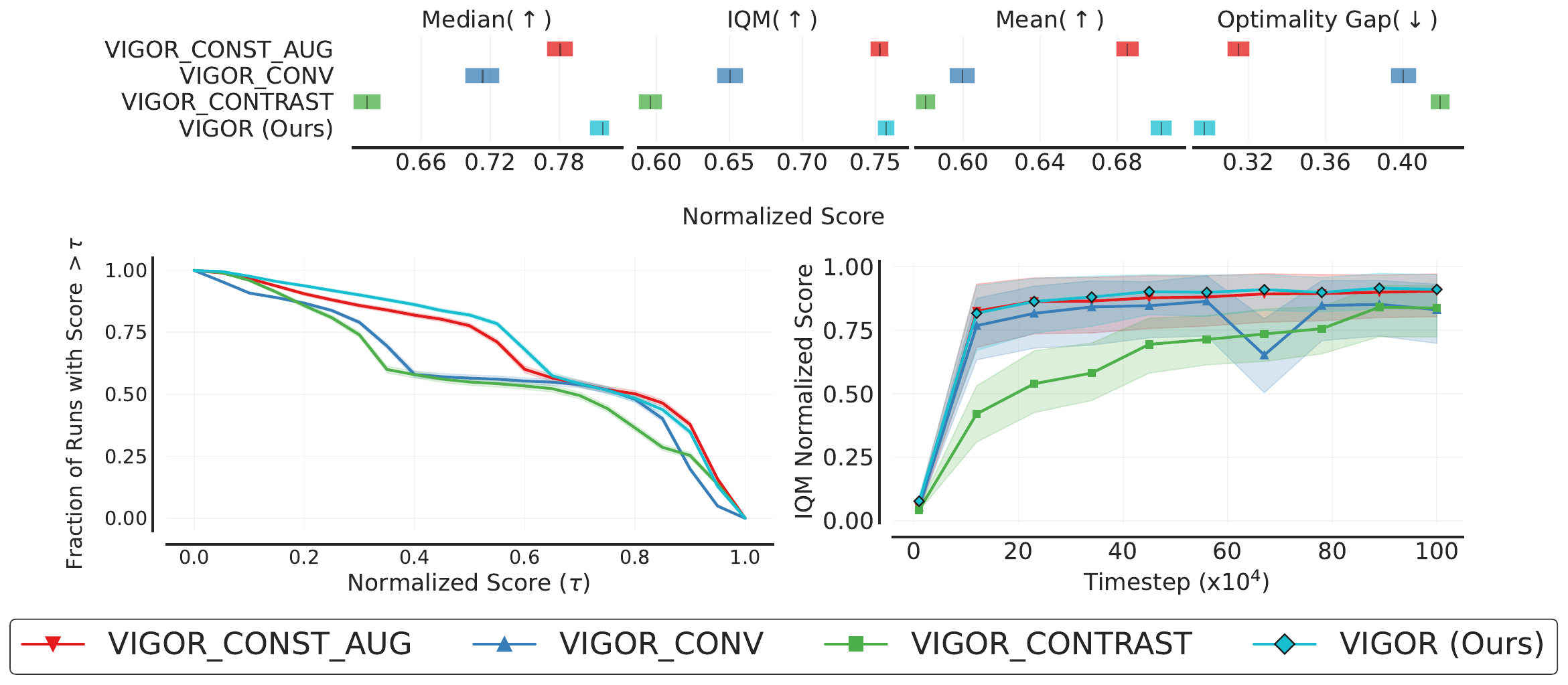}
    \captionsetup{font=small}
    \caption{\textbf{Design-choice ablation: aggregate metrics and sample efficiency.} Top: Zero-shot generalization (median, IQM, mean, optimality gap). Bottom left: Performance profile. Bottom right: Learning curves. Shaded regions denote 95\% stratified bootstrap confidence intervals.}
    \label{fig:app_design_choices}
\end{figure}

\paragraph{Sample efficiency and aggregate metrics.}
Figure~\ref{fig:app_design_choices} provides a comprehensive comparison. The aggregate statistics (top row) show that VIGOR achieves the best median, IQM, mean, and optimality gap, confirming its overall advantage. The performance profile (bottom-left) shows VIGOR's dominance across the full distribution of normalized scores, and the learning curve (bottom-right) shows that VIGOR retains the sample efficiency of TD-MPC2 despite training under visual perturbations.

\paragraph{Qualitative comparison of strong augmentations.}
Figure~\ref{fig:example_strong_aug} illustrates the visual effects of the two strong augmentations on \textit{walker-walk}. \textit{Random-overlay} blends natural images into the background, introducing realistic domain shifts that approximate variations encountered in deployment. \textit{Random-conv} applies convolutional filters that predominantly alter color statistics without structural changes. This distinction explains why \texttt{VIGOR\_CONV} performs well on color-based perturbations but fails to generalize to shifts involving spatial or contextual variations.

\paragraph{Summary.}
These ablations confirm that each design choice (dynamic augmentation, the choice of strong augmentation, and the auxiliary loss formulation) contributes meaningfully to VIGOR's generalization. Altering any one of them leads to clear performance degradation, demonstrating that these choices collectively enable robust zero-shot generalization.

\subsection{Latent-Space Consistency Analysis}\label{sec:appendix_embedding_vis}
This subsection provides detailed embedding visualizations that complement the latent-space analysis in Section~\ref{sec:embedding_vis}. 
We examine whether the two-level consistency observed in the main text, at both the encoder and rollout levels, generalizes across multiple distraction types.

\paragraph{Experimental setup.}
We compare VIGOR with its backbone TD-MPC2~\citep{hansen2023td} and Dr.~G~\citep{ha2023dream}, the most directly comparable prior method for zero-shot visual generalization in latent-space MBRL. All agents are trained in clean environments and evaluated \textit{zero-shot} under unseen visual distractions. For each agent, we roll out the learned world model from identical initial inputs to obtain a predicted latent trajectory $z_{t:t+H} = (z_t, z_{t+1}, \dots, z_{t+H})$, where each subsequent state is generated by the dynamics model as $z_{t+h+1} = d_\theta(z_{t+h}, a_{t+h})$, and we record the episode return accumulated during evaluation. To visualize how each world model evolves over consecutive timesteps, we additionally depict the preceding rollout $z_{t-1:t+H-1}$ as a lighter trajectory, providing temporal context. Comparing the two trajectories reveals whether the world model maintains consistent rollouts as the encoder processes successive observations under visual perturbations.

\paragraph{Results.}
We project the high-dimensional latent trajectories into 2D via UMAP~\citep{mcinnes2018umap}, with faded markers indicating earlier time steps. Episode returns and environment snapshots across evaluation levels (\textit{original}, \textit{easy}, \textit{hard}) are shown alongside.

\begin{figure}[ht]
    \centering
    \captionsetup[sub]{font=small}
    \subfigure[\textit{background-color}]{
    \centering
    \includegraphics[width=1.0\textwidth]{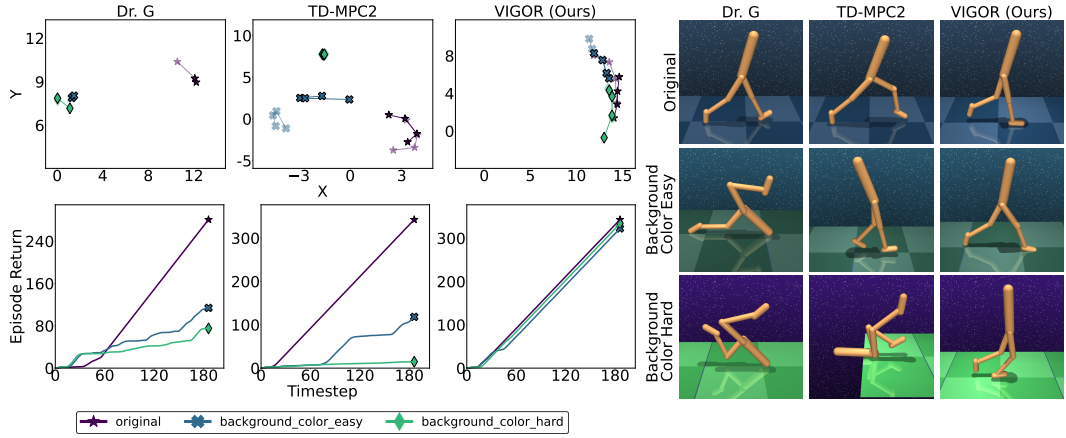}}
    \label{fig:appendix_visualization_embeddings_a}
    \\
    \subfigure[\textit{background-video}]{
    \centering
    \includegraphics[width=1.00\textwidth]{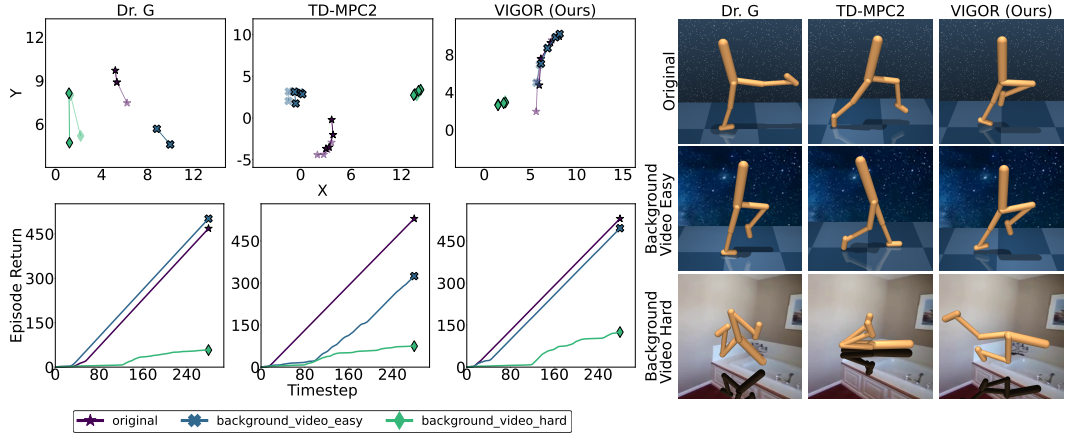}}
    \label{fig:appendix_visualization_embeddings_b}
    \\
    \subfigure[\textit{light-color}]{
    \centering
    \includegraphics[width=1.0\textwidth]{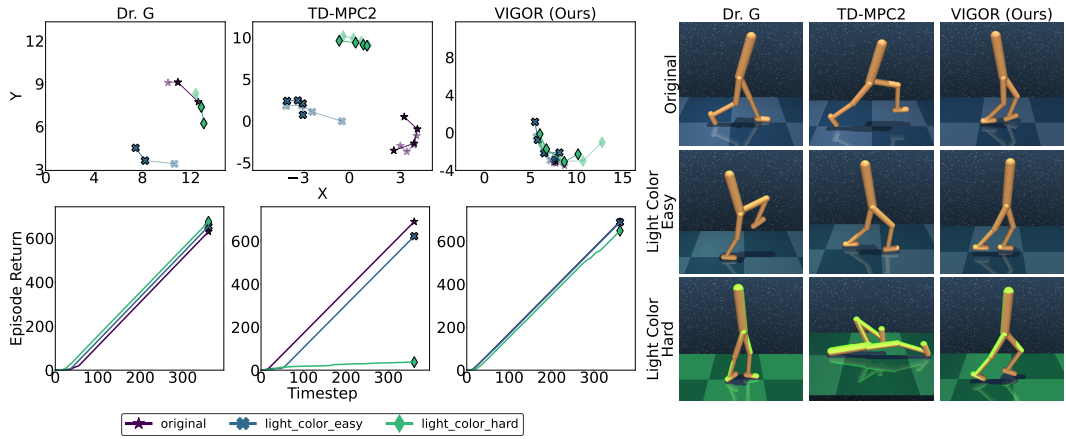}}
    \label{fig:appendix_visualization_embeddings_c}
    \captionsetup{font=small}
    \caption{\textbf{Latent-space consistency under visual perturbations.} Left: UMAP projections of predicted latent states (top) and episode returns (bottom) for \textit{walker-walk} under three distraction types, including \textit{background-color}, \textit{background-video}, and \textit{light-color}. Right: Environment snapshots across settings. Markers denote difficulty ({\color[RGB]{100,50,140}$\bigstar$}:~\textit{original}, {\color[RGB]{40,80,215}$\bigtimes$}:~\textit{easy}, {\color[RGB]{34,180,55}$\Diamond$}:~\textit{hard}); arrows indicate temporal progression. VIGOR maintains a consistent latent manifold and stable performance across levels, whereas TD-MPC2 and Dr.~G exhibit divergent rollouts and degraded execution.}
    \label{fig:appendix_visualization_embeddings}
\end{figure}

Figure~\ref{fig:appendix_visualization_embeddings} presents results for \textit{walker-walk} under three perturbation types: \textit{background-color}, \textit{background-video}, and \textit{light-color}. Across all three types, VIGOR exhibits consistency at both levels of the vulnerability identified in Figure~\ref{fig:distribution_shift}. At the encoder level, the initial latent states extracted under \textit{original}, \textit{easy}, and \textit{hard} settings are located within the same region of the latent manifold, indicating that the encoder maps visually distinct observations to similar task-relevant representations. At the rollout level, the subsequent latent trajectories $z_{t+1:t+H}$ produced by the dynamics model remain tightly aligned across difficulty levels, confirming that the dynamics model preserves this consistency throughout the planning horizon. This two-level stability translates directly into stable episode returns across all settings.

In contrast, TD-MPC2 and Dr.~G exhibit divergent patterns at both levels. Their initial encoder outputs scatter under perturbations, and this divergence is amplified through latent rollouts, producing trajectories that deviate substantially from the clean setting. The resulting return degradation and task failures (visible in both the episode-return curves and the environment snapshots) illustrate the two-level vulnerability in practice.

\paragraph{Summary.}
These visualizations provide further empirical support for the hypothesis (Figure~\ref{fig:distribution_shift}) that mapping OOD inputs onto the task-relevant latent manifold is essential for zero-shot generalization. Across all three perturbation types, VIGOR consistently preserves consistency at both the encoder and rollout levels, a property absent in the baselines.

\clearpage
\subsection{Performance under Longer Planning Horizons}\label{sec:rebuttal_horizons}
Our default experiments adopt a short MPC planning horizon of $H{=}3$, following standard practice in latent-space MPC. To examine whether VIGOR's latent-consistency mechanism remains effective under deeper imagined rollouts, we conduct comparisons at extended planning horizons.

\paragraph{Experimental setup.}
Longer planning horizons substantially increase wall-clock time because each optimization step requires deeper latent rollouts. Running $H{=}5$ or $H{=}10$ across all DMC and Robosuite tasks would therefore incur prohibitive computational cost. We thus evaluate $H \in \{3, 5, 10\}$ on one representative task per benchmark: \textit{finger-spin} on DMC and \textit{door-opening} on Robosuite. All other hyperparameters remain fixed to isolate the effect of the planning horizon.

\begin{figure}[ht]
\centering
\includegraphics[width=1.0\textwidth]{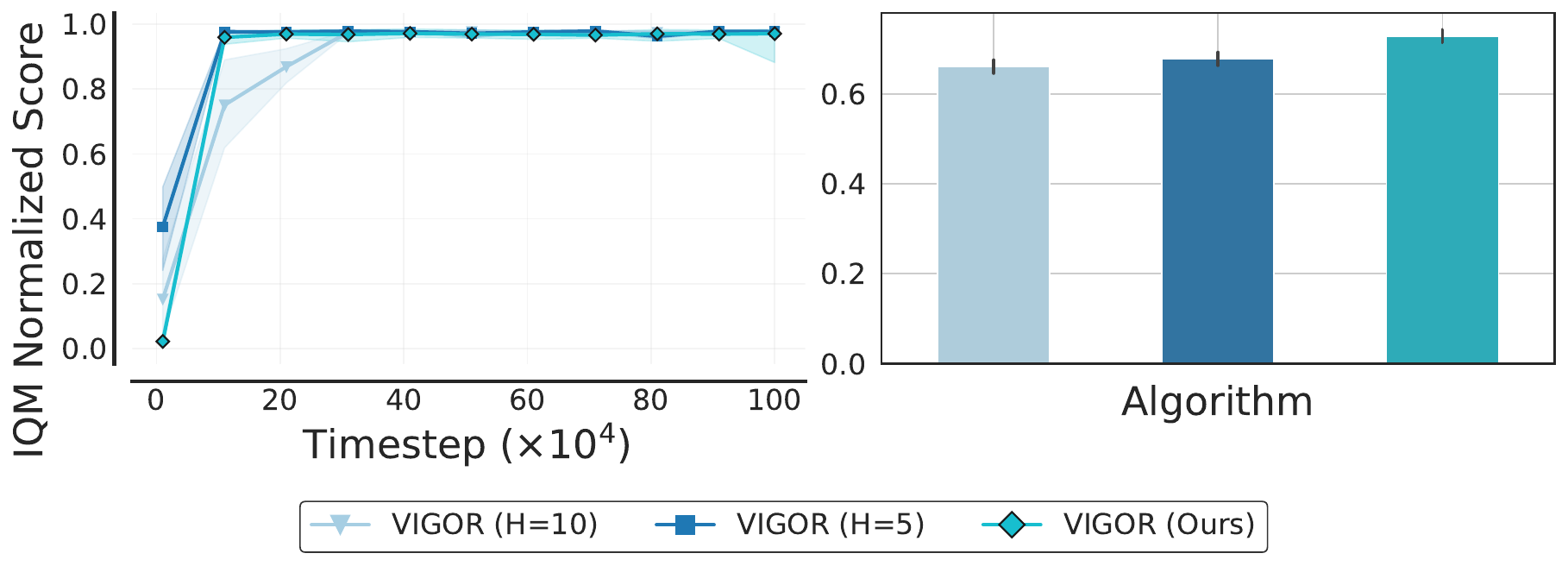}
\captionsetup{font=small}
\caption{\textbf{Performance under extended planning horizons.} Learning curves for VIGOR on \textit{finger-spin} (DMC) and \textit{door-opening} (Robosuite) with $H \in \{3, 5, 10\}$. VIGOR maintains stable learning and reaches comparable asymptotic returns across all horizons, despite slower convergence at longer horizons.}
\label{fig:horizon_results}
\end{figure}

\paragraph{Results and discussion.}
Figure~\ref{fig:horizon_results} shows learning curves for VIGOR under the three horizons. Across $H{=}3$, $H{=}5$, and $H{=}10$, VIGOR maintains stable optimization and reaches similar asymptotic returns. As expected, increasing the horizon slows convergence because compounding prediction errors accumulate over longer imagined trajectories, a well-documented limitation of MBRL. Nevertheless, VIGOR's latent-consistency regularization continues to stabilize the forward dynamics, preventing error accumulation from degrading policy learning even when the planning depth is tripled.

\paragraph{Summary.} 
These results confirm that VIGOR's latent-consistency mechanism generalizes beyond the default short-horizon setting. By mitigating the accumulation of prediction errors, VIGOR maintains stable learning and competitive performance under deeper imagined rollouts.

\subsection{Generalization to Unseen Camera Viewpoints}\label{sec:rebuttal_camera-viewpoints}
While RL-ViGen includes camera-position shifts (translations), these preserve the base camera orientation. To investigate whether models trained with appearance-focused perturbations can generalize to fundamentally different geometric transformations, we evaluate all methods on unseen camera rotations that modify projection geometry and the relative orientation between the agent and the camera.

\paragraph{Viewpoint-rotation settings.}
We consider two rotational viewpoint distractions from DMControl-GB~\citep{hansen2021generalization}:
\begin{itemize}
    \item \textbf{\texttt{cam-ori} (static rotation).}\quad At the start of each episode, the tracking camera's spherical parameters $c=(\phi, \theta, r, \theta_\text{roll})$ are sampled uniformly within bounds controlled by a difficulty parameter $\beta_\text{cam}\in \left[0,1\right]$ (easy: $0.3$; hard: $0.8$). Larger $\beta_\text{cam}$ permits wider rotation and distance ranges ( $-\frac{\pi}{2}\beta_\text{cam} \leq \phi,\theta,\theta_\text{roll} \leq \frac{\pi}{2}\beta_\text{cam}$ and $0.5\ r_\text{original}\beta_\text{cam} \leq r \leq 2.5\ r_\text{original}\beta_\text{cam}$), producing substantial viewpoint deviations without temporal variation.
    \item \textbf{\texttt{cam-dyn} (dynamic rotation).}\quad Camera parameters are initialized as in \texttt{cam-ori} and then evolve via a bounded random walk. Camera velocities are sampled as  
    \begin{equation}
    v_0\!\sim\!\mathcal{U}(v_{\min},v_{\max}),\quad 
    v_n = v_0 + \sum_{j=1}^{n}\mathcal{N}(0, \sigma\Sigma),
    \end{equation}
    where $n$ indexes the simulation step within the episode, $\sigma$ is a scalar noise scale, and $\Sigma$ is the per-axis covariance, generating continuous, temporally coherent viewpoint drift throughout the episode. Motion magnitudes scale with \(\beta_{\mathrm{cam}}\), producing time-varying geometric distortions.
\end{itemize}

\begin{figure}[ht]
    \centering
    \captionsetup[sub]{font=small}
    \subfigure[\textit{cartpole-swingup} under \texttt{cam-ori} and \texttt{cam-dyn}.]{
        \includegraphics[width=1.0\textwidth]{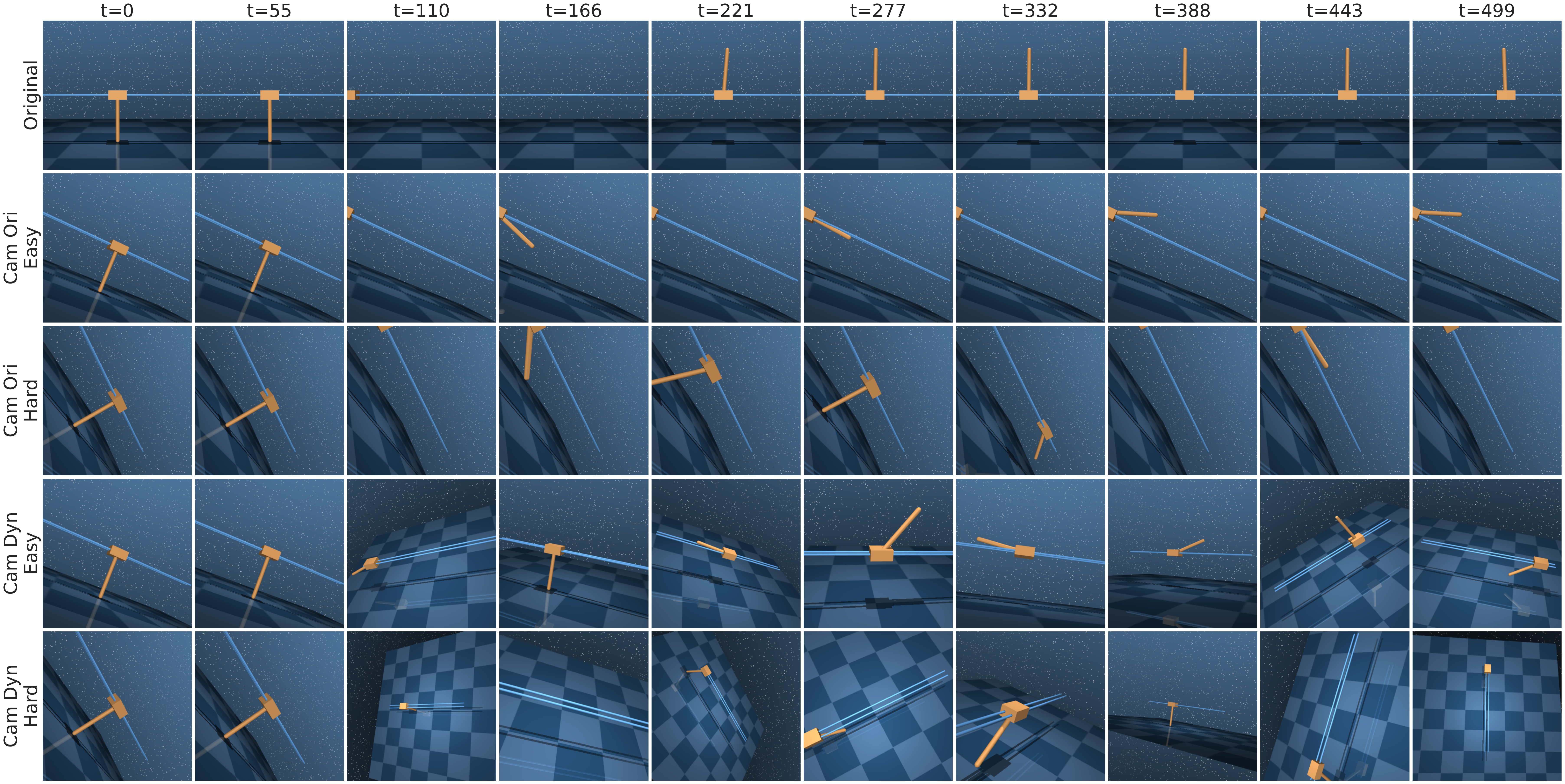}
        \label{fig:appendix_camera_viewpoint_comparison_a}
    }
    \\
    \subfigure[\textit{walker-walk} under \texttt{cam-ori} and \texttt{cam-dyn}.]{
        \includegraphics[width=1.0\textwidth]{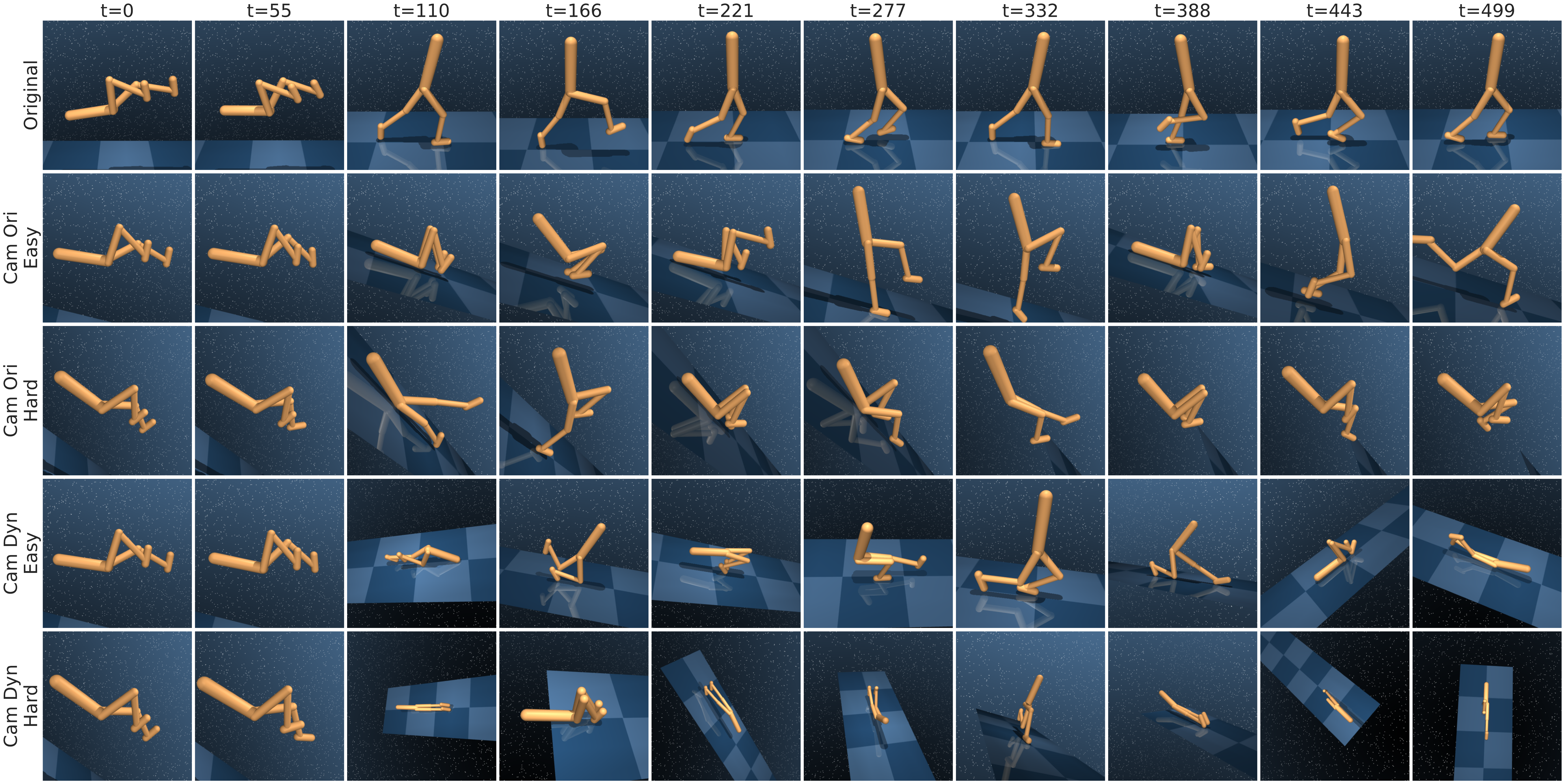}
        \label{fig:appendix_camera_viewpoint_comparison_b}
    }
        \caption{\textbf{Camera-viewpoint distractions.} Example frames from \textit{cartpole-swingup} and \textit{walker-walk} under \texttt{cam-ori} and \texttt{cam-dyn}. These settings introduce geometric distortions and projection changes that produce observations substantially different from training.}
    \label{fig:appendix_camera_viewpoint_comparison}
\end{figure}

As illustrated in Figure~\ref{fig:appendix_camera_viewpoint_comparison}, these viewpoint perturbations alter perspective, foreshorten limbs, reorder depth relationships, and in some cases cause the agent to leave the field of view entirely (\eg \textit{cartpole-swingup}).

\paragraph{Zero-shot evaluation.}
All models are trained under the RL-ViGen protocol (\emph{without} any viewpoint-rotation augmentations) and evaluated zero-shot on both settings.

\begin{figure}[ht]
\centering
\includegraphics[width=1.0\textwidth]{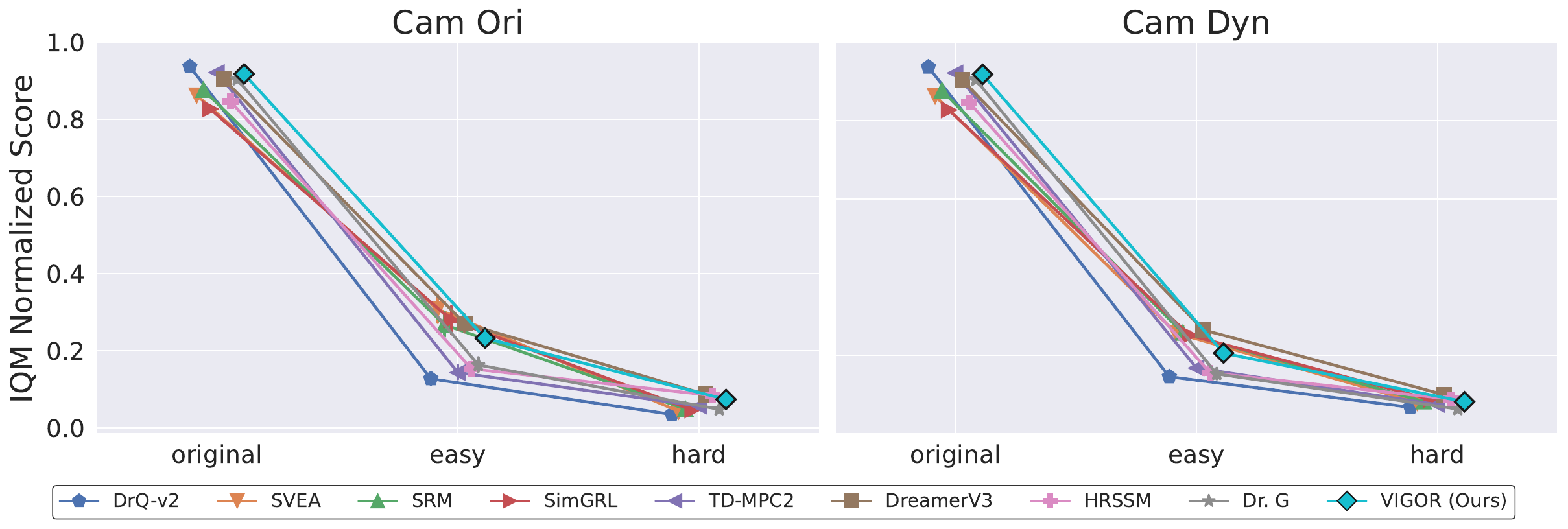}
\captionsetup{font=small}
\caption{\textbf{Zero-shot generalization under camera-viewpoint rotations.} VIGOR and baselines evaluated on \texttt{cam-ori} and \texttt{cam-dyn} settings. All models are trained without viewpoint-rotation augmentations. Performance drops sharply for all methods, indicating that zero-shot generalization under geometric viewpoint changes remains an open challenge.}
\label{fig:viewpoint_results}
\end{figure}

\paragraph{Results and discussion.}
As shown in Figure~\ref{fig:viewpoint_results}, performance drops sharply across all methods, and none maintains reliable zero-shot control under camera rotations. This degradation arises because rotational viewpoint changes induce geometric transformations that differ fundamentally from appearance-based perturbations (\eg lighting, background, or color). Such transformations violate the pixel-level and feature-level invariances implicitly learned by current encoder-based methods, including VIGOR.

While VIGOR is designed to improve robustness against appearance-based perturbations (background videos, lighting changes, color shifts, and camera translations), these results indicate that geometric viewpoint variations pose a more challenging generalization problem. Recent work has explored Spatial Transformer Networks (STNs)~\citep{jaderberg2015spatial} for few-shot viewpoint adaptation in visual RL~\citep{yang2024movie,yuan2024learning}, demonstrating improved spatial robustness by explicitly modeling spatial transformations. However, achieving simultaneous robustness to both appearance and geometric perturbations in a fully zero-shot setting remains largely unexplored. Developing geometry-aware latent representations, multi-view consistent world models, or scene-structured encoders constitutes an important direction for future work.

\paragraph{Summary.}
This evaluation reveals a clear limitation of current visual RL generalization methods. Although VIGOR significantly improves robustness to appearance-based perturbations, all evaluated methods struggle with the geometric projection changes induced by viewpoint rotation. These findings motivate future work on geometry-aware representations or multi-view consistent world models to achieve viewpoint-invariant control.

\section{MBRL Baseline Analysis}\label{sec:appendix_discussions}
This section provides an extended discussion of the MBRL baselines used in our experiments. Since VIGOR is designed as a framework that can be integrated into existing latent-space MBRL methods, understanding the characteristics of representative backbones is important. We focus on two influential families of latent-space MBRL, \textbf{Dreamer} and \textbf{TD-MPC}, which are widely regarded as state-of-the-art for pixel-based visuomotor control.

\paragraph{Dreamer family.}
The Dreamer family~\citep{hafner2019dream,hafner2020mastering,hafner2023mastering} learns a \emph{decoder-based} latent world model for long-horizon control from pixels. The model is trained on latent dynamics, reward prediction, and pixel-level reconstruction, forcing the latent state to retain sufficient information for observation recovery. The initial latent state is inferred through a recurrent state-space model (RSSM~\citep{hafner2019learning}) conditioned on past states, actions, and current observations. Actions are generated by a learned actor policy optimized via actor--critic RL in latent space, with the critic trained on imagined rollouts. DreamerV2~\citep{hafner2020mastering} and DreamerV3~\citep{hafner2023mastering} scale the architecture and introduce improved optimization strategies, making the family increasingly competitive on high-dimensional pixel-based control tasks. More recently, DreamerV4~\citep{hafner2025training} extends the family with a scalable Transformer architecture for long-horizon, language-conditioned tasks; however, it is trained from purely offline interaction data and thus falls outside the online RL setting considered in this work.

\paragraph{TD-MPC family.}
The TD-MPC family~\citep{hansen2022temporal,hansen2023td} takes a different approach by discarding reconstruction losses and adopting a \emph{decoder-free} latent world model. It jointly learns latent dynamics, reward, value, and a policy under a temporal-difference objective, operating entirely in latent space. Unlike the Dreamer family, which trains a separate actor for long-horizon imagination, the TD-MPC family integrates its policy into an online planning scheme: at each step, a model predictive controller searches over action sequences guided by Q-value predictions, using the learned policy as a prior, and executes the first action of the best trajectory. TD-MPC2 improves upon this formulation with architectural refinements and multi-task scalability, achieving strong sample efficiency on continuous visuomotor control benchmarks.

\paragraph{Comparison of the two families.}
Both families share the core principle of learning a transition model over compact latent representations. Their fundamental difference lies in how the world model is trained and how actions are selected: the Dreamer family relies on a decoder-based model and learns an actor policy via RL, while the TD-MPC family uses a decoder-free model and selects actions through MPC planning guided by Q-values. A further distinction concerns the role of imagined rollouts at deployment: the Dreamer family generates synthetic latent trajectories only during training for actor--critic optimization, while the actor directly selects actions at test time without rollouts. The TD-MPC family, by contrast, generates latent rollouts at both training and test time: during training for world-model learning and at test time for MPC planning. This distinction reflects two design philosophies, \emph{decoder-based actor learning} versus \emph{decoder-free latent planning}, with implications for sample efficiency and generalization.

\begin{figure}[ht]
\centering
\includegraphics[width=0.60\textwidth]{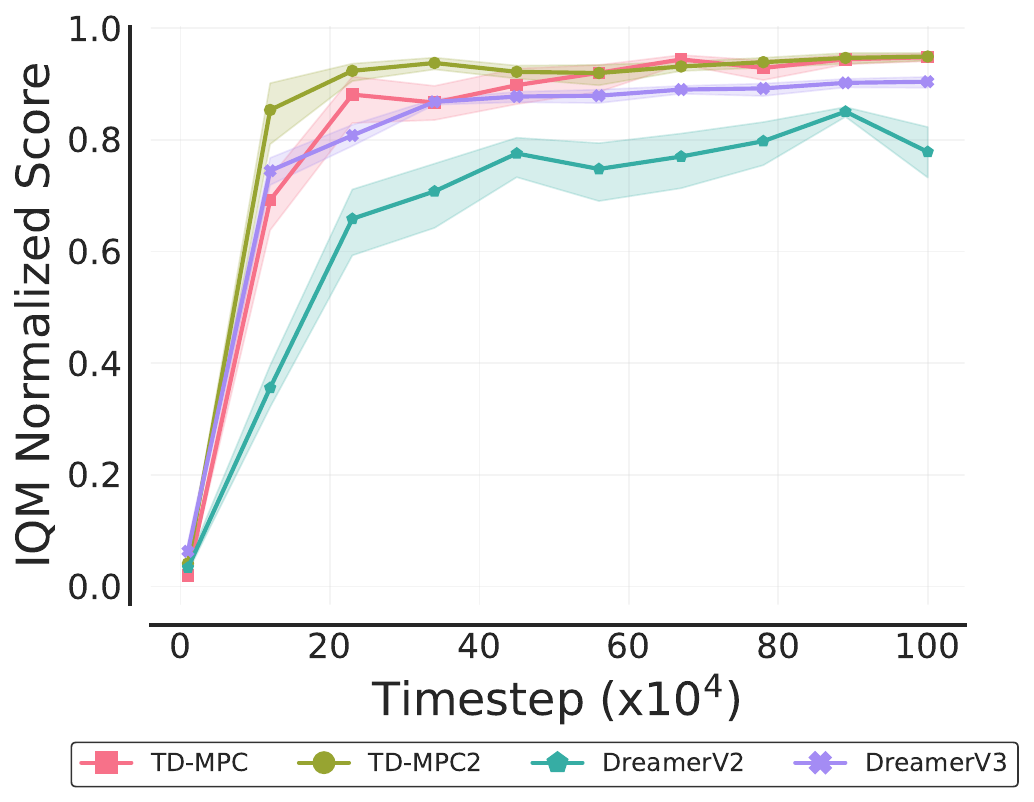}
\captionsetup{font=small}
\caption{\textbf{Sample efficiency of MBRL baselines.} TD-MPC2 achieves the highest sample efficiency across six DMC tasks, outperforming other MBRL baselines.}
\label{fig:tdmpc_dreamer_comparison}
\end{figure}

\paragraph{Rationale for backbone selection.}
Since VIGOR aims to enhance visual generalization without sacrificing sample efficiency, building upon an efficient backbone is essential. The Dreamer family's reconstruction-driven formulation increases model complexity and training cost, while the TD-MPC family's decoder-free design focuses directly on value learning and planning in latent space, yielding stronger sample efficiency in practice. As shown in Figure~\ref{fig:tdmpc_dreamer_comparison}, TD-MPC2 outperforms both the Dreamer family and its predecessor TD-MPC across diverse tasks in DMC, consistent with prior findings~\citep{hansen2023td}. We therefore adopt TD-MPC2 as VIGOR's backbone. Section~\ref{sec:results} demonstrates that VIGOR preserves this efficiency advantage while substantially improving zero-shot generalization. Nevertheless, to examine whether VIGOR's benefits depend on this choice, we additionally instantiate it on DreamerV3 in Appendix~\ref{sec:vigor_on_dreamerv3}.

\newpage
\section{VIGOR on DreamerV3}\label{sec:vigor_on_dreamerv3}
This section describes a second instantiation of VIGOR on DreamerV3~\citep{hafner2023mastering}, denoted DreamerV3+VIGOR, which examines whether the benefits of VIGOR extend beyond the TD-MPC2 backbone. It is evaluated under the same protocol as the main experiments (Tables~\ref{tab:dreamer_vigor_tasks}--\ref{tab:dreamer_vigor_tasks_backbones}; Figures~\ref{fig:dreamer_vigor_breakdown}--\ref{fig:eval_during_train_results_over_envs_dreamer_vigor_backbone}). Throughout this section, the default instantiation, referred to simply as VIGOR elsewhere in the paper, is denoted TD-MPC2+VIGOR. DreamerV3 provides a structurally distinct model-based RL backbone along several dimensions relevant to VIGOR. Unlike TD-MPC2~\citep{hansen2023td}, which learns a decoder-free implicit latent world model and selects actions through model-predictive control, DreamerV3 learns a generative world model built around a recurrent state-space model (RSSM) with stochastic categorical latents, trains its latent dynamics through KL-balanced dynamics and representation losses, and learns an actor--critic from imagined latent trajectories. We retain DreamerV3's architecture, observation reconstruction, reward, and continuation objectives, as well as its actor--critic learning procedure; the only change to the prediction objectives is that the decoder reconstructs the weakly augmented observation from the mixed-branch state (Algorithm~\ref{algo:dreamer_vigor}). VIGOR is incorporated only through the augmentation and consistency mechanisms described below: AW constructs the weak-only and weak-to-strong observation streams, DC re-targets the RSSM dynamics and representation KL terms across these streams, and ES adds an encoder-level consistency objective.

\subsection{Overview}
\begin{figure}[ht!]
    \centering
    \captionsetup[sub]{font=small}
    \subfigure[Overview of DreamerV3+VIGOR]{\centering\includegraphics[width=1.0\textwidth]{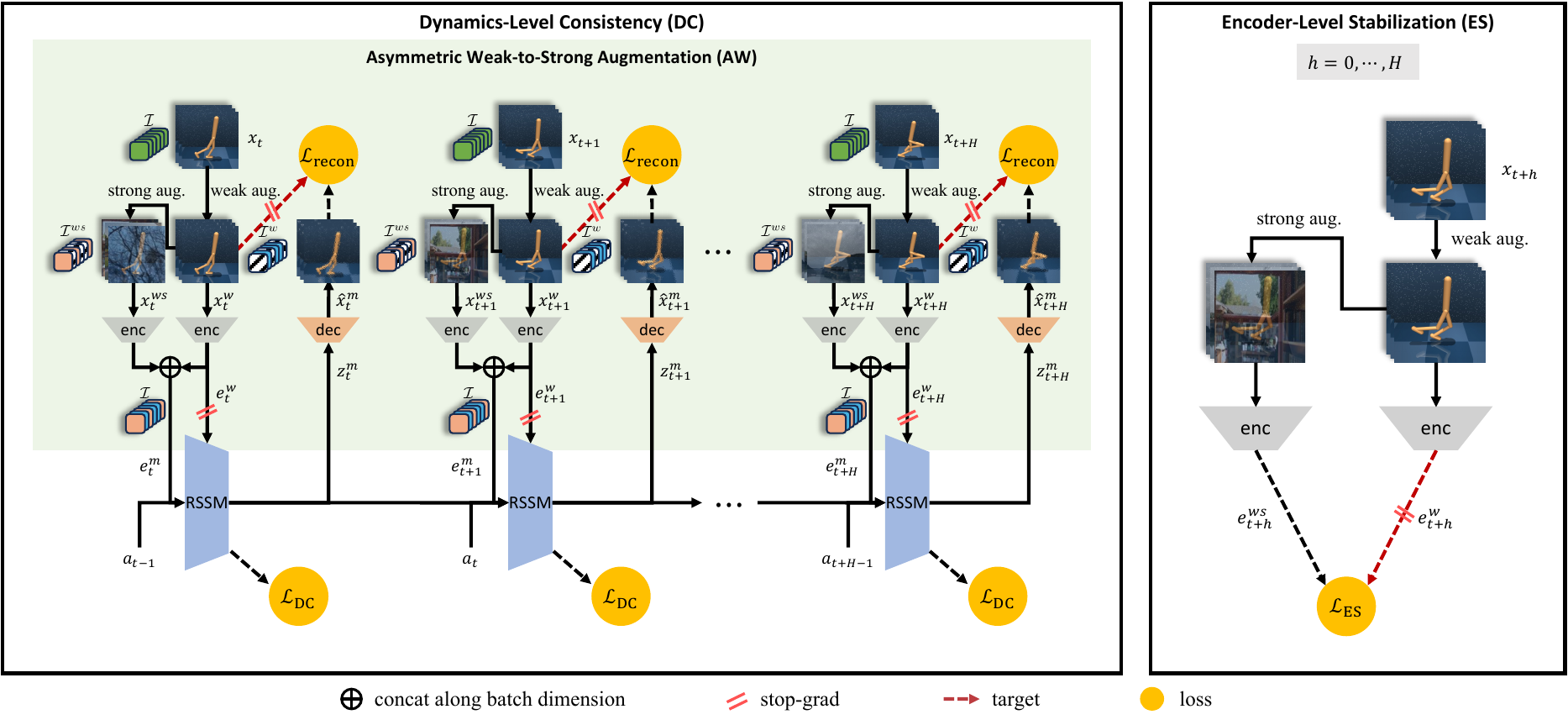}}
    ~~\subfigure[RSSM of DreamerV3+VIGOR]{\centering\includegraphics[width=1.0\textwidth]{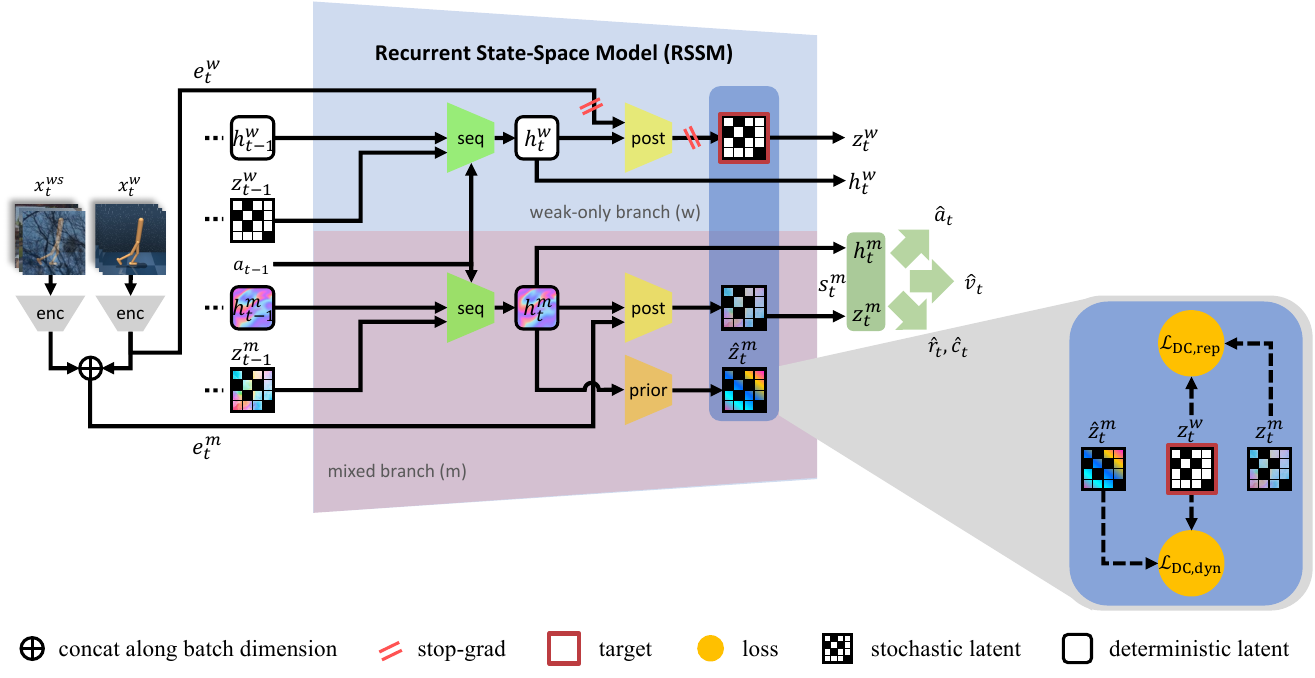}}
    \captionsetup{font=small}
    \caption{\textbf{Overview of integrating VIGOR with DreamerV3.} (a)~At every step of a sampled training sequence (steps $t,\dots,t+H$ are shown), AW constructs the weak-only and weak-to-strong views of the observation (subsets $\mathcal{I}^{w}$ and $\mathcal{I}^{ws}$, as in Figure~\ref{fig:VIGOR_architecture}) and concatenates them along the batch dimension; the shared encoder produces the mixed embedding $e_t^{m}$ and, under stop-gradient, the weak-only embedding $e_t^{w}$, both of which enter the RSSM together with the previous action $a_{t-1}$. The decoder reconstructs the weakly augmented observation from the mixed-branch state $s_t^{m}$ ($\mathcal{L}_{\mathrm{recon}}=-\ln p_\phi(x^w_t\mid h^m_t,z^m_t)$, the image term of $\mathcal{L}_{\mathrm{pred}}$ in Algorithm~\ref{algo:dreamer_vigor}), DC is applied inside the RSSM at every step ($\mathcal{L}_{\mathrm{DC}}$; detailed in (b)), and ES aligns the weak-to-strong embedding $e_{t+h}^{ws}$ with its weak-only counterpart $e_{t+h}^{w}$ at every step $h=0,\dots,H$. (b)~One time step of the RSSM, with the weak-only branch (top, stop-gradient) and the mixed branch (bottom). Each branch updates its recurrent state from its previous state and the action, $h_t=f_\phi(h_{t-1},z_{t-1},a_{t-1})$ (seq), and infers its posterior latent $z_t$ from $h_t$ and the embedding $e_t$ (post); the mixed branch additionally predicts the prior latent $\hat{z}_t^{m}$ from $h_t^{m}$ alone (prior). The weak-only posterior $z_t^{w}$ serves as the reference toward which DC trains the mixed-branch prior $\hat{z}_t^{m}$ ($\mathcal{L}_{\mathrm{DC,dyn}}$) and regularizes the mixed-branch posterior $z_t^{m}$ ($\mathcal{L}_{\mathrm{DC,rep}}$). The full mixed-branch state $s_t^{m}=(h_t^{m},z_t^{m})$ is passed to the decoder in (a) and to the reward, continuation, actor, and critic heads ($\hat{r}_t,\hat{c}_t,\hat{a}_t,\hat{v}_t$).}
    \label{fig:dreamer_vigor_overview}
\end{figure}

Figure~\ref{fig:dreamer_vigor_overview} illustrates the integration, and Algorithm~\ref{algo:dreamer_vigor} summarizes world-model learning in DreamerV3+VIGOR; as in Algorithm~\ref{algo:VIGOR}, only the steps highlighted in \colorbox{yellow!30}{yellow} are added to (or modify) the original DreamerV3 procedure.

\begin{algorithm}[h]
\caption{World model learning in DreamerV3+VIGOR}\label{algo:dreamer_vigor}
\begin{algorithmic}
\STATE {\bfseries Input:} Replay buffer $\mathcal{B}$; Sequence horizon $H$; Weak and strong augmentation functions $\tau^w,\tau^s$; Loss weights $\beta_{\mathrm{dyn}},\beta_{\mathrm{rep}},\beta_{\mathrm{ES}}$; Learning rate $\eta$
\WHILE{not converged}
    \FOR{gradient step $t_g=1,2,\dots,T_g$}
        \STATE $\Gamma=(x_i,a_{i-1},r_i,c_i)_{i=t}^{t+H}\sim\mathcal{B}$ \hfill \COMMENT{Sample a sequence from the replay buffer}
        \STATE \colorbox{yellow!30}{$\upsilon^w\sim\Upsilon^w,\upsilon^s\sim\Upsilon^s$} \hfill \COMMENT{Sample augmentation parameters per sequence}
        \STATE $L\leftarrow 0$ \hfill \COMMENT{Initialize cumulative loss}
        \STATE \colorbox{yellow!30}{$(h^m_{t-1},z^m_{t-1})\leftarrow(h_{t-1},z_{t-1})$} \hfill \COMMENT{Initialize from the latent stored in $\mathcal{B}$}
        \STATE \colorbox{yellow!30}{$(h^w_{t-1},z^w_{t-1})\leftarrow(h^m_{t-1},z^m_{t-1})$} \hfill \COMMENT{Start both branches from the same state}
        \FOR{$i=t,t+1,\dots,t+H$}
            \STATE \colorbox{yellow!30}{$x^w_i\leftarrow\tau^w(x_i;\upsilon^w),\ x^{ws}_i\leftarrow\tau^s(x^w_i;\upsilon^s)$} \hfill \COMMENT{Build weak-only and weak-to-strong views}
            \STATE \colorbox{yellow!30}{$e^m_i\leftarrow e_\phi(x^w_i)\oplus e_\phi(x^{ws}_i)$} \hfill \COMMENT{Form mixed encodings by batch-wise concat}
            \STATE \colorbox{yellow!30}{$h^w_i\leftarrow f_\phi(h^w_{i-1},z^w_{i-1},a_{i-1})$} \hfill \COMMENT{Weak-only branch (stop-gradient)}
            \STATE \colorbox{yellow!30}{$z^w_i\sim \mathrm{sg}(q_\phi(\cdot\mid h^w_i,\mathrm{sg}(e_\phi(x^w_i))))$}
            \STATE $h^m_i\leftarrow f_\phi(h^m_{i-1},z^m_{i-1},a_{i-1})$ \hfill \COMMENT{Mixed branch}
            \STATE $z^m_i\sim q_\phi(\cdot\mid h^m_i,{\setlength{\fboxsep}{1pt}\colorbox{yellow!30}{$e^m_i$}})$
            \STATE $\mathcal{L}_{\mathrm{pred}}\leftarrow-\ln p_\phi({\setlength{\fboxsep}{1pt}\colorbox{yellow!30}{$x^w_i$}},r_i,c_i\mid h^m_i,z^m_i)$ \hfill \COMMENT{Calculate prediction loss}
            \STATE \colorbox{yellow!30}{$\mathcal{L}_{\mathrm{DC}}\leftarrow\beta_{\mathrm{dyn}}\mathcal{L}_{\mathrm{DC,dyn}}+\beta_{\mathrm{rep}}\mathcal{L}_{\mathrm{DC,rep}}$} \hfill \COMMENT{Replaces $\mathcal{L}_{\mathrm{dyn}},\mathcal{L}_{\mathrm{rep}}$; Eq.~\eqref{eq:dreamer_dc} at step $i$}
            \STATE \colorbox{yellow!30}{$\mathcal{L}_{\mathrm{ES}}\leftarrow\|\bar e_\phi(x^{ws}_i)-\mathrm{sg}(\bar e_\phi(x^w_i))\|_2^2$} \hfill \COMMENT{$\bar e_\phi=e_\phi/\|e_\phi\|_2$; $e_\phi$: image encoder of $q_\phi$}
            \STATE $L\leftarrow L+\mathcal{L}_{\mathrm{pred}}+{\setlength{\fboxsep}{1pt}\colorbox{yellow!30}{$\mathcal{L}_{\mathrm{DC}}+\beta_{\mathrm{ES}}\mathcal{L}_{\mathrm{ES}}$}}$ \hfill \COMMENT{Accumulate loss}
        \ENDFOR
        \STATE $\phi\leftarrow\phi-\eta\frac{1}{H+1}\nabla_\phi L$ \hfill \COMMENT{Update world model parameters}
    \ENDFOR
\ENDWHILE
\end{algorithmic}
\end{algorithm}

\subsection{Notation}
Following the DreamerV3 notation, $x_t$ denotes the visual observation (here a single frame $o_t$, since DreamerV3 uses no frame stacking, whereas $x_t$ denotes the frame stack $\{o_t,o_{t-1},o_{t-2}\}$ in Section~\ref{sec:VIGOR}; see Table~\ref{tab:common_hyperparams}), $h_t$ denotes the deterministic recurrent state of the RSSM, and $z_t$ denotes its stochastic categorical latent; note that $z_t$ here is the stochastic state of the RSSM, whereas in Section~\ref{sec:VIGOR} it denotes the encoder output of TD-MPC2. We write $e_t=e_\phi(x_t)$ for the encoder embedding supplied to the posterior, with $e_t^{w}$, $e_t^{ws}$, and $e_t^{m}$ denoting the embeddings of the respective views (Figure~\ref{fig:dreamer_vigor_overview}(a)); ES operates on its $\ell_2$-normalized version $\bar e_\phi(x_t)=e_\phi(x_t)/\lVert e_\phi(x_t)\rVert_2$ (Algorithm~\ref{algo:dreamer_vigor}). 
The posterior $q_\phi(z_t \mid h_t, e_t)$ infers the stochastic latent from the current observation together with the recurrent state, whereas the prior $p_\phi(z_t \mid h_t)$ predicts the latent from the recurrent state without access to the current observation. Throughout this section, $h_t$ denotes the RSSM recurrent state and should not be confused with the visual encoder $h_\theta$ introduced in Section~\ref{sec:VIGOR}; likewise, $s_t=(h_t,z_t)$ in Figure~\ref{fig:dreamer_vigor_overview}(b) denotes the full RSSM state. Figure~\ref{fig:dreamer_vigor_overview}(a) illustrates consecutive steps $t,\dots,t+H$ of a sampled training sequence ($H+1$ steps), which Algorithm~\ref{algo:dreamer_vigor} indexes as $i=t,\dots,t+H$; as in Algorithm~\ref{algo:VIGOR}, $t$ is the first step of the sampled sequence and $i$ the running index. As in Section~\ref{sec:VIGOR}, superscripts $w$, $ws$, and $m$ denote quantities computed from the weak-only sequence, the weak-to-strong sequence, and the mixed (weak-only~$\oplus$~weak-to-strong) batch, respectively.

\subsection{Where the Two-Level Vulnerability Appears in DreamerV3}\label{sec:dreamer_vulnerability}
The recursive pathway differs from that of TD-MPC2. At deployment, DreamerV3 does not perform online planning; instead, the actor selects actions from the current RSSM state $(h_t,z_t)$, where $z_t$ is inferred from the current observation by the posterior. A visual perturbation can therefore alter the posterior state used for the current action and, because $z_t$ contributes to the recurrent update, can also influence subsequent deterministic states and latent predictions. Unlike in TD-MPC2, where representation errors can compound along a finite-horizon planning rollout, this second level of sensitivity in DreamerV3 arises through temporal recurrence of the RSSM across environment steps. During training, the actor--critic is optimized on imagined trajectories generated by the prior dynamics, so robustness of both posterior inference and prior prediction is important.

The DC objective introduced below (Appendix~\ref{sec:dreamer_integration}) targets both pathways: its representation term constrains posterior inference from perturbed observations, and its dynamics term constrains the prior dynamics propagated from them, with both anchored to the weak-only posterior at every time step of the sequence.

\subsection{Background: KL Balancing in DreamerV3}
DreamerV3 already regularizes consistency between its posterior and prior latent distributions through two KL terms that share the same posterior--prior pair but place the stop-gradient on opposite sides:
\begin{align}\label{eq:dreamer_kl}
\mathcal{L}_{\mathrm{dyn}}
  &= D_{\mathrm{KL}}\!\left[\,\mathrm{sg}\!\left(q_\phi(z_t \mid h_t, e_t)\right) \,\big\|\, p_\phi(z_t \mid h_t)\right],\\
\mathcal{L}_{\mathrm{rep}}
  &= D_{\mathrm{KL}}\!\left[\,q_\phi(z_t \mid h_t, e_t) \,\big\|\, \mathrm{sg}\!\left(p_\phi(z_t \mid h_t)\right)\right].
\end{align}
The dynamics loss trains the prior to predict the stop-gradient posterior, whereas the representation loss regularizes the posterior toward the stop-gradient prior, making the inferred representations easier for the dynamics model to predict. DreamerV3 weights these terms by $\beta_{\mathrm{dyn}}=1$ and $\beta_{\mathrm{rep}}=0.1$ and applies a 1-nat free-bits threshold to both KL divergences. This asymmetric posterior--prior optimization follows the KL-balancing principle introduced in DreamerV2~\citep{hafner2020mastering}, while DreamerV3 combines it with free bits to prevent the KL objective from forcing the representations toward an overly predictable but uninformative solution.

These KL terms differ from the DC objective of Eq.~\eqref{eq:dc_loss} in two important respects. First, they operate on stochastic categorical latent distributions rather than deterministic latent vectors. Second, they impose consistency between the posterior and prior computed within the same observation stream; they do not directly constrain corresponding latent distributions obtained from different augmented views to agree. Consequently, the standard DreamerV3 KL objective provides no explicit cross-augmentation invariance signal, motivating the cross-view re-targeting introduced by VIGOR's DC objective.

\subsection{Integration}\label{sec:dreamer_integration}
The three components are integrated as follows.

\paragraph{AW.}
AW follows the construction in Section~\ref{sec:weak_strong}. Each sampled training sequence is transformed according to the weak-only branch ($\tau^w$) or the weak-to-strong branch ($\tau^s \circ \tau^w$) with $\zeta=1$, and the two views are encoded by the shared CNN encoder; their embeddings are concatenated along the batch dimension into the mixed embedding $e^m$ (Figure~\ref{fig:dreamer_vigor_overview}(a)), which is then processed by the RSSM. Because the DreamerV3 posterior is conditioned on the encoded observation at every time step, the augmented observation stream directly influences posterior inference throughout the training sequence. This differs from the TD-MPC2 instantiation of Eq.~\eqref{eq:dc_loss}, where the mixed representation is used only to construct the initial latent state of the rollout segment.

\paragraph{DC.}
We instantiate dynamics-level consistency in DreamerV3 by re-targeting its two KL-balancing terms across the weak-only and mixed branches.
Specifically, the dynamics loss trains the prior predicted from the mixed sequence to match the stop-gradient posterior from the weak-only sequence, while the representation loss regularizes the mixed posterior toward the same stop-gradient posterior from the weak-only sequence:
\begin{align}\label{eq:dreamer_dc}
\mathcal{L}_{\mathrm{DC,dyn}}
  &= D_{\mathrm{KL}}\!\left[\,\mathrm{sg}\!\left(q_\phi(z^w_t \mid h^w_t, e^w_t)\right) \,\big\|\, p_\phi(\hat{z}^m_t \mid h^m_t)\right],\\
\mathcal{L}_{\mathrm{DC,rep}}\label{eq:dreamer_dc_rep}
  &= D_{\mathrm{KL}}\!\left[\,q_\phi(z^m_t \mid h^m_t, e^m_t) \,\big\|\, \mathrm{sg}\!\left(q_\phi(z^w_t \mid h^w_t, e^w_t)\right)\right].
\end{align}
Here, the weak-only and mixed sequences are processed by the same RSSM, producing their respective posterior and prior distributions at each time step. Stop-gradient is applied only to the reference distribution in each KL term, preserving the directional optimization of DreamerV3's original KL balancing. We replace the corresponding standard DreamerV3 dynamics and representation KL terms with Eqs.~\eqref{eq:dreamer_dc}--\eqref{eq:dreamer_dc_rep}, while retaining the original loss weights, $\beta_{\mathrm{dyn}}=1$ and $\beta_{\mathrm{rep}}=0.1$, as well as the 1-nat free-bits threshold.

This construction transfers the consistency principle of Eq.~\eqref{eq:dc_loss} to DreamerV3 while respecting the structure of its stochastic recurrent state-space model. Rather than regressing deterministic latent vectors with an $\ell_2$ distance, DreamerV3+VIGOR aligns categorical latent distributions through KL divergence. Moreover, because the RSSM operates recurrently over sequence batches, the cross-view alignment is imposed at every time step, with the weak-only posterior serving as a shared stop-gradient reference for both KL terms; this encourages the latent dynamics and representations induced by perturbed observations to remain consistent with those induced by the weakly augmented sequence. In contrast to the original DreamerV3 representation loss, the representation term of the DC objective therefore no longer regularizes the posterior toward the prior. As in the TD-MPC2 instantiation, we use the weakly augmented view, rather than an unaugmented clean view, as the reference branch for the reasons discussed in Section~\ref{sec:latent_consistency}.

\paragraph{ES.}
We apply the encoder-level stabilization objective of Eq.~\eqref{eq:es_loss} to the observation embedding $e_t$ supplied to the RSSM posterior. At each time step, the encoder embedding obtained from the weak-to-strong view ($e_t^{ws}$) is aligned with the stop-gradient embedding obtained from the corresponding weak-only view ($e_t^{w}$). This imposes augmentation invariance before the observation enters the recurrent state-space model, complementing DC, which enforces consistency at the level of the stochastic latent dynamics. By reducing augmentation-induced variation in the encoder output, ES provides the posterior with a more stable observation representation across the two views, including at the beginning of a sequence where recurrent context is limited.

\subsection{Experimental Setup}
DreamerV3+VIGOR is trained and evaluated under the protocol of the main experiments (Section~\ref{sec:exp_setup}; Appendix~\ref{sec:appendix_evaluation_details}): agents are trained in clean environments on the six DMC and four Robosuite tasks and evaluated zero-shot under the seven RL-ViGen distraction types at every difficulty level, with the same metrics as in Section~\ref{sec:results}. The DreamerV3 baseline of the main experiments serves as the backbone reference, so the comparison between DreamerV3 and DreamerV3+VIGOR isolates the effect of VIGOR. DreamerV3+VIGOR uses the same architecture, optimization settings, sequence length, batch size, and world-model loss scales as the DreamerV3 baseline (DreamerV3 column of Table~\ref{tab:common_hyperparams}; official defaults otherwise), including the KL loss weights and free-bits threshold stated above. No backbone-specific hyperparameters are changed. The VIGOR-specific hyperparameters (the weak and strong augmentations $\tau^w$ and $\tau^s$ and the mixture ratio $\zeta$) are shared with the TD-MPC2 instantiation and correspond to the highlighted rows of Table~\ref{tab:VIGOR_hyperparams}. The ES loss weight is set to $\beta_{\mathrm{ES}}=\,1$. All results are averaged over ten random seeds.

\subsection{Results and Discussion}
\begin{figure}[ht]
\centering
\includegraphics[width=\textwidth]{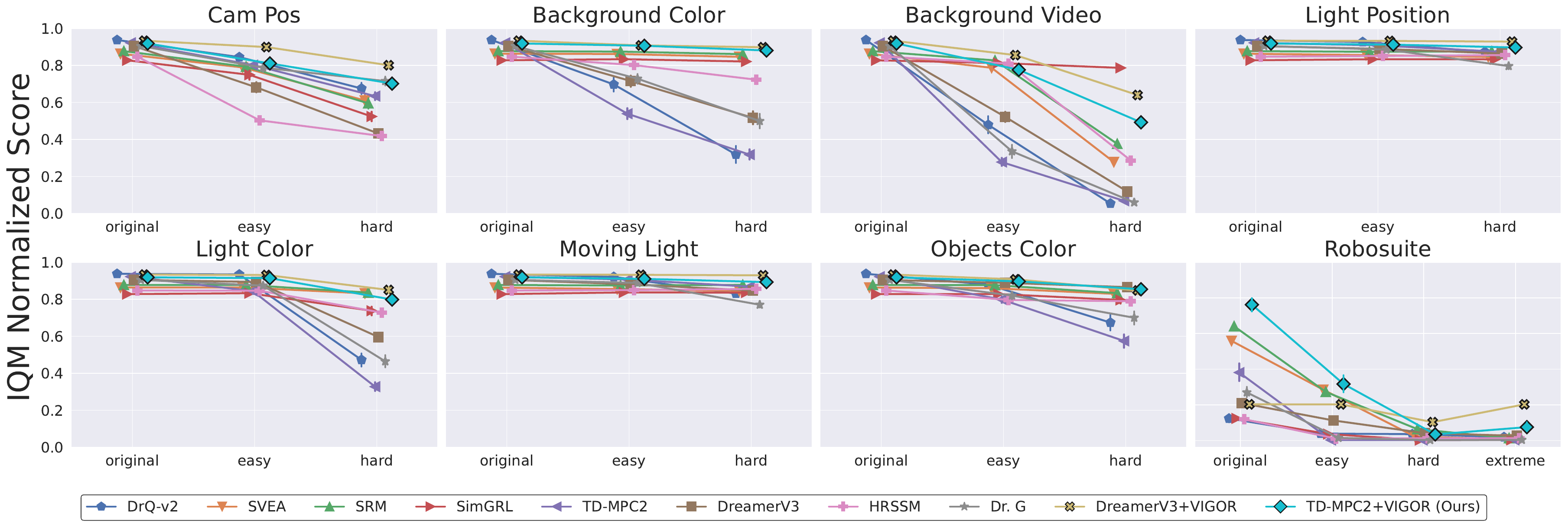}
\captionsetup{font=small}
\caption{\textbf{Zero-shot generalization across distraction types and difficulty levels, including DreamerV3+VIGOR.} Layout follows Figure~\ref{fig:eval_results_over_eval_types}: each subplot reports IQM normalized scores for a single distraction type across difficulty levels. DMC: seven distraction types $\times$ three settings (\textit{original}, \textit{easy}, \textit{hard}). Robosuite: four settings (\textit{original}, \textit{easy}, \textit{hard}, \textit{extreme}).}
\label{fig:dreamer_vigor_breakdown}
\end{figure}

\begin{figure}[ht]
\centering
\includegraphics[width=\textwidth]{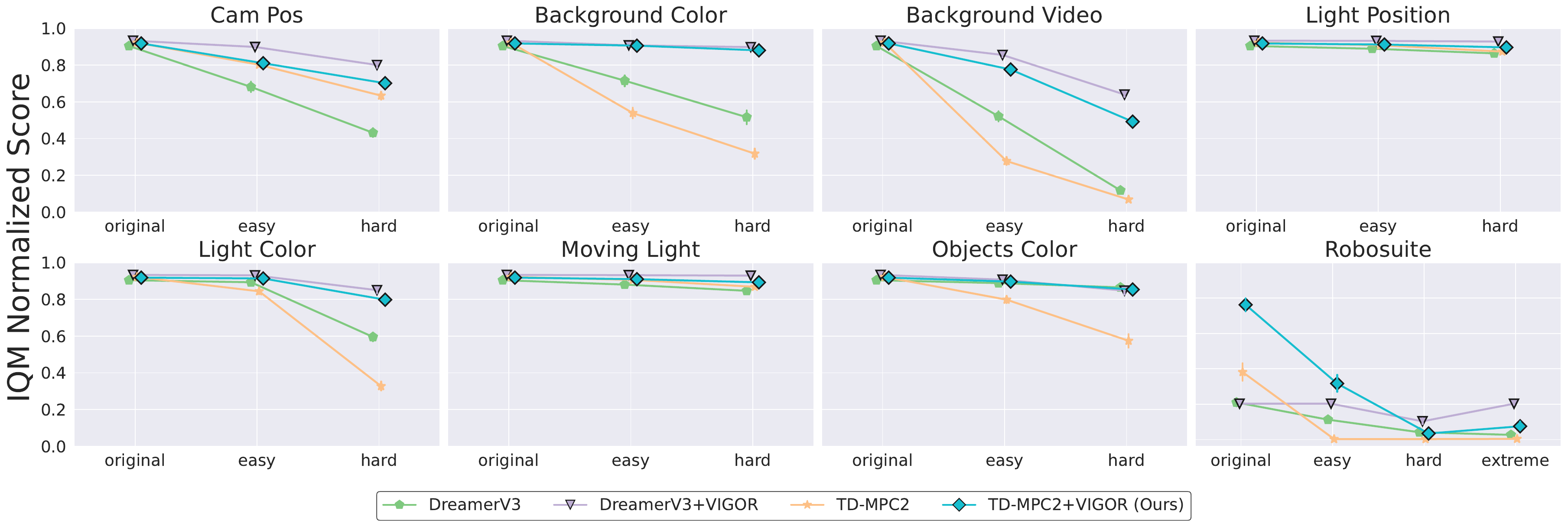}
\captionsetup{font=small}
\caption{\textbf{Zero-shot generalization across distraction types and difficulty levels for DreamerV3 and TD-MPC2 backbones.} Each subplot reports IQM normalized scores for a single distraction type across difficulty levels. DMC: seven distraction types $\times$ three settings (\textit{original}, \textit{easy}, \textit{hard}). Robosuite: four settings (\textit{original}, \textit{easy}, \textit{hard}, \textit{extreme}).}
\label{fig:dreamer_vigor_backbone_breakdown}
\end{figure}

\begin{figure}[ht]
\centering
\includegraphics[width=\textwidth]{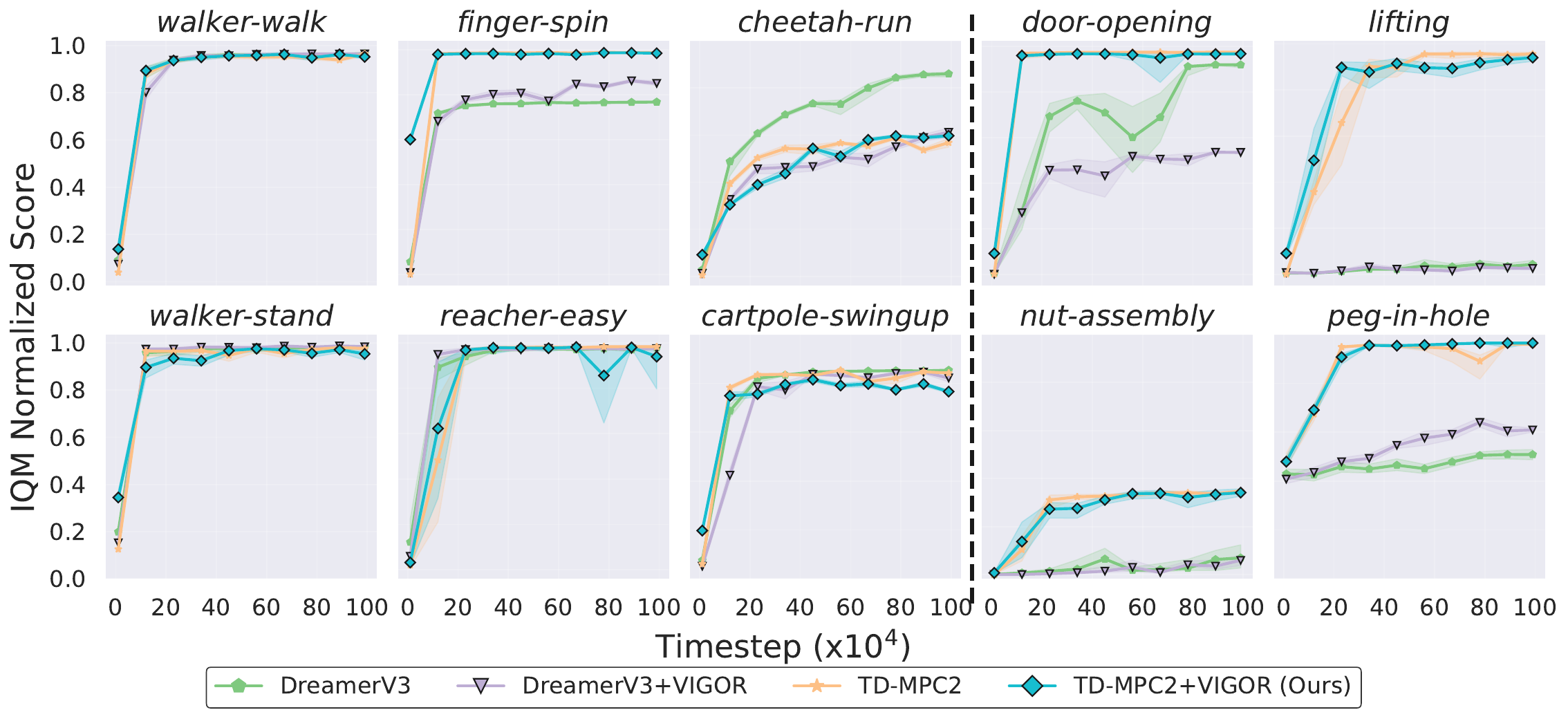}
\captionsetup{font=small}
\caption{\textbf{Per-task sample efficiency for DreamerV3 and TD-MPC2 backbones.} Learning curves reported as IQM normalized scores across methods. Shaded regions denote 95\% stratified bootstrap confidence intervals.}
\label{fig:eval_during_train_results_over_envs_dreamer_vigor_backbone}
\end{figure}

Table~\ref{tab:dreamer_vigor_tasks} reports per-task zero-shot generalization for DreamerV3+VIGOR alongside all baselines and TD-MPC2+VIGOR, and Figure~\ref{fig:dreamer_vigor_breakdown} breaks the results down by distraction type and difficulty level. Table~\ref{tab:dreamer_vigor_tasks_backbones} and Figure~\ref{fig:dreamer_vigor_backbone_breakdown} isolate the two backbones and their VIGOR instantiations, and Figure~\ref{fig:eval_during_train_results_over_envs_dreamer_vigor_backbone} reports the corresponding learning curves.

\paragraph{Zero-shot generalization.}
DreamerV3+VIGOR improves the average zero-shot return of its backbone from $635.3$ to $767.8$ on DMC ($+20.9\%$) and from $64.3$ to $98.1$ on Robosuite ($+52.6\%$). On DMC, the improvement holds on every task, and DreamerV3+VIGOR attains the best score among all methods on five of the six tasks; its average is second only to TD-MPC2+VIGOR ($777.6$) and exceeds every baseline, including the strongest baseline, SRM ($752.0$). The exception is \textit{finger-spin} ($458.3 \pm 441.4$), where the DreamerV3 backbone itself performs poorly ($268.2$) and the large variance indicates seed-dependent training. On Robosuite, the gains are concentrated on \textit{door-opening} ($35.0 \rightarrow 175.2$) and \textit{nut-assembly} ($3.4 \rightarrow 12.6$), whereas \textit{lifting} decreases ($23.7 \rightarrow 12.6$) and \textit{peg-in-hole} is essentially unchanged ($195.2 \rightarrow 192.0$); the average again exceeds every baseline (SRM: $91.8$) but remains well below TD-MPC2+VIGOR ($131.8$).

Figures~\ref{fig:dreamer_vigor_breakdown} and~\ref{fig:dreamer_vigor_backbone_breakdown} show that the gains arise where the backbone is most fragile. On DMC, DreamerV3 degrades sharply under hard \textit{cam-pos}, \textit{background-color}, \textit{background-video}, and \textit{light-color} distractions, whereas DreamerV3+VIGOR degrades gracefully; on \textit{background-video} hard, it even retains a higher IQM normalized score than TD-MPC2+VIGOR. On Robosuite, DreamerV3+VIGOR scores considerably lower than TD-MPC2+VIGOR in the \textit{original} setting but maintains a nearly flat profile from \textit{easy} to \textit{extreme} and attains the highest IQM normalized score among all methods under the \textit{hard} and \textit{extreme} settings.

\begin{table}[ht]
\captionsetup{font=small}
\caption{\textbf{Zero-shot generalization scores including DreamerV3+VIGOR.} Mean episode returns ($\pm$ std) averaged over all distraction types and difficulty levels for each task. Baseline columns are identical to Table~\ref{tab:visual_generalization}. Boldface denotes the best score per task.}
\label{tab:dreamer_vigor_tasks}
\centering
\begin{small}
\begin{sc}
\resizebox{\textwidth}{!}{
\renewcommand{\arraystretch}{1.5}
\begin{tabular}{ll|cccc|cccc|cc}
\toprule
\multirow{2}{*}{Env} & \multirow{2}{*}{Task}
& \multicolumn{4}{c|}{MFRL}
& \multicolumn{4}{c|}{MBRL}
& \multicolumn{2}{c}{\textbf{VIGOR (Ours)}} \\
\cmidrule(lr){3-6}\cmidrule(lr){7-10}\cmidrule(lr){11-12}
 & & DrQ-v2 & SVEA & SRM & SimGRL
 & TD-MPC2 & DreamerV3 & HRSSM & Dr.~G
 & TD-MPC2+VIGOR (default) & DreamerV3+VIGOR \\
\midrule
\midrule
\multirow[c]{7}{*}{DMC}
 & walker-walk
 & $731.5 \pm 344.4$
 & $738.5 \pm 177.0$
 & $730.2 \pm 153.6$
 & $719.3 \pm 128.5$
 & $642.2 \pm 344.8$
 & $795.5 \pm 279.0$
 & $775.3 \pm 153.0$
 & $769.8 \pm 331.9$
 & \cellcolor{cyan!15}{$811.3 \pm 201.7$}
 & \cellcolor{cyan!15}{$\mathbf{927.7 \pm 144.7}$} \\
 & finger-spin
 & $721.6 \pm 373.0$
 & $739.8 \pm 246.4$
 & $801.5 \pm 319.8$
 & $787.9 \pm 242.5$
 & $643.9 \pm 336.7$
 & $268.2 \pm 301.6$
 & $734.6 \pm 319.1$
 & $584.2 \pm 321.9$
 & \cellcolor{cyan!15}{$\mathbf{809.1 \pm 212.2}$}
 & \cellcolor{cyan!15}{$458.3 \pm 441.4$} \\
 & cheetah-run
 & $379.3 \pm 295.5$
 & $429.1 \pm 179.1$
 & $443.7 \pm 228.9$
 & $206.1 \pm 25.6$
 & $318.4 \pm 243.4$
 & $364.3 \pm 281.2$
 & $323.8 \pm 85.7$
 & $413.7 \pm 286.7$
 & \cellcolor{cyan!15}{$461.6 \pm 176.7$}
 & \cellcolor{cyan!15}{$\mathbf{490.9 \pm 167.2}$} \\
 & walker-stand
 & $754.3 \pm 313.7$
 & $933.5 \pm 132.0$
 & $932.7 \pm 105.0$
 & $929.5 \pm 110.3$
 & $737.1 \pm 320.5$
 & $891.9 \pm 233.4$
 & $915.7 \pm 116.0$
 & $834.6 \pm 274.9$
 & \cellcolor{cyan!15}{$938.6 \pm 72.9$}
 & \cellcolor{cyan!15}{$\mathbf{979.7 \pm 81.8}$} \\
 & reacher-easy
 & $501.8 \pm 449.9$
 & $829.8 \pm 322.1$
 & $858.2 \pm 291.8$
 & $875.5 \pm 286.8$
 & $621.9 \pm 422.5$
 & $758.5 \pm 345.7$
 & $571.2 \pm 427.5$
 & $500.9 \pm 388.8$
 & \cellcolor{cyan!15}{$899.0 \pm 232.3$}
 & \cellcolor{cyan!15}{$\mathbf{925.2 \pm 187.5}$} \\
 & cartpole-swingup
 & $668.3 \pm 289.2$
 & $686.5 \pm 281.8$
 & $745.7 \pm 143.4$
 & $742.7 \pm 87.5$
 & $631.1 \pm 241.7$
 & $733.5 \pm 212.5$
 & $657.8 \pm 192.1$
 & $638.7 \pm 274.7$
 & \cellcolor{cyan!15}{$746.0 \pm 173.8$}
 & \cellcolor{cyan!15}{$\mathbf{825.2 \pm 138.3}$} \\
\cmidrule(lr){2-12}
 & Average
 & $626.1 \pm 375.1$
 & $726.2 \pm 279.4$
 & $752.0 \pm 269.8$
 & $710.2 \pm 292.9$
 & $599.1 \pm 349.8$
 & $635.3 \pm 363.2$
 & $663.1 \pm 308.4$
 & $623.7 \pm 347.5$
 & \cellcolor{cyan!30}{$\mathbf{777.6 \pm 241.5}$}
 & \cellcolor{cyan!30}{$767.8 \pm 309.7$} \\
\midrule
\multirow[c]{5}{*}{Robosuite}
 & door-opening
 & $7.6 \pm 32.2$
 & $76.1 \pm 98.1$
 & $92.5 \pm 90.9$
 & $1.0 \pm 0.4$
 & $1.0 \pm 1.0$
 & $35.0 \pm 41.5$
 & $6.8 \pm 12.7$
 & $13.2 \pm 69.0$
 & \cellcolor{cyan!15}{$143.9 \pm 201.6$}
 & \cellcolor{cyan!15}{$\mathbf{175.2 \pm 77.7}$} \\
 & lifting
 & $27.0 \pm 25.1$
 & $20.6 \pm 58.9$
 & $25.5 \pm 78.7$
 & $21.6 \pm 74.1$
 & $1.8 \pm 5.0$
 & $23.7 \pm 27.7$
 & $1.5 \pm 2.7$
 & $1.3 \pm 2.5$
 & \cellcolor{cyan!15}{$\mathbf{35.4 \pm 74.9}$}
 & \cellcolor{cyan!15}{$12.6 \pm 6.1$} \\
 & nut-assembly
 & $0.6 \pm 1.6$
 & $11.8 \pm 31.7$
 & $10.2 \pm 26.8$
 & $1.3 \pm 0.0$
 & $0.7 \pm 1.1$
 & $3.4 \pm 21.7$
 & $5.2 \pm 3.8$
 & $0.7 \pm 1.0$
 & \cellcolor{cyan!15}{$\mathbf{15.5 \pm 39.3}$}
 & \cellcolor{cyan!15}{$12.6 \pm 15.1$} \\
 & peg-in-hole
 & $128.8 \pm 45.0$
 & $222.6 \pm 86.6$
 & $239.1 \pm 114.2$
 & $137.3 \pm 12.9$
 & $157.4 \pm 35.9$
 & $195.2 \pm 39.1$
 & $166.9 \pm 24.1$
 & $200.8 \pm 28.7$
 & \cellcolor{cyan!15}{$\mathbf{332.6 \pm 114.2}$}
 & \cellcolor{cyan!15}{$192.0 \pm 24.8$} \\
\cmidrule(lr){2-12}
 & Average
 & $41.0 \pm 59.9$
 & $82.8 \pm 111.9$
 & $91.8 \pm 123.4$
 & $40.3 \pm 68.0$
 & $40.2 \pm 70.1$
 & $64.3 \pm 83.5$
 & $45.1 \pm 71.7$
 & $54.0 \pm 92.8$
 & \cellcolor{cyan!30}{$\mathbf{131.8 \pm 176.1}$}
 & \cellcolor{cyan!30}{$98.1 \pm 95.3$}\\
\bottomrule
\end{tabular}
}
\end{sc}
\end{small}
\end{table}

\begin{table}[ht]
\captionsetup{font=small}
\caption{\textbf{Zero-shot generalization scores for DreamerV3 and TD-MPC2 backbones.} Mean episode returns ($\pm$ std) averaged over all distraction types and difficulty levels for each task. Boldface denotes the best score per task.}
\label{tab:dreamer_vigor_tasks_backbones}
\centering
\begin{small}
\begin{sc}
\resizebox{0.8\textwidth}{!}{
\renewcommand{\arraystretch}{1.5}
\begin{tabular}{ll|cccc}
\toprule
Env & Task  & DreamerV3 & DreamerV3+VIGOR & TD-MPC2 & TD-MPC2+VIGOR (Ours) \\
\midrule
\midrule
\multirow[c]{7}{*}{DMC}
 & walker-walk
 & $795.5 \pm 279.0$
 & \cellcolor{cyan!15}{$\mathbf{927.7 \pm 144.7}$}
 & $642.2 \pm 344.8$
 & \cellcolor{cyan!15}{$811.3 \pm 201.7$} \\
 & finger-spin
 & $268.2 \pm 301.6$
 & \cellcolor{cyan!15}{$458.3 \pm 441.4$}
 & $643.9 \pm 336.7$
 & \cellcolor{cyan!15}{$\mathbf{809.1 \pm 212.2}$} \\
 & cheetah-run
 & $364.3 \pm 281.2$
 & \cellcolor{cyan!15}{$\mathbf{490.9 \pm 167.2}$}
 & $318.4 \pm 243.4$
 & \cellcolor{cyan!15}{$461.6 \pm 176.7$} \\
 & walker-stand
 & $891.9 \pm 233.4$
 & \cellcolor{cyan!15}{$\mathbf{979.7 \pm 81.8}$}
 & $737.1 \pm 320.5$
 & \cellcolor{cyan!15}{$938.6 \pm 72.9$} \\
 & reacher-easy
 & $758.5 \pm 345.7$
 & \cellcolor{cyan!15}{$\mathbf{925.2 \pm 187.5}$}
 & $621.9 \pm 422.5$
 & \cellcolor{cyan!15}{$899.0 \pm 232.3$} \\
 & cartpole-swingup
 & $733.5 \pm 212.5$
 & \cellcolor{cyan!15}{$\mathbf{825.2 \pm 138.3}$}
 & $631.1 \pm 241.7$
 & \cellcolor{cyan!15}{$746.0 \pm 173.8$} \\
\cmidrule(lr){2-6}
 & Average
 & $635.3 \pm 363.2$
 & \cellcolor{cyan!30}{$767.8 \pm 309.7$}
 & $599.1 \pm 349.8$
 & \cellcolor{cyan!30}{$\mathbf{777.6 \pm 241.5}$} \\
\midrule
\multirow[c]{5}{*}{Robosuite}
 & door-opening
 & $35.0 \pm 41.5$
 & \cellcolor{cyan!15}{$\mathbf{175.2 \pm 77.7}$}
 & $1.0 \pm 1.0$
 & \cellcolor{cyan!15}{$143.9 \pm 201.6$} \\
 & lifting
 & $23.7 \pm 27.7$
 & \cellcolor{cyan!15}{$12.6 \pm 6.1$}
 & $1.8 \pm 5.0$
 & \cellcolor{cyan!15}{$\mathbf{35.4 \pm 74.9}$} \\
 & nut-assembly
 & $3.4 \pm 21.7$
 & \cellcolor{cyan!15}{$12.6 \pm 15.1$}
 & $0.7 \pm 1.1$
 & \cellcolor{cyan!15}{$\mathbf{15.5 \pm 39.3}$} \\
 & peg-in-hole
 & $195.2 \pm 39.1$
 & \cellcolor{cyan!15}{$192.0 \pm 24.8$}
 & $157.4 \pm 35.9$
 & \cellcolor{cyan!15}{$\mathbf{332.6 \pm 114.2}$} \\
\cmidrule(lr){2-6}
 & Average
 & $64.3 \pm 83.5$
 & \cellcolor{cyan!30}{$98.1 \pm 95.3$}
 & $40.2 \pm 70.1$
 & \cellcolor{cyan!30}{$\mathbf{131.8 \pm 176.1}$} \\
\bottomrule
\end{tabular}
}
\end{sc}
\end{small}
\end{table}

\paragraph{Sample efficiency.}
Figure~\ref{fig:eval_during_train_results_over_envs_dreamer_vigor_backbone} shows that DreamerV3+VIGOR tracks the learning curves of DreamerV3 on most tasks and learns faster on \textit{peg-in-hole}, but converges to noticeably lower training returns on \textit{cheetah-run} and \textit{door-opening}. The added augmentation and consistency objectives thus incur a task-dependent cost in training return on the DreamerV3 backbone, in contrast to the near-lossless integration observed on TD-MPC2 (Section~\ref{sec:results}; Appendix~\ref{sec:appendix_se}).

\subsection{Summary}
The DreamerV3 instantiation demonstrates that VIGOR can be applied to a structurally distinct RSSM-based world model while preserving the latent-space consistency principle underlying VIGOR. AW retains the same weak-only and weak-to-strong construction, DC re-targets DreamerV3's KL-balancing terms toward the weak-only posterior across the two augmentation branches, and ES imposes consistency at the encoder level. Without backbone-specific tuning, DreamerV3+VIGOR improves the aggregate zero-shot generalization of DreamerV3 on both benchmarks and exceeds every baseline in mean return, although it remains below TD-MPC2+VIGOR, does not improve every Robosuite task, and incurs lower training returns on some tasks. The two instantiations thus differ in absolute performance and in their efficiency--robustness profiles, reflecting differences between the underlying backbones. Taken together, the results suggest that the benefits of VIGOR are not limited to the TD-MPC2 backbone, while the TD-MPC2 instantiation remains the stronger of the two.

\end{document}